\documentclass[]{academic_template}

\usepackage[toc,page,header]{appendix}

\usepackage{amsmath}
\usepackage{amssymb}

\usepackage{booktabs}
\usepackage{makecell}
\usepackage{arydshln}
\usepackage{pifont}

\usepackage{graphicx}
\usepackage{wrapfig}

\usepackage{xcolor}
\usepackage{colortbl}
\definecolor{kcgreen}{rgb}{0.1, 0.7, 0.2}
\definecolor{kcred}{rgb}{0.8, 0.1, 0.2}
\usepackage{algorithm}      
\usepackage{algpseudocode}  

\usepackage{listings}
\usepackage{url}

\usepackage{caption}
\usepackage{subcaption}

\setlogoheight{15mm}   
\setlogospacing{5mm}
\setsjtublue 

\settitlerulethickness{3pt}  

\abstractboxon 
\setabstractframecolor{gray} 
\setabstractbgcolor{gray!10} 
\setlogotolineshift{5mm} 
\setabstractfont{\fontsize{12}{14}\selectfont} 

\settoprulethickness{2.5pt}  
\setbottomrulethickness{1.5pt} 

\makeatletter
\renewcommand\authorformat[2][]{\authorfont{\sffamily #2$^{#1}$}}
\makeatother

\title{\includegraphics[width=1.0cm]{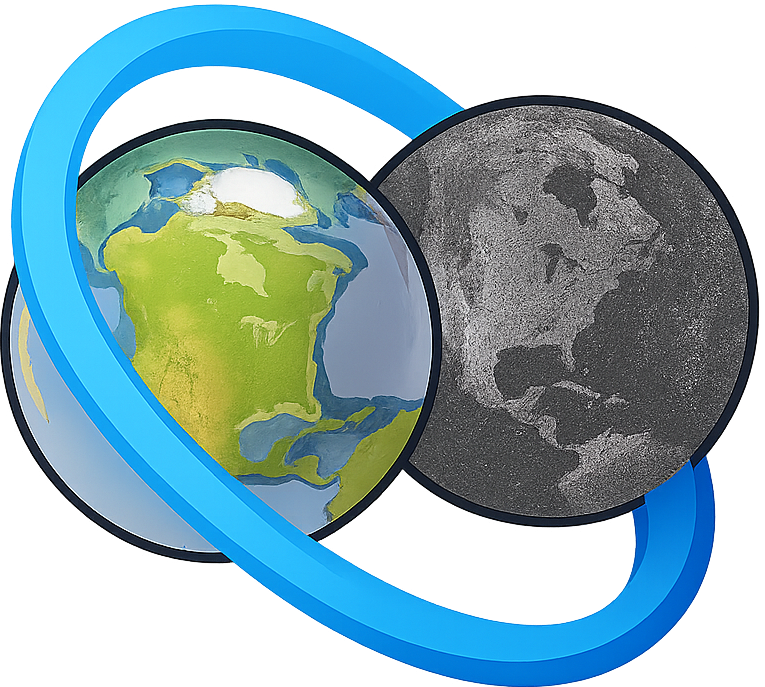} CrossEarth-SAR: A SAR-Centric and Billion-Scale Geospatial Foundation Model for Domain Generalizable Semantic Segmentation}

\author[1,2*]{Ziqi Ye}
\author[3*\dagger]{Ziyang Gong}
\author[3*]{Ning Liao}  
\author[2,3,4]{Xiaoxing Hu}
\author[5,6]{Di Wang}
\author[7]{Hongruixuan Chen} 
\author[8]{Chen Huang}
\author[3]{Yiguo He}
\author[9,10]{Yuru Jia}
\author[3]{Xiaoxing Wang}
\author[1\ddagger]{Haipeng Wang}
\author[3\ddagger]{Xue Yang}
\author[3,2\ddagger]{Junchi Yan}

\affiliation[1]{Fudan University}
\affiliation[2]{Shanghai Innovation Institute}
\affiliation[3]{Shanghai Jiao Tong University}
\affiliation[4]{Beijing Institute of Technology}
\affiliation[5]{Wuhan University}
\affiliation[6]{Zhongguancun Academy}
\affiliation[7]{The University of Tokyo}
\affiliation[8]{Sun Yat-sen University}
\affiliation[9]{KU Leuven}
\affiliation[10]{KTH}

\contribution[*]{Equal contribution}
\contribution[\dagger]{Project lead}
\contribution[\ddagger]{Corresponding Author}

\abstract{
Synthetic Aperture Radar (SAR) enables global, all-weather earth observation. However, owing to diverse imaging mechanisms, domain shifts across sensors and regions severely hinder its semantic generalization. To address this, we present CrossEarth-SAR, the first billion-scale SAR vision foundation model built upon a novel physics-guided sparse mixture-of-experts (MoE) architecture incorporating physical descriptors, explicitly designed for cross-domain semantic segmentation. To facilitate large-scale pre-training, we develop CrossEarth-SAR-200K, a weakly and fully supervised dataset that unifies public and private SAR imagery. We also introduce a benchmark suite comprising 22 sub-benchmarks across 8 distinct domain gaps, establishing the first unified standard for domain generalization semantic segmentation on SAR imagery. Extensive experiments demonstrate that CrossEarth-SAR achieves state-of-the-art results on 20 benchmarks, surpassing previous methods by over 10\% mIoU on some benchmarks under multi-gap transfer. All code, benchmark and datasets will be publicly available.
}

\date{\today}

\checkdata[Project Page]{\url{https://crossearth.cn/}}
\checkdata[Code]{\url{https://github.com/VisionXLab/CrossEarth-SAR}}
\checkdata[Benchmark Datasets]{\url{https://huggingface.co/datasets/conquer997/CrossEarth-SAR-Benchmarks}}

\begin{document}
\maketitle


\section{Introduction}
\label{sec:intro}

\begin{figure}[t]
\centering
\vspace{10pt}

\begin{minipage}[t]{0.49\textwidth}
  \centering
  \includegraphics[width=\linewidth]{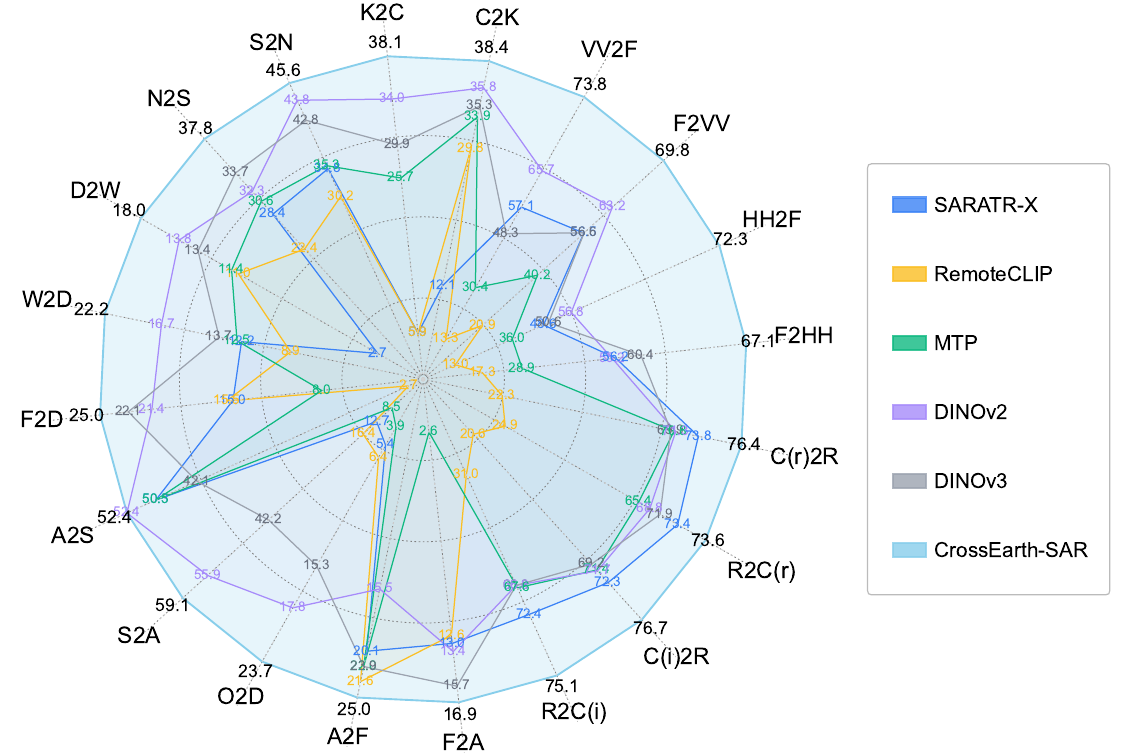}
  \vspace{3pt} 
  \caption{We evaluate representative models on 22 valuation benchmarks, where CrossEarth-SAR achieves SOTA performance (mIoU) on 20 settings across various segmentation scenes, demonstrating strong generalizability.}
  \label{fig:radar}
\end{minipage}
\hfill
\begin{minipage}[t]{0.49\textwidth}
  \centering
  \includegraphics[width=\linewidth]{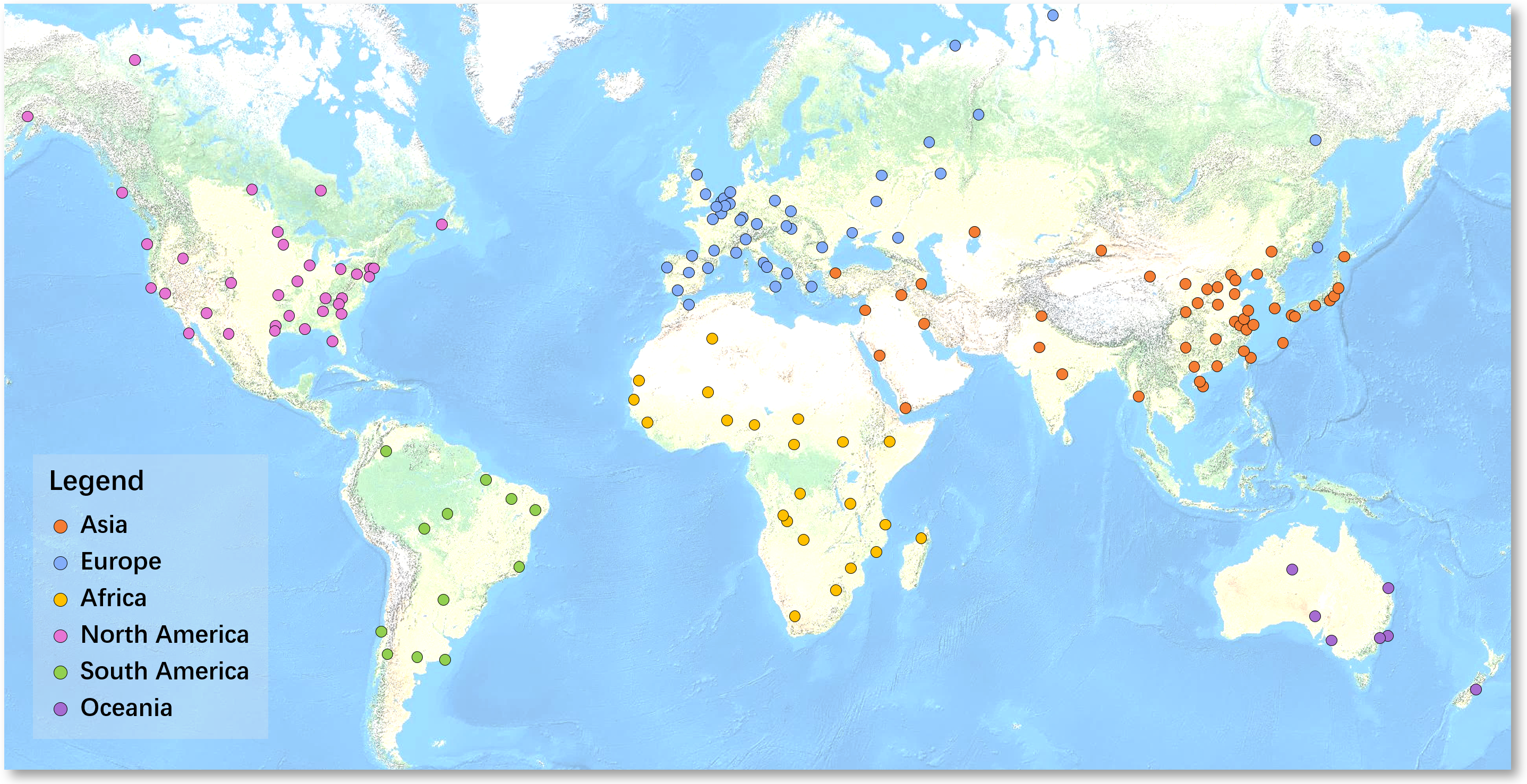}
  \vspace{3pt} 
  \caption{Geographic distribution of the CrossEarth-SAR-200K, demonstrating its comprehensive coverage across hundreds of cities on six continents.}
  \label{fig:earthmap}
\end{minipage}
\end{figure}

Synthetic Aperture Radar (SAR) is an indispensable tool for earth observation, prized for its ability to capture high-resolution imagery regardless of weather conditions or time of day. This all-weather, all-day capability is critical for time-sensitive applications, especially in disaster mitigation ~\cite{jensen_remote_2015}, as well as for long-term environmental monitoring ~\cite{kumar2012environmental} and urban management~\cite{mcgill1998urban}. At the core of translating complex SAR data into actionable insights lies in semantic segmentation, the task of assigning a class label to every pixel in a captured scene.

However, achieving robust semantic understanding in SAR is fundamentally more challenging than in the visible spectrum domain. This difficulty stems not only from the unfamiliar appearance of radar images but from three core challenges that directly conflict with the assumptions of modern vision models. First, SAR's coherent imaging process creates inherent multiplicative speckle, a granular noise that corrupts local feature statistics and textures, severely degrading the performance of models reliant on learned textural patterns~\cite{sar1,sar2,sar3,sar4}. Second, SAR's side-looking geometry introduces severe spatial distortions, such as layover, foreshortening, and shadow, which drastically alter the apparent shape and topology of objects like buildings and mountains, severely affect models trained on standard geometric priors~\cite{lee2005segmentation, yang2025fusarklip, wu2021novel}. Third, SAR measures radar backscatter, not color. This physical measurement, tied to surface roughness and dielectric properties, creates profound semantic ambiguity: dissimilar classes can share identical dark appearances, while a single class can appear vastly different based solely on moisture content or crop row orientation~\cite{robertson2023monitoring,wegmuller2011progress}.

Beyond these inherent pixel-level and geometric complexities lies the most significant barrier to scalable applications, \textit{i.e.}, the domain specificity of SAR. Different from web images widely used in the vision community, SAR data is not a monolith. Its characteristics are fragmented by a multitude of acquisition parameters, including sensor platform (e.g., Sentinel-1, ALOS-2, Capella), frequency band (C, L, X bands), polarization mode (HH, HV, VH, and VV), and incidence angle. A model trained on data from one sensor, in one region, under one set of conditions, catastrophically fails when applied to another. This extreme domain fragmentation has hindered the development of a truly generalizable and robust SAR understanding model.

In this context, achieving cross-domain generalization is the essential yet largely underexplored challenge for the SAR community~\cite{gong2024crossearth, hu2025earth}. To bridge this gap, we argue that unlocking large-scale generalization requires a massive model capacity to absorb the extreme diversity of SAR data. We introduce \textbf{CrossEarth-SAR}, the first \textbf{billion-scale} SAR vision foundation model. To make this scale computationally feasible, CrossEarth-SAR is built upon a sparse Mixture-of-Experts (MoE) architecture. This design allows the parameter count to scale into the billions (to capture extreme domain diversity) while maintaining a manageable compute cost per image. Additionally, we also incorporate a novel physics-guided routing mechanism to further stabilize expert selection for SAR's unique physical properties.

To support the training and rigorous evaluation of such a large-scale model, we introduce two additional resources. First, the CrossEarth-SAR-200K dataset, which integrates 200K supervised images by combining public and privately collected SAR data. Second, a comprehensive benchmark collection encompassing 22 benchmarks across 8 domain gaps, providing the first unified standard for assessing DG in SAR imagery.

Our main contributions are summarized as follows:
\begin{itemize}
    \item We propose \textbf{CrossEarth-SAR}, the first \textbf{billion-scale} SAR vision foundation model featuring a physics-guided sparse MoE architecture, specifically optimized for cross-domain semantic segmentation. In addition, we also provide small, base, and large versions of CrossEarth-SAR to support hierarchical application needs.

    \item We construct \textbf{CrossEarth-SAR-200K}, a large-scale semantic segmentation dataset integrating weakly and fully supervised SAR imagery to support global-scale continuous pre-training.

    \item We curate a standardized benchmark collection encompassing 22 sub-benchmarks across 8 domain gaps, enabling unified evaluation of SAR domain generalization.

    \item Extensive experiments and analyses show the superiority of CrossEarth-SAR and provide new insights into domain-generalizable SAR representation learning.
\end{itemize}

\begin{figure*}[!tb]
  \centering
  \includegraphics[width=0.9\textwidth]{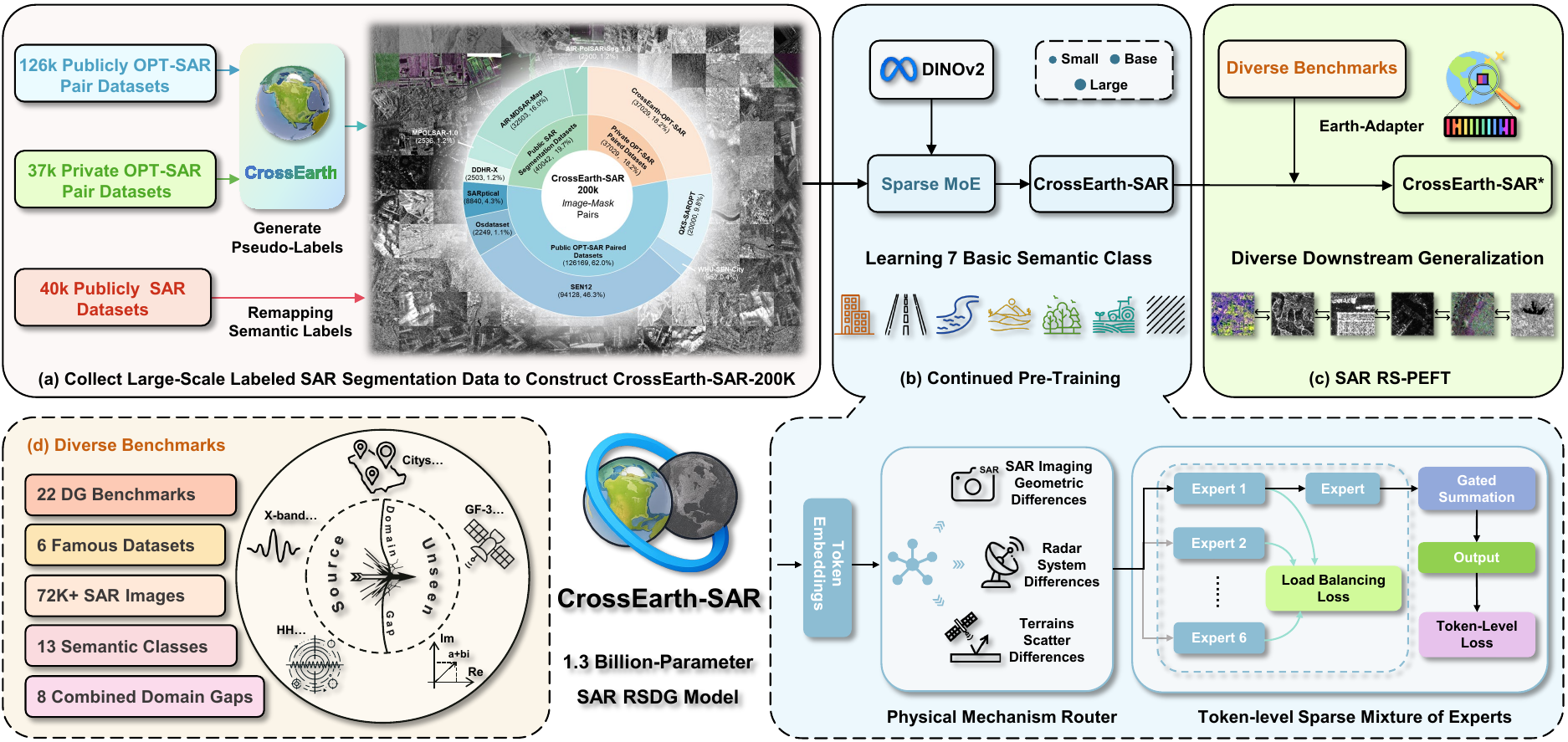}
  \caption{Framework of CrossEarth-SAR. (a) Collect large-scale labeled SAR segmentation data to construct CrossEarth-SAR-200K. (b) The continued pre-training structure of CrossEarth-SAR. (c) SAR RS-PEFT by Earth-Adapter. (d) The information of diverse benchmarks.} 
  \label{fig:CrossEarth-SAR}
\end{figure*}
\section{Related Work}
\label{sec:related}


\subsection{Cross-Domain Semantic Segmentation in RS}
\label{subsec:area1}

Cross-domain semantic segmentation is a fundamental task for testing model generalization. While notable progress has been made in fields like medicine~\cite{guan2021domain, qiu2025noise, qiu2025adaptively} and autonomous driving~\cite{coda, sta, parsing, jia2025can}, the problem remains particularly acute in RS. RS imagery suffers from severe domain gaps caused by differences in sensors, resolution, season, spectral bands, and geographic conditions~\cite{Reichstein2019Nature}. Within RS, efforts in the visible spectrum domain have evolved from conventional adversarial alignment and self-training methods~\cite{Zou2018ECCVCBST,Zou2019ICCVCRST,Chen2019BMVCUDAST,Zhu2017CycleGAN,Benjdira2019RS,Li2020RemoteSensingCrossCity,Marmanis2018ISPRS} to the recent adoption of Vision Foundation Models (VFMs)~\cite{guo2024skysense,Zhang2024HyperSIGMA,Zhang2023PerSAM,wang2024mtp}. Notably, CrossEarth~\cite{gong2024crossearth}, the first VFM for domain-generalized segmentation, has set a strong benchmark for optical generalization.

However, in challenging SAR scenarios, achieving generalization remains difficult~\cite{Saha2020IGARSSUDA,Nguyen2021IGARSSDG} due to the unique characteristics of SAR imagery, such as speckle noise and complex electromagnetic scattering mechanisms~\cite{ye2024convolutional,Schmitt2019SEN12MS,Wang2022SARTransformerSeg}. Existing mainstream methods often rely on paired optical images to support SAR semantic segmentation~\cite{zhang2024asanet, shi2021object, wei2024mgfnet, xia2023openearthmap}, but their generalization capabilities remain limited to relatively narrow domain gaps. How to fully unlock the potential of FMs for cross-domain generalization in SAR imagery remains largely underexplored.

\subsection{SAR Vision Foundation Models}
The development of geospatial foundation models has been largely concentrated on optical and temporal multispectral data. Prominent examples include SatMAE~\cite{cong2022satmae}, HyperSIGMA~\cite{wang2025hypersigma}, SkySense~\cite{guo2024skysense}, SkySense++~\cite{wu2025skysensepp} and SkySenseV2~\cite{zhang2025skysensev2}. While powerful in their domain, their architectures and pre-training strategies do not address the unique backscattering physics and noise characteristics of SAR. Recently, VFMs focusing on SAR have emerged. A significant portion of these efforts centers on object-level applications, such as object detection~\cite{li2024sardet} or automatic target recognition \cite{li2025saratr}. These models are optimized for sparse object localization, which does not directly address dense, pixel-level semantic segmentation. Other efforts are geared towards learning general-purpose representations~\cite{li2024predicting, wang2025complex}. While they show promise, they are not designed for cross-domain generalization. Thus, a large-scale VFM that is built to achieve robust domain-generalizable semantic segmentation in SAR remains a critical missing piece.
\section{Method}
\label{sec:method}


\subsection{Overview}
\label{subsec:overview}
The architecture of CrossEarth-SAR is a physics-guided sparse MoE integrated into a DINOv2 backbone $f_{\phi}$. The backbone follows the standard ViT stacking. In each block, the original Feed-Forward Network (FFN) is replaced by a MoE composed of a router $R_{\psi}$ and $n$ experts $\{E_k\}_{k=1}^{n}$, which are initialized from the DINOv2 FFN weights.

At each training iteration, an input SAR image is first copied to 3-channel $X \in \mathbb{R}^{3 \times H \times W} $ and forwarded through $ f_{\phi}$. In parallel, three SAR physics descriptors $ s \in \mathbb{R}^{3} $ are computed. For each ViT block, the token embeddings produced by attention are augmented with $s$ and passed to the router $ R_{\psi} $ to activate an expert. A load-balancing constraint $ \mathcal{L}_{\mathrm{BC}} $ is applied during the routing process.

After propagating through all blocks, the final embeddings are decoded by a Mask2Former $ f_{\theta} $ to generate the prediction $\hat{Y} $. During training, the parameters of $ f_{\phi} $, $ R_{\psi} $, $\{E_k\}_{k=1}^{n} $, and $ f_{\theta} $ are jointly optimized using the segmentation objective $\mathcal{L}_{seg}$ together with $\mathcal{L}_{\mathrm{BC}}$. In the following, we illustrate the details and motivations of the components in CrossEarth-SAR.

\subsection{SAR Physical Descriptor}
\label{subsec:component1}
A critical challenge in applying sparse MoE architectures to SAR is `Routing Instability'. Standard routers select experts based on learned token embeddings. However, these embeddings fluctuate dramatically when processing heterogeneous SAR data due to severe physical domain shifts. This prevents the router from consistently activating the appropriate expert for a given domain, undermining the MoE's specialization objective.

To mitigate this issue, CrossEarth-SAR introduces a SAR Physical Operator $g_{\mathrm{sar}}(\cdot)$ to compute three representative descriptors for each input image. These descriptors provide stable, physically grounded cues that complement token embeddings and supply physics-aware guidance to the router, enabling more consistent expert activation. 

Notably, before calculation, we transform all input $X$ to a log-intensity representation $X^\prime$ for numerical stability by
\begin{equation}
    X^\prime=\log\big(1+\lvert X\rvert\big).
\end{equation}
Next, we will illustrate the details of three descriptors.

\textbf{Imaging Geometry.}
SAR image appearance is sensitive to imaging geometry. Changes in sensor position or viewing angle alter object shapes and edge orientations, causing inconsistent features of the same target across domains.
To quantify this difference, we introduce the Directional Entropy $H_{DE}$, which evaluates the structural regularity of an image by analyzing the distribution of gradient directions in $X^\prime$.
Given the Sobel gradients ($g_x$, $g_y$), we first compute the gradient direction for each pixel:
\begin{equation}
\theta = \operatorname{atan2}(g_y, g_x).
\end{equation}
These direction angles are then quantized into $N$ discrete bins to form a histogram of gradient orientations.
Finally, the entropy of this histogram is computed as:
\begin{equation}
H_{DE} = - \sum_{i=1}^{N} p_i \ln p_i,
\end{equation}
where ($p_i$) denotes the normalized proportion of pixels whose gradient direction falls into the $i-th$ bin.

A low $H_{DE}$ indicates that image gradients are concentrated in a dominant direction, typically corresponding to clear linear or edge-like structures. A high value suggests that gradient directions are more evenly distributed and the scene contains richer, more irregular structural patterns.

\textbf{Radar System.} 
Variations in radar parameters, such as frequency band and polarization, lead to distinct speckle statistics across sensors, introducing inconsistent noise levels that challenge generalization. 
To capture speckle variations induced by the system imaging mechanism, we use the Equivalent Number of Looks (ENL), a widely adopted measure of speckle strength in SAR imaging. ENL is defined as:
\begin{equation}
\text{ENL} = \left( \frac{\mu}{\sigma} \right)^2,
\end{equation}
where $\mu$ and ($\sigma$) are the mean and standard deviation of the $X^\prime$. A higher ENL indicates more stable intensity statistics and weaker speckle, while a lower ENL reflects stronger noise fluctuations. This scalar provides a simple and reliable way to compare system-level noise properties across different SAR sources.

\textbf{Object Scattering.} 
Surface scattering varies with material roughness and dielectric properties, resulting in diverse texture patterns that differ significantly across domains and hinder consistent feature generalization.
To capture this texture variability, we compute a Local Roughness descriptor by applying spatial averaging to $X^\prime$ and measuring the variance among blockwise mean values:
\begin{equation}
R_{\text{LR}} = \text{Var}\left( { \mu_j } \right)_{j=1}^M,
\end{equation}
where $M$ is the total number of spatial blocks obtained from average pooling. Each blockwise mean $\mu_j$ is defined as:
\begin{equation}
\mu_j = \frac{1}{\lvert B_j\rvert}\sum_{(u,v)\in B_j} X^\prime(u,v),
\end{equation}
where ($B_j$) denotes the set of pixels within the $j-th$ local block of $X^\prime$. Consequently, a higher $R_{\text{LR}}$ value implies large variations in brightness across regions, corresponding to a scene composed of complex texture. Conversely, a lower value indicates a smooth texture.

Finally, the three physical descriptors are concatenated to form a 3D tensor:
\begin{equation}
s
= \big[ H_{DE}, \text{ENL}, R_{\mathrm{LR}} \big] \in \mathbb{R}^{3}.
\end{equation}

Ablation studies (see Tab.~\ref{tab:ablation_left}, Section 5.5) demonstrate that incorporating these physical quantities guides the model to focus on the inter-domain physical features of SAR images, effectively improving its domain generalization capability and downstream segmentation performance.

\begin{table*}[t]
\centering
\caption{CrossEarthSAR benchmark collection. North/South: northern/southern China. KR/CN: Korea/China. VV/HH: single-polarization. Full: full-polarization. Complex(r)/Complex(i): real or imaginary part of complex-valued PolSAR data.}
\label{tab:benchmark}
\small 
\setlength{\tabcolsep}{10pt}  
\renewcommand{\arraystretch}{1.2} 
\resizebox{1.0\textwidth}{!}{
\begin{tabular}{cccccccc}
\toprule
Domain Gap & Dataset & Source Domain & Target Domain & Abbreviation & Train & Test & Categories \\
\midrule
\multirow{4}{*}{\makecell{Unseen Region}} & \multirow{2}{*}{AIR-PolSAR-Seg 2.0~\cite{zhirui2025air}} & North & South & N2S & 600 & 159 & 5 \\
 &  & South & North & S2N & 633 & 150 & 5 \\
 & \multirow{2}{*}{DDHR-SK~\cite{ren2022dual}} & KR & CN & K2C & 4916 & 1566 & 6 \\
 &  & CN & KR & C2K & 6265 & 1229 & 6 \\
\cmidrule{1-8}
\multirow{4}{*}{\makecell{Unseen Polarization}} & \multirow{4}{*}{AIR-PolSAR-Seg 2.0~\cite{zhirui2025air}} & VV & Full & VV2F & \multirow{4}{*}{\centering 1233} & \multirow{4}{*}{\centering 309} & 6 \\
 &  & Full & VV & F2VV &  &  & 6 \\
 &  & HH & Full & HH2F &  &  & 6 \\
 &  & Full & HH & F2HH &  &  & 6 \\
\cmidrule{1-8}
\multirow{4}{*}{\makecell{Unseen Complex Values}} & \multirow{4}{*}{AIR-PolSAR-Seg 2.0~\cite{zhirui2025air}} & Complex(r) & Real & C(r)2R & \multirow{4}{*}{\centering 1233} & \multirow{4}{*}{\centering 309} & 6 \\
 &  & Real & Complex(r) & R2C(r) &  &  & 6 \\
 &  & Complex(i) & Real & C(i)2R &  &  & 6 \\
 &  & Real & Complex(i) & R2C(i) &  &  & 6 \\
\cmidrule{1-8}
Unseen Region and & FUSAR-Map~\cite{shi2021object} & FUSAR & AIR & F2A & 1952 & 309 & 5 \\
Polarization & AIR-PolSAR-Seg 2.0~\cite{zhirui2025air} & AIR & FUSAR & A2F & 1233 & 488 & 5 \\
\cmidrule{1-8}
Unseen Region and & OpenEarthMap~\cite{xia2023openearthmap} & OpenEarthMap & DDHR & O2D & 13865 & 1567 & 6 \\
Platform & DDHR-SK~\cite{ren2022dual} & DDHR & OpenEarthMap & D2O & 6265 & 3467 & 6 \\
\cmidrule{1-8}
Unseen Region and & SARBuD~\cite{wu2021built} & SARBuD & AIR & S2A & 20000 & 309 & 1 \\
Microwave Band & AIR-PolSAR-Seg 2.0~\cite{zhirui2025air} & AIR & SARBuD & A2S & 1233 & 5000 & 1 \\
\cmidrule{1-8}
Unseen Region, Polarization & DDHR-SK~\cite{ren2022dual} & DDHR & FUSAR & D2F & 6265 & 488 & 5 \\
and Microwave Band & FUSAR-Map~\cite{shi2021object} & FUSAR & DDHR & F2D & 1952 & 1566 & 5 \\
\cmidrule{1-8}
Unseen Region, Platform & WHU-OPT-SAR~\cite{li2022mcanet} & WHU-SAR & DDHR & W2D & 7040 & 1567 & 6 \\
and Microwave Band & DDHR-DK~\cite{ren2022dual} & DDHR & WHU-SAR & D2W & 6265 & 1760 & 6 \\
\bottomrule
\end{tabular}
}
\end{table*}

\subsection{Physics-Guided Sparse MoE}
\label{subsec:component2}
To model the inherent multi-source variability of SAR imagery, we try to curate a model with sufficient capacity to capture its diverse and complex imaging mechanisms. However, simply enlarging a dense network leads to prohibitive computational costs. The sparse MoE architecture provides a principled solution: its multi-expert design allows different experts to specialize in different SAR characteristics, while sparse activation ensures that the inference cost remains comparable to a standard FFN. Motivated by this, we design physics-guided Sparse MoE.

\textbf{Router.}
Given the token embeddings $Z\in\mathbb{R}^{B\times N\times C}$ and the descriptors $s$. We tile 
$s$ across tokens to get $S\in\mathbb{R}^{B\times N\times3}$. The router computes a score over $n$ experts and selects the top-$k$ candidates for each token:
\begin{equation}
    \pi = \mathrm{softmax}(W_r [Z\| S] + b_r),
\end{equation}
where $[\cdot]$ means the concat and $ \pi \in \mathbb{R}^{B\times N\times n}$ denotes the expert scores.

\textbf{Token-Level MoE.}
For each token embedding $z\in\mathbb{R}^{C}$, the router selects the top-$k$ experts according to scores $\pi\in\mathbb{R}^{n}$. Let $\mathcal{I}$ denote the set of selected expert indices; the gating weights are normalized as:
\begin{equation}
 g_{k} = \frac{\pi_{k}}{\sum_{m\in\mathcal{I}}\pi_{m}}, \quad k\in\mathcal{I},
\end{equation}
and the final output is the sparse aggregation, defined as:
\begin{equation}
 \tilde{z} = \sum_{k\in\mathcal{I}} g_{k}\cdot E_k(z).
\end{equation}
This design preserves computational efficiency comparable to the original FFN while allowing dynamic specialization across SAR imaging conditions.

\textbf{Training Objective.}
To prevent expert collapse, we impose a load-balancing loss. Let $f_k$ be the fraction of tokens assigned to expert $k$ and $p_k$ the mean routing probability. The balancing loss is as follows:
\begin{equation}
 \mathcal{L}_{\mathrm{BC}} = \lambda_{\mathrm{BC}}\,n\sum_{k=1}^{n} f_k p_k,
\end{equation}
where $\lambda_{\mathrm{BC}}$ is the weight set to 0.005. The complete training objective combines segmentation and balancing terms:
\begin{equation}
 \mathcal{L} = \mathcal{L}_{\mathrm{seg}} + \mathcal{L}_{\mathrm{BC}}.
\end{equation}

\section{CrossEarth-SAR-200K \& Benchmark}
\subsection{CrossEarth-SAR-200K}
\label{subsec:dataset}

CrossEarth-SAR-200K consists of publicly available SAR images with real labels and collected SAR images with pseudo labels, as shown in Figure~\ref{fig:CrossEarth-SAR}(a).

For data without semantic segmentation labels, the strongest RSDG model CrossEarth, is adopted to segment the optical images paired with SAR images. The R2U weights trained on 7 classes from the LoveDA dataset are used. The resulting predictions are assigned as labels for the corresponding SAR images. 
This pseudo-labeling pipeline follows a widely accepted practice in SAR-related semantic segmentation, and has been adopted by several representative benchmarks, such as DDHR~\cite{ren2022dual}, FUSAR-Map~\cite{shi2021object}, WHU-OPT-SAR~\cite{li2022mcanet}, and OpenEarthMap-SAR~\cite{xia2023openearthmap}. Additionally, to estimate label confidence, we apply four models (SatMAE~\cite{cong2022satmae}, ScaleMAE~\cite{reed2023scale}, MTP~\cite{wang2024mtp}, and CrossEarth~\cite{gong2024crossearth}) to predict pseudo labels on 1K samples. The mean agreement~\cite{xia2023openearthmap} across these models reaches 75.88\%, exceeding the 63.20\% computed on OpenEarthMap-SAR~\cite{xia2023openearthmap}, which supports the reliability of the generated labels.

For the continuous pre-training stage, CrossEarth-SAR-200K-Val is a 4k-image split sampled from CrossEarth-SAR-200K with ground-truth labels, and is used for validation. As part of preprocessing, all SAR images are cropped or resized to 512×512. More details on the composition and structure of CrossEarth-SAR-200K are provided in Fig.~\ref{fig:earthmap}, Fig.~\ref{fig:CrossEarth-SAR}(a). All CrossEarth-SAR-200K, including our collected dataset, will be fully available.

\subsection{Benchmark}
\label{subsec:benchmark}

For fair evaluation of generalizability of existing models in SAR modality, we curate 22 benchmarks based on 6 widely-used SAR RS semantic segmentation datasets including AIR-PolSAR-Seg-2.0~\cite{zhirui2025air}, DDHR-SK~\cite{ren2022dual}, FUSAR-Map~\cite{shi2021object}, OpenEarthMap-SAR~\cite{xia2023openearthmap}, SARBuD~\cite{wu2021built}, and WHU-OPT-SAR~\cite{li2022mcanet}, and extend them to DG settings, as shown in Tab.~\ref{tab:benchmark}. Our benchmark comprises \textbf{22} different tasks across 8 compositional domain gaps: (1) Unseen Region; (2) Unseen Polarization; (3) Unseen Complex Number; (4) Unseen Region and Polarization; (5) Unseen Region and Platform; (6) Unseen Region and Microwave Band; (7) Unseen Region, Polarization and Microwave Band; (8) Unseen Region, Platform and Microwave Band. The benchmark datasets are non-overlapping with CrossEarth-SAR-200K, and we confirm that there is no data leakage across datasets.

\begin{table*}[t]
\centering
\caption{Performance comparison across one-domain-gap benchmarks.
All models are frozen during downstream fine-tuning. \textbf{Bold} and \underline{underlined} denote the best and second-best results. Performances in $(\cdot)$ show changes relative to Baseline. * mark means the use of Earth-Adapter for PEFT. Parameters in $(\cdot)$ mean the activated parameters, representing the efficient inference cost of CrossEarth-SAR.}
\label{tab:1gap}
\small  
\renewcommand{\arraystretch}{1.1} 
\setlength{\tabcolsep}{4pt}  
\resizebox{1.0\textwidth}{!}{
\begin{tabular}{cccccccccccccccc}
\toprule
\multirow{2}{*}{Method} & \multirow{2}{*}{Backbone}&\multirow{2}{*}{Param.}  & \multicolumn{4}{c}{Unseen Region} & \multicolumn{4}{c}{Unseen Polarization} & \multicolumn{4}{c}{Unseen Complex Value} & \multirow{2}{*}{Avg.} \\
\cmidrule{4-15}
 & & &N2S & S2N & K2C & C2K & VV2F & F2VV & HH2F & F2HH & C(r)2R & R2C(r) & C(i)2R & R2C(i) &  \\
\midrule
S12-MoCo \cite{stewart2023ssl4eo} & ViT-S \cite{dosovitskiy2020image} &20M & 23.2 & 24.0 & 8.9 & 18.6 & 21.9 & 16.1 & 19.8 & 29.9 & 29.9 & 48.1 & 29.3 & 37.7 & 25.6 \\
S12-DINO \cite{stewart2023ssl4eo} & ViT-S \cite{dosovitskiy2020image} & 20M&22.5 & 27.0 & 12.6 & 30.8 & 23.0 & 37.5 & 15.7 & 35.1 & 25.0 & 59.0 & 23.4 & 55.0 & 30.5 \\
S12-MAE \cite{stewart2023ssl4eo} & ViT-S \cite{dosovitskiy2020image} &20M& 22.9 & 29.9 & 5.9 & 34.8 & 12.8 & 17.7 & 12.6 & 15.6 & 23.7 & 33.3 & 19.5 & 22.8 & 21.0 \\
DOFA \cite{xiong2024neural} & ViT-B \cite{dosovitskiy2020image} & 80M& 25.3 & 27.5 & 15.8 & 31.9 & 20.4 & 30.7 & 21.6 & 30.6 & 24.6 & 42.6 & 24.7 & 26.3 & 26.8 \\
\noalign{\vskip 1pt}  
\hdashline  
\noalign{\vskip 1pt}  
SatMAE \cite{cong2022satmae} & ViT-L \cite{dosovitskiy2020image} &300M& 26.2 & 32.1 & 14.5 & 33.4 & 22.5 & 33.4 & 17.1 & 30.0 & 46.4 & 34.4 & 43.4 & 35.6 & 30.7 \\
ScaleMAE \cite{reed2023scale} & ViT-L \cite{dosovitskiy2020image} &300M& 26.4 & 27.2 & 12.9 & 30.5 & 15.1 & 33.3 & 14.9 & 20.2 & 36.8 & 40.6 & 50.0 & 30.6 & 28.2 \\
RemoteCLIP \cite{liu2024remoteclip} & ViT-L \cite{dosovitskiy2020image} & 300M& 22.4 & 30.2 & 5.9 & 29.8 & 13.3 & 20.9 & 13.0 & 17.3 & 22.3 & 24.9 & 20.6 & 31.0 & 21.0 \\
MTP \cite{wang2024mtp} & ViT-L \cite{dosovitskiy2020image} & 300M& 30.6 & 35.3 & 25.7 & 33.9 & 30.4 & 40.2 & 36.0 & 28.9 & 70.8 & 65.4 & 71.4 & 67.6 & 44.7 \\
DINOv2 (Baseline) \cite{oquab2023dinov2} & ViT-L \cite{dosovitskiy2020image} &300M& 32.3 & 43.8 & 34.0 & 35.8 & 65.7 & 63.2 & 56.8 & 55.2 & 71.3 & 68.8 & 71.7 & 66.8 & 55.5 \\
DINOv3 \cite{simeoni2025dinov3} & ViT-L \cite{dosovitskiy2020image} & 300M& 33.7 & 42.8 & 29.9 & 35.3 & 48.3 & 56.6 & 50.6 & 60.4 & 69.9 & 71.9 & 69.2 & 66.8 & 53.0 \\
\hdashline  
\noalign{\vskip 2pt}  
SARATR-X \cite{hoyer2022hrda} & HiViT-B \cite{zhang2023hivit} & 60M& 28.4 & 34.8 & 5.9 & 12.1 & 57.1 & 56.5 & 48.6 & 56.2 & 73.8 & 73.4 & 72.3 & 72.4 & 49.3 \\
\noalign{\vskip 1pt}  
\hdashline  
\noalign{\vskip 1pt}
\rowcolor[RGB]{235,235,235}
& & 90M & 
34.6 & 43.2 & 35.4 & 35.6 & 71.3 & 68.0 & 68.5 & 65.0 & 74.5 & 73.6 & 74.2 & 72.9 & 59.7 \\
\rowcolor[RGB]{235,235,235}
\multirow{-2}{*}{\textbf{CrossEarth-SAR-S}} &
\multirow{-2}{*}{ViT-S \cite{dosovitskiy2020image}} &
(20M) &
\textbf{(\textcolor{kcgreen}{+2.3})} & (\textcolor{kcred}{-0.6}) & \textbf{(\textcolor{kcgreen}{+1.4})} &
(\textcolor{kcred}{-0.2}) & \textbf{(\textcolor{kcgreen}{+5.6})} & \textbf{(\textcolor{kcgreen}{+4.8})} &
\textbf{(\textcolor{kcgreen}{+11.7})} & \textbf{(\textcolor{kcgreen}{+9.8})} &
\textbf{(\textcolor{kcgreen}{+3.2})} & \textbf{(\textcolor{kcgreen}{+4.8})} &
\textbf{(\textcolor{kcgreen}{+2.5})} & \textbf{(\textcolor{kcgreen}{+6.1})} &
\textbf{(\textcolor{kcgreen}{+4.2})} \\

\noalign{\vskip 1pt}
\hdashline
\noalign{\vskip 1pt}

\rowcolor[RGB]{235,235,235}
& & 300M &
36.7 & 43.4 & 37.7 & 37.1 & 73.5 & 68.2 & 70.6 & 66.5 & 76.1 & \underline{74.5} & 75.1 & 73.5 & 61.1 \\
\rowcolor[RGB]{235,235,235}
\multirow{-2}{*}{\textbf{CrossEarth-SAR-B}} &
\multirow{-2}{*}{ViT-B \cite{dosovitskiy2020image}} &
(80M) &
\textbf{(\textcolor{kcgreen}{+4.4})} & (\textcolor{kcred}{-0.4}) &
\textbf{(\textcolor{kcgreen}{+3.7})} & \textbf{(\textcolor{kcgreen}{+1.3})} &
\textbf{(\textcolor{kcgreen}{+7.8})} & \textbf{(\textcolor{kcgreen}{+5.0})} &
\textbf{(\textcolor{kcgreen}{+13.8})} & \textbf{(\textcolor{kcgreen}{+11.3})} &
\textbf{(\textcolor{kcgreen}{+4.8})} & \textbf{(\textcolor{kcgreen}{+5.7})} &
\textbf{(\textcolor{kcgreen}{+3.4})} & \textbf{(\textcolor{kcgreen}{+6.7})} &
\textbf{(\textcolor{kcgreen}{+5.6})} \\

\noalign{\vskip 1pt}
\hdashline
\noalign{\vskip 1pt}

\rowcolor[RGB]{235,235,235}
& & 1.3B &
\underline{37.8} & \underline{45.6} & \underline{38.1} & \underline{38.4} &
\underline{73.8} & \underline{69.8} & \textbf{72.3} & \underline{67.1} &
\underline{76.4} & 73.6 & \underline{76.4} & \underline{73.6} & \underline{61.9} \\
\rowcolor[RGB]{235,235,235}
\multirow{-2}{*}{\textbf{CrossEarth-SAR-L}} &
\multirow{-2}{*}{ViT-L \cite{dosovitskiy2020image}} &
(300M)  &
\textbf{(\textcolor{kcgreen}{+5.5})} & \textbf{(\textcolor{kcgreen}{+1.8})} &
\textbf{(\textcolor{kcgreen}{+4.1})} & \textbf{(\textcolor{kcgreen}{+2.6})} &
\textbf{(\textcolor{kcgreen}{+8.1})} & \textbf{(\textcolor{kcgreen}{+6.6})} &
\textbf{(\textcolor{kcgreen}{+15.5})} & \textbf{(\textcolor{kcgreen}{+11.9})} &
\textbf{(\textcolor{kcgreen}{+5.1})} & \textbf{(\textcolor{kcgreen}{+4.8})} &
\textbf{(\textcolor{kcgreen}{+4.7})} & \textbf{(\textcolor{kcgreen}{+6.8})} &
\textbf{(\textcolor{kcgreen}{+6.4})} \\

\noalign{\vskip 1pt}
\hdashline
\noalign{\vskip 1pt}

\rowcolor[RGB]{235,235,235}
& & 1.3B &
\textbf{38.0} & \textbf{46.7} & \textbf{39.1} & \textbf{38.6} &
\textbf{73.9} & \textbf{72.2} & \underline{71.8} & \textbf{68.6} &
\textbf{76.9} & \textbf{75.2} & \textbf{76.7} & \textbf{75.1} & \textbf{62.7} \\
\rowcolor[RGB]{235,235,235}
\multirow{-2}{*}{\textbf{CrossEarth-SAR-L$^*$}} &
\multirow{-2}{*}{ViT-L \cite{dosovitskiy2020image}} &
(300M) &
\textbf{(\textcolor{kcgreen}{+5.7})} & \textbf{(\textcolor{kcgreen}{+2.9})} &
\textbf{(\textcolor{kcgreen}{+5.1})} & \textbf{(\textcolor{kcgreen}{+2.8})} &
\textbf{(\textcolor{kcgreen}{+8.2})} & \textbf{(\textcolor{kcgreen}{+9.0})} &
\textbf{(\textcolor{kcgreen}{+15.0})} & \textbf{(\textcolor{kcgreen}{+13.4})} &
\textbf{(\textcolor{kcgreen}{+5.6})} & \textbf{(\textcolor{kcgreen}{+6.4})} &
\textbf{(\textcolor{kcgreen}{+5.0})} & \textbf{(\textcolor{kcgreen}{+8.3})} &
\textbf{(\textcolor{kcgreen}{+7.2})} \\


\noalign{\vskip 1pt}  
\bottomrule
\end{tabular}
}
\end{table*}
\begin{table*}[t]
\centering
\caption{Performance comparison on benchmarks across two and three domain gaps. Results show that CrossEarth-SAR-S outperforms other larger FMs. CrossEarth-SAR-L$^*$ performance also demonstrates further improvement potential in PEFT of CrossEarth-SAR.}
\label{tab:23gap}
\small  
\renewcommand{\arraystretch}{1} 
\setlength{\tabcolsep}{4pt}  
\resizebox{1.0\textwidth}{!}{
\begin{tabular}{cccccccccccccc}
\toprule
\multirow{2}{*}{Method} & \multirow{2}{*}{Backbone} &\multirow{2}{*}{Param.} & \multicolumn{2}{c}{\makecell{Unseen Region and\\ Polarization}} & \multicolumn{2}{c}{\makecell{Unseen Region and\\ Platform}} & \multicolumn{2}{c}{\makecell{Unseen Region and\\ MW Band}} & \multicolumn{2}{c}{\makecell{Unseen Region and\\ Pol, MW Band}} & \multicolumn{2}{c}{\makecell{Unseen Region and\\ Platform, MW Band}} & \multirow{2}{*}{Avg.} \\
\cmidrule{4-13}
 & & & F2A & A2F & O2D & D2O & S2A & A2S & D2F & F2D & W2D & D2W &  \\
\midrule
S12-MoCo \cite{stewart2023ssl4eo} & ViT-S \cite{dosovitskiy2020image} & 20M & 7.4 & 14.7 & 5.5 & 8.1 & 45.2 & 42.2 & 6.7 & 17.4 & 10.3 & 5.2 & 16.3 \\
S12-DINO \cite{stewart2023ssl4eo} & ViT-S \cite{dosovitskiy2020image} & 20M& 8.9 & 23.6 & 4.3 & 9.5 & 53.9 & 51.1 & 15.5 & 15.9 & 13.6 & 10.1 & 20.6 \\
S12-MAE \cite{stewart2023ssl4eo} & ViT-S \cite{dosovitskiy2020image}& 20M & 12.1 & 11.4 & 5.3 & 8.8 & 31.3 & 51.8 & 19.6 & 16.1 & 11.9 & 14.8 & 18.3 \\
DOFA \cite{xiong2024neural} & ViT-B \cite{dosovitskiy2020image} & 80M& 10.3 & 22.3 & 4.8 & 10.2 & 50.9 & 27.5 & 8.3 & 16.0 & 12.3 & 12.5 & 17.5 \\
\noalign{\vskip 1pt}  
\hdashline  
\noalign{\vskip 1pt}  
SatMAE \cite{cong2022satmae} & ViT-L \cite{dosovitskiy2020image} & 300M& 11.8 & 13.1 & 12.1 & 9.9 & 49.2 & 48.1 & 20.3 & 17.5 & 10.2 & 7.7 & 20.0 \\
ScaleMAE \cite{reed2023scale} & ViT-L \cite{dosovitskiy2020image} & 300M& 2.4 & 8.6 & 9.9 & 8.2 & 42.9 & 47.7 & 18.6 & 16.3 & 7.4 & 5.3 & 16.7 \\
RemoteCLIP \cite{liu2024remoteclip} & ViT-L \cite{dosovitskiy2020image} & 300M& 12.6 & 21.6 & 6.4 & \underline{10.7} & 16.4 & 2.7 & 21.7 & 15.5 & 8.9 & 11.0 & 12.8 \\
MTP \cite{wang2024mtp} & ViT-L \cite{dosovitskiy2020image} & 300M& 2.6 & 22.0 & 3.9 & 4.1 & 8.5 & 50.5 & 7.1 & 8.0 & 12.5 & 11.4 & 13.1 \\
DINOv2 (Baseline) \cite{oquab2023dinov2} & ViT-L \cite{dosovitskiy2020image} & 300M& 13.4 & 15.5 & 17.8 & 10.1 & 55.9 & \underline{52.4} & \underline{26.0} & 21.4 & 16.7 & 13.8 & 24.3 \\
DINOv3 \cite{simeoni2025dinov3} & ViT-L \cite{dosovitskiy2020image} & 300M& 15.7 & 23.9 & 15.3 & 8.8 & 42.2 & 42.1 & 22.1 & 19.5 & 13.7 & 13.4 & 21.7 \\%
\hdashline  
\noalign{\vskip 2pt}  
SARATR-X \cite{hoyer2022hrda} & HiViT-B \cite{zhang2023hivit}& 60M & 13.0 & 20.1 & 5.4 & 4.4 & 12.7 & 50.3 & 21.7 & 15.0 & 12.2 & 2.7 & 15.8 \\
\noalign{\vskip 1pt}  
\hdashline  
\noalign{\vskip 1pt}  
\rowcolor[RGB]{235,235,235}
& & 90M &
14.1 & 21.3 & 19.0 & \underline{10.7} & 53.1 & 50.5 & 22.6 & 21.7 & 16.1 & \underline{18.9} & 24.8 \\
\rowcolor[RGB]{235,235,235}
\multirow{-2}{*}{\textbf{CrossEarth-SAR-S}} &
\multirow{-2}{*}{ViT-S \cite{dosovitskiy2020image}} &
(20M) &
\textbf{(\textcolor{kcgreen}{+0.7})} & \textbf{(\textcolor{kcgreen}{+5.8})} &
\textbf{(\textcolor{kcgreen}{+1.2})} & \textbf{(\textcolor{kcgreen}{+0.6})} &
(\textcolor{kcred}{-2.8}) & (\textcolor{kcred}{-1.9}) &
(\textcolor{kcred}{-3.4}) & \textbf{(\textcolor{kcgreen}{+0.3})} &
(\textcolor{kcred}{-0.6}) & \textbf{(\textcolor{kcgreen}{+5.1})} &
\textbf{(\textcolor{kcgreen}{+0.5})} \\

\noalign{\vskip 1pt}\hdashline\noalign{\vskip 1pt}

\rowcolor[RGB]{235,235,235}
& & 300M &
15.2 & 24.2 & 20.0 & 9.4 & 57.1 & 48.4 & 21.2 & 23.2 & 20.0 & 15.3 & 25.4 \\
\rowcolor[RGB]{235,235,235}
\multirow{-2}{*}{\textbf{CrossEarth-SAR-B}} &
\multirow{-2}{*}{ViT-B \cite{dosovitskiy2020image}} &
(80M) &
\textbf{(\textcolor{kcgreen}{+1.8})} & \textbf{(\textcolor{kcgreen}{+8.7})} &
\textbf{(\textcolor{kcgreen}{+2.2})} & (\textcolor{kcred}{-0.7}) &
\textbf{(\textcolor{kcgreen}{+1.2})} & (\textcolor{kcred}{-4.0}) &
(\textcolor{kcred}{-4.8}) & \textbf{(\textcolor{kcgreen}{+1.8})} &
\textbf{(\textcolor{kcgreen}{+3.3})} & \textbf{(\textcolor{kcgreen}{+1.5})} &
\textbf{(\textcolor{kcgreen}{+1.1})} \\

\noalign{\vskip 1pt}\hdashline\noalign{\vskip 1pt}

\rowcolor[RGB]{235,235,235}
& & 1.3B &
\textbf{16.9} & \underline{25.0} & \textbf{23.7} & 9.6 & \textbf{59.1} & \underline{52.4} & 25.1 & \textbf{25.0} & \underline{22.2} & 18.0 & \underline{27.7} \\
\rowcolor[RGB]{235,235,235}
\multirow{-2}{*}{\textbf{CrossEarth-SAR-L}} &
\multirow{-2}{*}{ViT-L \cite{dosovitskiy2020image}} &
(300M) &
\textbf{(\textcolor{kcgreen}{+3.5})} & \textbf{(\textcolor{kcgreen}{+9.5})} &
\textbf{(\textcolor{kcgreen}{+5.9})} & (\textcolor{kcred}{-0.5}) &
\textbf{(\textcolor{kcgreen}{+3.2})} & \textbf{(\textcolor{kcgreen}{+0.0})} &
(\textcolor{kcred}{-0.9}) & \textbf{(\textcolor{kcgreen}{+3.6})} &
\textbf{(\textcolor{kcgreen}{+5.5})} & \textbf{(\textcolor{kcgreen}{+4.2})} &
\textbf{(\textcolor{kcgreen}{+3.4})} \\

\noalign{\vskip 1pt}\hdashline\noalign{\vskip 1pt}

\rowcolor[RGB]{235,235,235}
& & 1.3B &
\underline{16.1} & \textbf{27.0} & \underline{23.1} & \textbf{11.3} &
\underline{57.9} & \textbf{53.5} & \textbf{26.5} & \underline{24.1} &
\textbf{25.6} & \textbf{19.4} & \textbf{28.5} \\
\rowcolor[RGB]{235,235,235}
\multirow{-2}{*}{\textbf{CrossEarth-SAR-L$^*$}} &
\multirow{-2}{*}{ViT-L \cite{dosovitskiy2020image}} &
(300M) &
\textbf{(\textcolor{kcgreen}{+2.7})} & \textbf{(\textcolor{kcgreen}{+11.5})} &
\textbf{(\textcolor{kcgreen}{+5.3})} & \textbf{(\textcolor{kcgreen}{+1.2})} &
\textbf{(\textcolor{kcgreen}{+2.0})} & \textbf{(\textcolor{kcgreen}{+1.1})} &
\textbf{(\textcolor{kcgreen}{+0.5})} & \textbf{(\textcolor{kcgreen}{+2.7})} &
\textbf{(\textcolor{kcgreen}{+8.9})} & \textbf{(\textcolor{kcgreen}{+5.6})} &
\textbf{(\textcolor{kcgreen}{+4.2})} \\
\noalign{\vskip 1pt}  
\bottomrule
\end{tabular}
}
\end{table*}

\begin{figure*}[t]
  \centering
  \includegraphics[width=1.0\columnwidth]{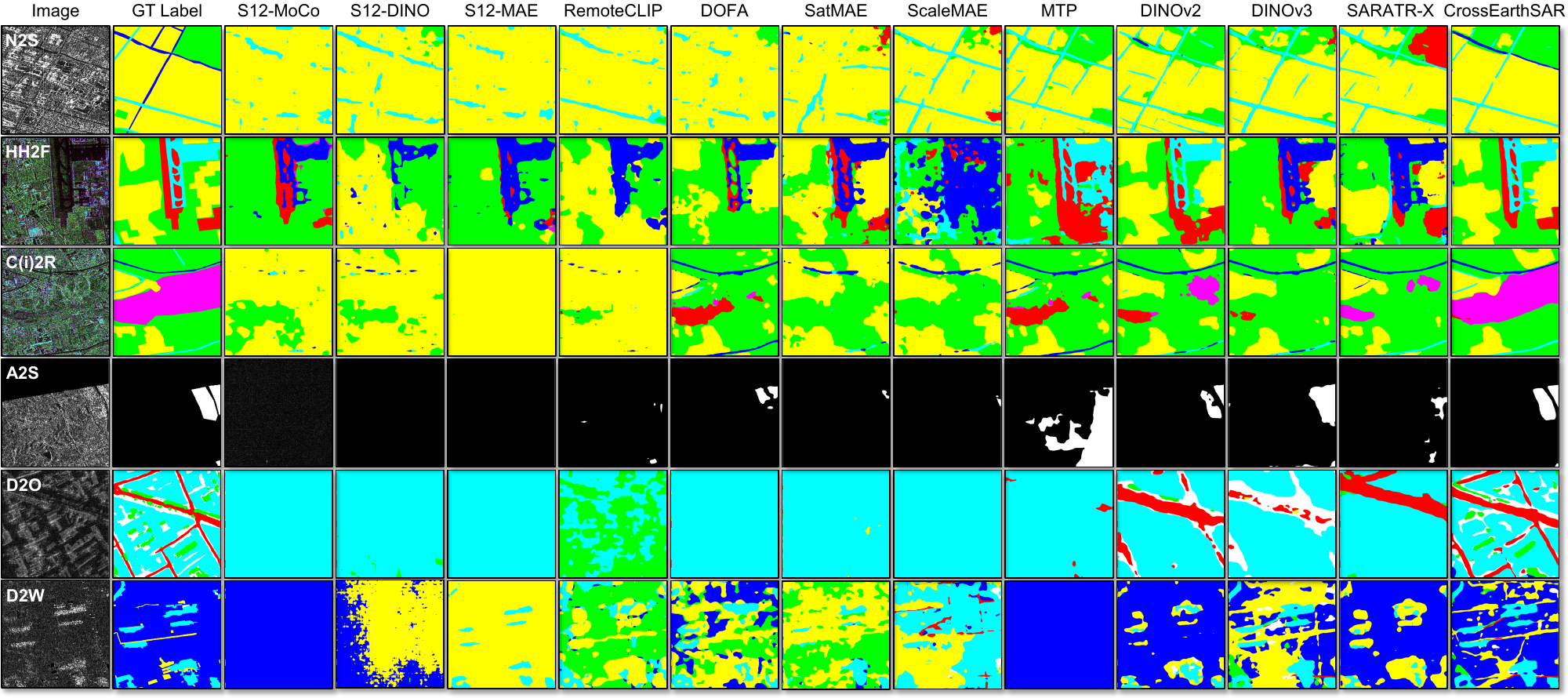}
  \caption{Visualizations of predicted segmentation maps on six representative benchmarks. For the top three rows color map, \textcolor[RGB]{0, 0, 255}{blue} is water, \textcolor[RGB]{0, 255, 0}{green} is vegetation, \textcolor[RGB]{255, 0, 0}{red} is ground, \textcolor[RGB]{0, 255, 255}{cyan} is road, \textcolor[RGB]{245, 245, 0}{yellow} is building, and \textcolor[RGB]{255, 0, 255}{purple} is mountain. For the bottom two rows color map, \textcolor[RGB]{0, 0, 255}{blue} is farmland, \textcolor[RGB]{0, 255, 0}{green} is greenery, \textcolor[RGB]{255, 0, 0}{red} is road, \textcolor[RGB]{0, 255, 255}{cyan} is building, \textcolor[RGB]{245, 245, 0}{yellow} is water, and white is background.}
  \label{fig:visualization}
\end{figure*}
\section{Experiments}
\label{sec:experiments}


\subsection{Preliminary}
\label{subsec:Preliminary}
We perform continuous pre-training (CPT) on DINOv2 using the CrossEarth-SAR-200K, where all backbone and decoder parameters are trainable. CPT runs for 18 epochs with a batch size of 4, optimized by AdamW~\cite{adamw} with a learning rate of 3e-5. For the CPT stage, we use CrossEarth-SAR-200K-Val as the validation set. All experiments during the CPT stage are conducted in PyTorch using 16 NVIDIA A100 (80 GB) GPUs.

After CPT, we fine-tune CrossEarth-SAR on the source domain and evaluate on unseen target domains. During downstream tasks, all models' backbones are frozen and only the decoder is trained for 40k iterations with a batch size of 2 on a single NVIDIA 4090 (48 GB) GPU, using a learning rate of 1e-4.

\subsection{Generalization across One Gap}
\label{subsec:First result}
In this section, the region, polarization, and complex values are the primary gaps between the source and unseen domains, and there is only one gap in each experiment. Results are presented in Tab.~\ref{tab:1gap}. For instance, CrossEarth-SAR-L achieves the best performance on four benchmarks of the common unseen regions in RS. Next, we mainly discuss the unseen polarization and complex-valued gaps, which are unique to SAR imagery.

Different polarization channels interact with terrain through distinct scattering mechanisms, thereby shaping the contrast and class separability in SAR imagery. 
To probe the generalizability of models when cross-polarization, we conduct experiments on unseen polarization across 4 DG benchmarks: VV2F, F2VV, HH2F, and F2HH. 
For single to full polarization benchmarks (VV2F and HH2F), CrossEarth-SAR achieves substantial improvements over the baseline. On VV2F, CrossEarth-SAR-L achieves 73.8\% mIoU (+8.4\%), while on HH2F, CrossEarth-SAR-L reaches 72.3\% mIoU (+15.5\%). For full to single polarization benchmarks (F2VV and F2HH), CrossEarth-SAR-L also shows remarkable improvements with +6.6\% on F2VV and +11.9\% on F2HH. Notably, compared with 300M DINOv2 and v3, 90M CrossEarth-SAR-S also shows its superiority with 11.7\% improvement in HH2F.

In addition to polarization, SAR inherently produces coherent measurements represented as complex-valued signals. However, most public SAR products only provide real-valued intensity images, which limits the investigation of generalization under complex-valued shifts. To address this, we evaluate generalization across complex-to-real and real-to-complex settings in the one-gap scenario.For complex to real benchmarks (C(r)2R and C(i)2R), CrossEarth-SAR significantly outperforms the baseline across all scales. For example, CrossEarth-SAR-L achieves 76.4\% (+5.1\%) on C(r)2R and 73.6\% (+4.8\%) on R2C(r).The similar trends are also shown in benchmarks (C(i)2R and R2C(i)). This indicates that CrossEarth-SAR excels in understanding the complex values of SAR images. Moreover, the improved performance of CrossEarth-SAR-L$^*$ highlights the further potential of PEFT. 

These results demonstrate that CrossEarth-SAR achieves optimal generalization performance across all one-gap DG benchmarks. The qualitative results for the one-gap setting are provided in the first three rows of Fig.~\ref{fig:visualization}.

\subsection{Generalization across Two Gaps}
\label{subsec:Second result}
Experiments in this section primarily demonstrate the results of two DG gaps, which are formed by pairwise combinations of region, polarization, platform, and microwave band. Notably, the D2O and O2D benchmarks assess global generalization capabilities, while the S2A and A2S benchmarks evaluate building extraction capabilities.

In experiments across two gaps of unseen region and polarization shown in Tab.~\ref {tab:23gap}, CrossEarth-SAR achieves the best mIoU on both F2A and A2F benchmarks. Furthermore, CrossEarth-SAR-L delivers performance at 25.0\%, surpassing baseline by 9.5\% and DINOv3 (second-best) by 1.1\% on the A2F benchmark. 
Similar in experiments across region and platform, CrossEarth-SAR-L$^*$ exhibits global generalizability on O2D and D2O benchmarks, surpassing the baseline with 5.3\% and 1.2\% improvements, respectively.
Although the base and large variants of CrossEarth-SAR on D2O do not surpass the baseline (-0.7\% and -0.5\%), they outperform DINOv3 by 0.6\% and 0.8\%, and exceed the SAR foundation model SARATR-X by 5.0\% and 5.2\%. 

In addition to the above, how accurately models perform in the building extraction task also serves as a key indicator of their generalizability. Thus, we conduct experiments on the S2A and A2S benchmarks. Our strongest model, CrossEarth-SAR-L$^*$, achieves improvements of 2.0\% and 1.1\% mIoU over the baseline on S2A and A2S, respectively. These results indicate that even under two distinct domain gaps, the CrossEarth-SAR maintains a clear advantage in building extraction. The qualitative results for the two-gap setting are provided in the fourth and fifth rows of Fig.~\ref{fig:visualization}.

\subsection{Generalization across Three Gaps}
\label{subsec:Third result}

In this section, we thoroughly investigate the three-gap case, which exemplifies one of the most complex challenges in DG scenarios. There are a total of 4 benchmarks shown in Tab.~\ref{tab:23gap}. Excluding the D2F benchmark, CrossEarth-SAR demonstrates remarkable improvements in mIoU across the other three benchmarks: F2D, D2W, and W2D, highlighting its superior performance. 

Specifically, CrossEarth-SAR-L$^*$ outperforms the baseline by 0.5\% and 2.7\% mIoU on the D2F and F2D benchmarks, respectively. Although CrossEarth-SAR-L does not surpass the baseline on the D2F benchmark (-0.9\%), it both outperform DINOv3 by 3.0\% and SARATR-X by 3.4\%. Extensive experiments validate CrossEarth-SAR's exceptional ability to learn domain-invariant representations under diverse and complex SAR domains.

\subsection{Ablation Study}

As mentioned earlier, we use CrossEarth-SAR-200K-Val for the ablation study during the CPT stage, as summarized in Tab.~\ref{tab:ablation_left}. Specifically, the training paradigm, MoE design, and learning rate ablations are performed on CrossEarth-SAR-L, with Top-$k$=1 and Expert Number $n$=6, as this configuration yields the best overall performance. Additionally, experiments, except for data construction, utilize all CrossEarth-SAR-200K data.

\textbf{Pseudo-label quality.}
We first compare training on only fully supervised samples against CrossEarth-SAR-200K, as reported in Tab.~\ref{tab:ablation_left}. Specifically, the latter achieves 59.4\% mIoU, surpassing the only 40K setting by a significant 14.3\%, showing that extra pseudo labels generated by CrossEarth improve the SAR representation learning.
Additionally, the ablation results in Tab.~\ref{tab:ablation_rebuttal} validate the utility of the pseudo labels. Notably, ground-truth annotations are often constrained to limited geographic regions, whereas pseudo-labeled data are collected at a global scale. The results show that using 40K pseudo-labeled samples yields larger improvements on downstream tasks than using 40K real labels. Moreover, Rows 1 and 3 of Tab.~\ref{tab:ablation_rebuttal} show that adding pseudo labels to real-label training yields an average +3.6\% gain on OOD downstream tasks.

\begin{table}[t]
\centering
\hspace*{-2.5em}
\begin{minipage}{\textwidth}
    \centering
    \caption{Ablation Study on the Structure and Data Validity of CrossEarth-SAR.}
    \label{tab:ablation_rebuttal}
\end{minipage}

\vspace{0.0em} 

\resizebox{\textwidth}{!}{%
    \renewcommand{\arraystretch}{1.2}
    \setlength{\tabcolsep}{5pt}
    \begin{tabular}{lccccccccc}
    \toprule
    Method with CPT dataset & N2S & VV2F & C(r)2R & A2F & O2D & S2A & D2F & W2D & avg. \\
    \midrule
    DINOv2 + \textbf{40K} real label datasets & 32.7 &  68.6 &  67.7 & 16.3 & 16.1 & 52.9 & 20.4 & 17.1 & 36.5 \\

    DINOv2 + \textbf{40K} pseudo label datasets (randomly selected) &  33.2 & 66.3 & 66.9 &  18.6 & 17.2 &  53.6 & 21.3 &  17.6 &  36.8 \\

    DINOv2 + CrossEarth-SAR-\textbf{200K} &  36.6 &  70.6 &  74.5 &  23.4 &  19.7 &  55.2 &  21.9 &  18.9 &  40.1 \\

    \textbf{CrossEarth-SAR} &  \textbf{37.8} &  \textbf{73.8} &  \textbf{76.4} &  \textbf{25.0} &  \textbf{23.7} &  \textbf{59.1} &  \textbf{25.1} &  \textbf{22.2} &  \textbf{42.9} \\
    \bottomrule
    \end{tabular}
}
\end{table}
\begin{table*}[t]
\centering
\begin{minipage}[t]{0.48\textwidth}
\vspace{0pt}
        \caption{\textbf{Ablation study}. The \textit{italic font} means the baseline performance of DINOv2-L, and the \textbf{bold font} means the best score of CrossEarth-SAR.}
        \label{tab:ablation_left}
        \vspace{-2mm}
        \centering
        \scriptsize 
        \setlength{\tabcolsep}{1.0pt} 
        \resizebox{\linewidth}{!}{%
        \begingroup
        \setlength{\arrayrulewidth}{0.6pt} 
        \renewcommand{\arraystretch}{1.2} 
        \begin{tabular}{l|>{\centering\arraybackslash}p{1.0cm}>{\centering\arraybackslash}p{1.0cm}>{\centering\arraybackslash}p{1.0cm}>{\centering\arraybackslash}p{1.0cm}|cc}
            \hline
            \multicolumn{1}{c|}{Experiments} & \multicolumn{4}{c|}{Components} & \multicolumn{2}{c}{Performance} \\
            \hline

            \multirow{2}*{\makecell[l]{Data Construction}} & \multicolumn{2}{c}{40$K$ Fully} &\multicolumn{2}{c|}{160$K$ Weakly}  & mIoU & Gain \\ \cline{2-7}
            \multirow{2}*{\makecell[l]{(DINOv2)}} & \multicolumn{2}{c}{\ding{51}} & \multicolumn{2}{c|}{\quad}  & 45.1 & -14.3  \\
            &\multicolumn{2}{c}{\ding{51}} & \multicolumn{2}{c|}{\ding{51}} & \textit{59.4} & \textit{+0.0} \\    \hline

            \multirow{3}*{\makecell[l]{Training Paradigm}} & \multicolumn{2}{c}{CPT} & \multicolumn{2}{c|}{PEFT}  & mIoU & Gain \\ \cline{2-7}
            & \multicolumn{2}{c}{\ding{51}} & \multicolumn{2}{c|}{\quad} & 62.1 & +2.7  \\
            & \multicolumn{2}{c}{\ding{51}} & \multicolumn{2}{c|}{\ding{51}} & \textbf{62.4} &\textbf{+3.0} \\ \hline

            \multirow{5}*{Learning Rate} & $1e{-}5$ & $3e{-}5$ & $6e{-}5$ & $1e{-}4$ & mIoU & Gain \\ \cline{2-7}
            &\ding{51} & \quad & \quad & \quad & 61.2 & +1.8  \\
            & \quad & \ding{51} & \quad & \quad & \textbf{62.4} & \textbf{+3.0} \\
            \multirow{2}*{}& \quad & \quad & \ding{51} & \quad & 62.3 & +2.9 \\
            & \quad & \quad & \quad & \ding{51} & 61.9 & +2.5 \\ \hline
            
            \multirow{5}*{\makecell[l]{MoE Design}} & \multicolumn{2}{c}{$\mathcal{L}_{BC}$} & \multicolumn{2}{c|}{$S$} & mIoU & Gain \\ \cline{2-7}
            & & & & & 61.1 & +1.7\\
            &\multicolumn{2}{c}{\ding{51}} & \multicolumn{2}{c|}{\quad}  & 62.2 & +2.8 \\
            &\multicolumn{2}{c}{\quad} & \multicolumn{2}{c|}{\ding{51}}  & 61.6 & +2.2 \\
            &\multicolumn{2}{c}{\ding{51}} & \multicolumn{2}{c|}{\ding{51}}  & \textbf{62.4} & \textbf{+3.0} \\ \hline
            
            \multirow{4}*{Expert Number n} & $3$ & $4$ & $5$ & $6$ & mIoU & Gain \\ \cline{2-7}
            &\ding{51} & \quad & \quad & \quad & 60.9 & +1.5  \\
            \multirow{2}*{(Top-$k$=1)}& \quad & \ding{51} & \quad & \quad & 61.6 & +2.2 \\
            \multirow{2}*{}& \quad & \quad & \ding{51} & \quad & 61.9 & +2.5 \\
            & \quad & \quad & \quad & \ding{51} &\textbf{62.4} &\textbf{+3.0} \\ \hline
            
            \multirow{3}*{Activate Top-$k$} & \multicolumn{4}{c|}{$k=1$ \qquad $k=2$ \qquad $k=3$} & mIoU & Gain\\ \cline{2-7}
            \multirow{3}*{($n$=6)} &\multicolumn{4}{c|}{\;\;\ding{51}\qquad\qquad\qquad\qquad\qquad\quad}  & \textbf{62.4} & \textbf{+3.0} \\
            &   \multicolumn{4}{c|}{\qquad \; \ding{51} \qquad \qquad}  & 61.7 & +2.3 \\
            &  \multicolumn{4}{c|}{\qquad\qquad\qquad\qquad\quad\;\; \ding{51}}  & 61.3 & +0.9 \\ 
            \hline
        \end{tabular}
        \endgroup
        }
        \vspace{0.0mm}
\end{minipage}%
\hfill
\begin{minipage}[t]{0.48\textwidth}
        \vspace{2pt} 
        \caption{Experiments on the sensitivity of individual physical descriptors to downstream gaps.}
        \label{tab:ablation_right}
        \vspace{0mm}
        \centering
        \scriptsize 
        \resizebox{\linewidth}{!}{
        \renewcommand{\arraystretch}{1.4}
        \begin{tabular}{c*{5}{>{\centering\arraybackslash}m{1.3cm}}}
        \toprule
        Domain Gap & Region & Polarization & Complex Value & Platform & Microwave Band \\
        \midrule
        Test Bench & N2S & VV2F & Cr2R & O2D & S2A \\
        \midrule
        \textit{$H_{DE}$} & 37.45 & \textbf{73.47} & 75.75 & 19.70 & \textbf{59.18} \\
        ENL & 37.48 & 73.33 & \textbf{75.97} & 19.68 & 59.02 \\
        \textit{$R_{LR}$} & \textbf{37.49} & 73.43 & 75.91 & \textbf{19.83} & 59.06 \\
        \bottomrule
        \end{tabular}
        }

        \vspace{7mm} 
        \centering
        \includegraphics[width=\linewidth]{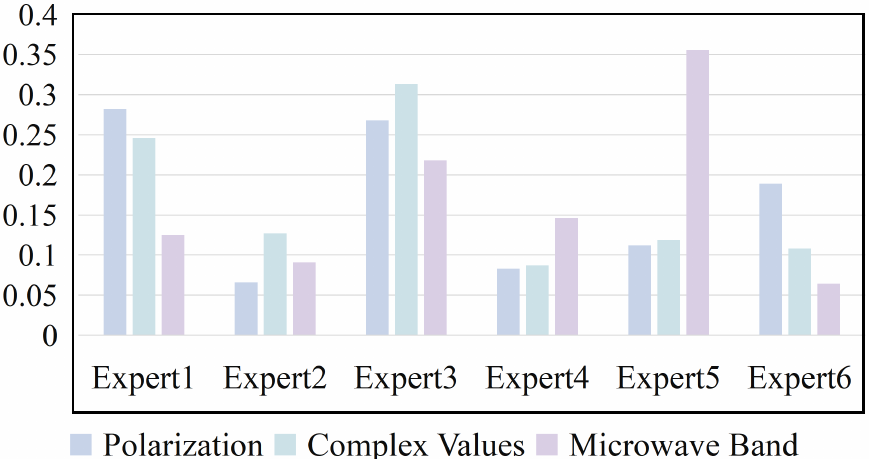} 
        \vspace{0mm} 
        \captionof{figure}{Relationship Between Expert Activations and the SAR Physical Domain.}
        \label{fig:downstream_viz}
        
    \end{minipage}
    \vspace{0mm} 
\end{table*}

\textbf{Train Paradigm.}
We evaluate two training paradigms as in Tab.~\ref{tab:ablation_left}, CPT and PEFT, and observe consistent gains: CPT reaches 62.1\% mIoU (+2.7\%), and CPT+PEFT further improves to 62.4\% mIoU (+3.0\%), in line with the downstream results in Tab.~\ref{tab:1gap} and~\ref{tab:23gap}. Besides, the learning rate is also important, though it does not show strong sensitivity.

\textbf{MoE Desgin.}
For the MoE design in Tab.~\ref{tab:ablation_left}, we start from a plain MoE setup without any additional constraints or priors, which improves performance from DINOv2 to 61.1 mIoU (+1.7\%). Introducing $\mathcal{L}_{BC}$ further enhances expert utilization and raises the result to 62.2\% mIoU (+2.8\%). Incorporating physics descriptors $S$ also brings improvement to 61.6\% mIoU (+2.2\%), and combining both components achieves the best performance of 62.4\% mIoU (+3.0\%).

\textbf{Experts Activation.}
We ablate the expert number $n$ and Top-$k$ activation. As shown in Tab.~\ref{tab:ablation_left}, with Top-$k$=1, increasing $n$ from 3 to 6 improves mIoU from 60.9\% to 62.4\%, suggesting that more experts better capture heterogeneous SAR. However, with $n$=6, raising Top-$k$ to 2 or 3 reduces mIoU (62.4\% $\rightarrow$ 61.3\%), likely because 200K samples favor single-expert specialization and multi-expert activation dilutes it. Therefore, $n$=6 and $k$=1 offer the best trade-off between performance and parameters given the data scale.

\textbf{Physical Descriptors.}
We enable only one physical descriptor at a time across five domain gaps, evaluating each gap on a representative benchmark. The results on mIoU in Tab.~\ref{tab:ablation_right} show that $H_{DE}$ is more sensitive to Polarization (73.47\%) and Microwave Band (59.18\%), ENL to Complex Value (75.97\%), and $R_{\text{LR}}$ to Region (37.49\%) and Platform (19.83\%). Additionally, the results show that all 3 physical descriptors are important, and combining them yields better performance, as reported in Tab.~\ref{tab:1gap} and Tab.~\ref{tab:23gap}.

\textbf{Sensitivity of experts to the SAR domain.}
We randomly selected one benchmark for each of three representative domain gaps to visualise the activation proportions of each expert and the SAR domain layer by layer. As shown in the Fig.~\ref{fig:downstream_viz}, polarization is dominated by Expert 1, complex values by Expert 3, and the microwave band by Expert 5.

\begin{figure}[t]
\centering
\begin{minipage}[t]{0.48\textwidth}
  \centering
  \includegraphics[width=\linewidth]{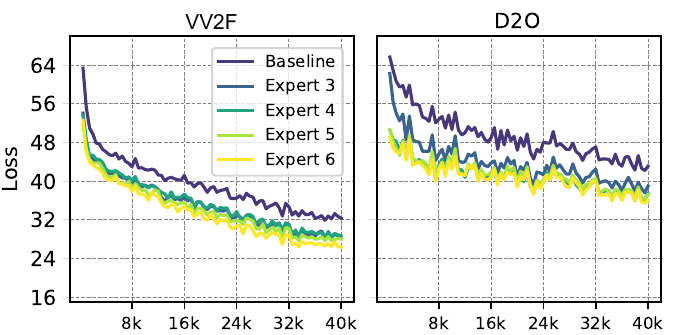}
  \caption{Loss Curve of CrossEarth-SAR in two downstream tasks fine-tuning. CrossEarth-SAR consistently starts and converges at lower losses than DINOv2, illustrating the effectiveness of continuous pre-training across diverse SAR domains.}
  \label{fig:loss}
\end{minipage}
\hfill
\begin{minipage}[t]{0.48\textwidth}
  \centering
  \includegraphics[width=\linewidth]{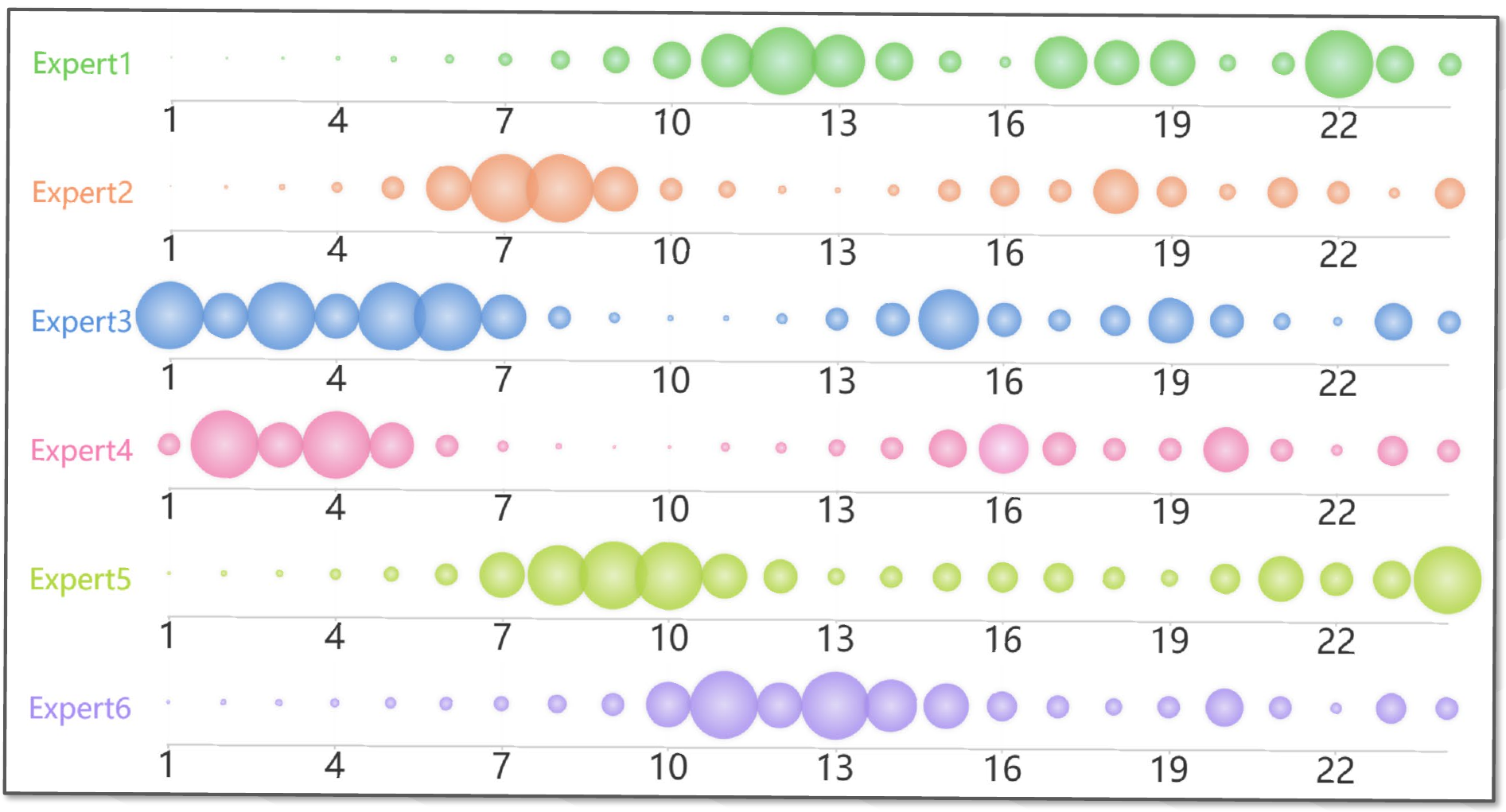}
  \caption{Ratio of expert activations over all benchmarks.
For each of the 24 layers, we count how frequently each expert is activated, and larger circles denote higher activation ratios.}
  \label{fig:bubble}
\end{minipage}
\end{figure}

\subsection{Visualization}

In this section, we primarily present visualizations of loss and expert activation ratio. Additional visualization results are provided in the Appendix.

\textbf{Loss Curve in Downstream Tasks.} We visualize the loss curves of CrossEarth-SAR-L during downstream fine-tuning. As shown in Fig.~\ref{fig:loss}, CrossEarth-SAR consistently starts and converges at lower losses than DINOv2. In the D2O task, for instance, the initial loss drops from over 64.0 to around 50.0, and the final loss remains lower as training progresses. Similar patterns appear in VV2F, demonstrating the effectiveness of continuous pre-training and its improved scalability in downstream optimization. More visualizations are attached in the Appendix.

\textbf{Expert Activation Ratio.} As shown in Fig.~\ref{fig:bubble}, the experts exhibit a clear layer-wise specialization. Experts 3 and 4 dominate the early layers (1–6), indicating their focus on low-level SAR cues such as speckle statistics. In the mid-layers (7–16), Experts 1, 2, 5, and 6 show the highest activation, reflecting their role in modeling geometric structures, textures, and scattering patterns. In the deepest layers (17–24), activation concentrates on Experts 1 and 5, suggesting that they are responsible for high-level semantic learning. These observations show that CrossEarth-SAR learns an effective hierarchical specialization across layers.
\section{Conclusion}
\label{sec:conclusion}

In this paper, we present \textbf{CrossEarth-SAR}, the first \textbf{billion-scale} SAR VFM featuring a physics-guided sparse MoE architecture, designed for cross-domain semantic segmentation.
Additionally, we construct \textbf{CrossEarth-SAR-200K} dataset, which contains over 200K SAR weakly and fully labeled semantic segmentation data, enabling continual pre-training at a global scale for cross-domain semantic understanding.
Furthermore, to encourage SAR RSDG research, we also curate a DG benchmark collection with 22 sub-benchmarks, providing rigorous and comprehensive evaluations of models' generalizability. 
Experiments validate the superiority and potential of CrossEarth-SAR across regions, polarization, complex values, microwave band, and platform.
We hope this work can promote research on cross-domain generalization within the SAR community, and we will continue to explore its DG capabilities in other downstream tasks, such as target recognition and change detection.

\bibliographystyle{plainnat}
\bibliography{CrossEatrhSAR}

@String(CVPR= {IEEE Conf. Comput. Vis. Pattern Recog.})

@String(ICCV= {Int. Conf. Comput. Vis.})

@String(ECCV= {Eur. Conf. Comput. Vis.})

@String(BMVC= {Brit. Mach. Vis. Conf.})

@String(AAAI = {AAAI})

@String(CVPR  = {CVPR})

@String(ICCV  = {ICCV})

@String(ECCV  = {ECCV})

@String(BMVC  =	{BMVC})

@inproceedings{Zhu2017CycleGAN,
title     = {Unpaired Image-to-Image Translation Using Cycle-Consistent Adversarial Networks},
author    = {Zhu, Jun-Yan and Park, Taesung and Isola, Phillip and Efros, Alexei A.},
booktitle = {ICCV},
year      = {2017},
pages     = {2242--2251},
doi       = {10.1109/ICCV.2017.244}
}

@inproceedings{Zou2018ECCVCBST,
title     = {Unsupervised Domain Adaptation for Semantic Segmentation via Class-Balanced Self-Training},
author    = {Zou, Yang and Yu, Zhiding and Kumar, B.V.K. Vijaya and Wang, Jinsong},
booktitle = {ECCV},
year      = {2018},
pages     = {289--305},
doi       = {10.1007/978-3-030-01219-9_18}
}

@inproceedings{Zou2019ICCVCRST,
title     = {Confidence Regularized Self-Training},
author    = {Zou, Yang and Yu, Zhiding and Liu, Xinkun and Kumar, B.V.K. Vijaya and Wang, Jinsong},
booktitle = {ICCV},
year      = {2019},
pages     = {5982--5991},
doi       = {10.1109/ICCV.2019.00608}
}

@inproceedings{Chen2019BMVCUDAST,
title     = {Domain Adaptive Semantic Segmentation with Self-Training and Uncertainty Guidance},
author    = {Chen, Yunlong and Li, Wen and Sakaridis, Christos and Dai, Dengxin and Van Gool, Luc},
booktitle = {BMVC},
year      = {2019}
}

@inproceedings{Saha2020IGARSSUDA,
title     = {Unsupervised Domain Adaptation for Semantic Segmentation of Satellite Images},
author    = {Saha, Sudipan and Krauß, Thomas and Reinartz, Peter},
booktitle = {IGARSS},
year      = {2020},
pages     = {5171--5174},
doi       = {10.1109/IGARSS39084.2020.9323954}
}

@inproceedings{Nguyen2021IGARSSDG,
title     = {Domain Generalization for Land Cover Mapping with Meta-Learning},
author    = {Nguyen, Van-Hoai and Drummond, Tom and Petersson, Lars and Harandi, Mehrtash},
booktitle = {IGARSS},
year      = {2021},
pages     = {1156--1159},
doi       = {10.1109/IGARSS47720.2021.9554748}
}

@article{Benjdira2019RS,
title   = {Car Detection Using Unmanned Aerial Vehicles: Image Enhancement and Domain Adaptation},
author  = {Benjdira, Bilel and Bazi, Yakoub and Koubaa, Anis and Ammar, Ayoub},
journal = {Remote Sensing},
year    = {2019},
volume  = {11},
number  = {7},
pages   = {747},
doi     = {10.3390/rs11070747}
}

@article{Li2020RemoteSensingCrossCity,
title   = {Cross-City Domain Adaptation for Semantic Segmentation of Aerial Images},
author  = {Li, Yansheng and Liu, Sijie and Zhou, Wei and et al.},
journal = {Remote Sensing},
year    = {2020},
volume  = {12},
number  = {19},
pages   = {3132},
doi     = {10.3390/rs12193132}
}

@article{Marmanis2018ISPRS,
title   = {Classification with an Edge: Improving Semantic Segmentation of Aerial Images Using Boundary Detection},
author  = {Marmanis, Dimitrios and Datcu, Mihai and Esch, Thomas and Stilla, Uwe},
journal = {ISPRS Journal of Photogrammetry and Remote Sensing},
year    = {2018},
volume  = {135},
pages   = {158--172},
doi     = {10.1016/j.isprsjprs.2017.11.009}
}

@article{Reichstein2019Nature,
title   = {Deep Learning and Process Understanding for Data-Driven Earth System Science},
author  = {Reichstein, Markus and Camps-Valls, Gustau and Stevens, Bjorn and et al.},
journal = {Nature},
year    = {2019},
volume  = {566},
pages   = {195--204},
doi     = {10.1038/s41586-019-0912-1}
}

@article{Zhang2023PerSAM,
title   = {Personalize Segment Anything Model for Remote Sensing Zero/Few-Shot Segmentation},
author  = {Zhang, Runpei and Li, Zhen and Wang, Qiang and et al.},
journal = {arXiv preprint arXiv:2306.05457},
year    = {2023}
}

@inproceedings{ye2024convolutional,
  title={Convolutional modulated scattering feature network for aircraft classification in SAR images},
  author={Ye, Ziqi and Xiao, Xiayang and Wang, Haipeng},
  booktitle={IGARSS 2024-2024 IEEE International Geoscience and Remote Sensing Symposium},
  pages={9329--9332},
  year={2024},
  organization={IEEE}
}

@article{Zhang2024HyperSIGMA,
title   = {HyperSIGMA: A Foundation Model for Hyperspectral and Multispectral Remote Sensing},
author  = {Zhang, X. and Wang, Y. and Li, C. and et al.},
journal = {arXiv preprint arXiv:2405.12667},
year    = {2024}
}

@inproceedings{Schmitt2019SEN12MS,
title     = {SEN12MS -- A Curated Dataset of Georeferenced Multi-Spectral Sentinel-1/2 Imagery for Deep Learning and Data Fusion},
author    = {Schmitt, Michael and Hughes, Lloyd H. and Qiu, Chenying and Zhu, Xiao Xiang},
booktitle = {ISPRS Annals},
year      = {2019},
volume    = {IV-2/W7},
pages     = {153--160},
doi       = {10.5194/isprs-annals-IV-2-W7-153-2019}
}

@article{Wang2022SARTransformerSeg,
title   = {Semantic Segmentation of SAR Images with Transformer-Based Networks},
author  = {Wang, Y. and Liu, H. and Zhang, J.},
journal = {Remote Sensing},
year    = {2022},
volume  = {14},
number  = {3},
pages   = {512},
doi     = {10.3390/rs14030512}
}

@article{gong2024crossearth,
  title={Crossearth: Geospatial vision foundation model for domain generalizable remote sensing semantic segmentation},
  author={Gong, Ziyang and Wei, Zhixiang and Wang, Di and Ma, Xianzheng and Chen, Hongruixuan and Jia, Yuru and Deng, Yupeng and Ji, Zhenming and Zhu, Xiangwei and Yokoya, Naoto and others},
  journal={arXiv preprint arXiv:2410.22629},
  year={2024}
}

@InProceedings{guo2024skysense,
    author    = {Guo, Xin and Lao, Jiangwei and Dang, Bo and Zhang, Yingying and Yu, Lei and Ru, Lixiang and Zhong, Liheng and Huang, Ziyuan and Wu, Kang and Hu, Dingxiang and He, Huimei and Wang, Jian and Chen, Jingdong and Yang, Ming and Zhang, Yongjun and Li, Yansheng},
    title     = {SkySense: A Multi-Modal Remote Sensing Foundation Model Towards Universal Interpretation for Earth Observation Imagery},
    booktitle = {Proceedings of the IEEE/CVF Conference on Computer Vision and Pattern Recognition (CVPR)},
    month     = {June},
    year      = {2024},
    pages     = {27672-27683}
}

@article{wu2025skysensepp,
    author  = {Wu, Kang and Zhang, Yingying and Ru, Lixiang and Dang, Bo and Lao, Jiangwei and Yu, Lei and Luo, Junwei and Zhu, Zifan and Sun, Yue and Zhang, Jiahao and Zhu, Qi and Wang, Jian and Yang, Ming and Chen, Jingdong and Zhang, Yongjun and Li, Yansheng},
    title   = {A semantic-enhanced multi-modal remote sensing foundation model for Earth observation},
    journal = {Nature Machine Intelligence},
    volume  = {7},
    pages   = {1235-1249},
    year    = {2025},
    doi     = {10.1038/s42256-025-01078-8}
}

@article{zhang2025skysensev2,
    author  = {Zhang, Yingying and Ru, Lixiang and Wu, Kang and Yu, Lei and Liang, Lei and Li, Yansheng and Chen, Jingdong},
    title   = {SkySense V2: A Unified Foundation Model for Multi-modal Remote Sensing},
    journal = {arXiv preprint arXiv:2507.13812},
    year    = {2025}
}

@article{wang2025hypersigma,
  title={Hypersigma: Hyperspectral intelligence comprehension foundation model},
  author={Wang, Di and Hu, Meiqi and Jin, Yao and Miao, Yuchun and Yang, Jiaqi and Xu, Yichu and Qin, Xiaolei and Ma, Jiaqi and Sun, Lingyu and Li, Chenxing and others},
  journal={IEEE Transactions on Pattern Analysis and Machine Intelligence},
  year={2025},
  publisher={IEEE}
}

@article{wang2025complex,
  title={A Complex-valued SAR Foundation Model Based on Physically Inspired Representation Learning},
  author={Wang, Mengyu and Bi, Hanbo and Feng, Yingchao and Xin, Linlin and Gong, Shuo and Wang, Tianqi and Yan, Zhiyuan and Wang, Peijin and Diao, Wenhui and Sun, Xian},
  journal={arXiv preprint arXiv:2504.11999},
  year={2025}
}

@article{li2025saratr,
  title={SARATR-X: Towards building a foundation model for SAR target recognition},
  author={Li, Weijie and Yang, Wei and Hou, Yuenan and Liu, Li and Liu, Yongxiang and Li, Xiang},
  journal={IEEE Transactions on Image Processing},
  year={2025},
  publisher={IEEE}
}

@article{li2024predicting,
  title={Predicting gradient is better: Exploring self-supervised learning for SAR ATR with a joint-embedding predictive architecture},
  author={Li, Weijie and Yang, Wei and Liu, Tianpeng and Hou, Yuenan and Li, Yuxuan and Liu, Zhen and Liu, Yongxiang and Liu, Li},
  journal={ISPRS Journal of Photogrammetry and Remote Sensing},
  volume={218},
  pages={326--338},
  year={2024},
  publisher={Elsevier}
}

@book{jensen_remote_2015,
title = {Introductory Digital Image Processing: A Remote Sensing Perspective},
author = {Jensen, John R.},
edition = {4},
year = {2015},
publisher = {Pearson},
address = {Boston},
isbn = {978-0-13-405816-0}
}

@article{hu2025earth,
  title={Earth-Adapter: Bridge the Geospatial Domain Gaps with Mixture of Frequency Adaptation},
  author={Hu, Xiaoxing and Gong, Ziyang and Wang, Yupei and Jia, Yuru and Luo, Gen and Yang, Xue},
  journal={arXiv preprint arXiv:2504.06220},
  year={2025}
}

@article{zhang2024asanet,
  title={ASANet: Asymmetric Semantic Aligning Network for RGB and SAR image land cover classification},
  author={Zhang, Pan and Peng, Baochai and Lu, Chaoran and Huang, Quanjin and Liu, Dongsheng},
  journal={ISPRS Journal of Photogrammetry and Remote Sensing},
  volume={218},
  pages={574--587},
  year={2024},
  publisher={Elsevier}
}

@article{shi2021object,
  title={Object-level semantic segmentation on the high-resolution Gaofen-3 FUSAR-map dataset},
  author={Shi, Xianzheng and Fu, Shilei and Chen, Jin and Wang, Feng and Xu, Feng},
  journal={IEEE Journal of Selected Topics in Applied Earth Observations and Remote Sensing},
  volume={14},
  pages={3107--3119},
  year={2021},
  publisher={IEEE}
}

@article{wei2024mgfnet,
  title={MGFNet: An MLP-dominated gated fusion network for semantic segmentation of high-resolution multi-modal remote sensing images},
  author={Wei, Kan and Dai, Jinkun and Hong, Danfeng and Ye, Yuanxin},
  journal={International Journal of Applied Earth Observation and Geoinformation},
  volume={135},
  pages={104241},
  year={2024},
  publisher={Elsevier}
}

@inproceedings{xia2023openearthmap,
  title={Openearthmap: A benchmark dataset for global high-resolution land cover mapping},
  author={Xia, Junshi and Yokoya, Naoto and Adriano, Bruno and Broni-Bediako, Clifford},
  booktitle={Proceedings of the IEEE/CVF Winter Conference on Applications of Computer Vision},
  pages={6254--6264},
  year={2023}
}

@article{xie2021segformer,
  title={SegFormer: Simple and efficient design for semantic segmentation with transformers},
  author={Xie, Enze and Wang, Wenhai and Yu, Zhiding and Anandkumar, Anima and Alvarez, Jose M and Luo, Ping},
  journal={NeurIPS},
  volume={34},
  pages={12077--12090},
  year={2021}
}

@inproceedings{hoyer2022hrda,
  title={Hrda: Context-aware high-resolution domain-adaptive semantic segmentation},
  author={Hoyer, Lukas and Dai, Dengxin and Van Gool, Luc},
  booktitle={ECCV},
  pages={372--391},
  year={2022}
}

@article{stewart2023ssl4eo,
  title={Ssl4eo-l: Datasets and foundation models for landsat imagery},
  author={Stewart, Adam and Lehmann, Nils and Corley, Isaac and Wang, Yi and Chang, Yi-Chia and Ait Ali Braham, Nassim Ait and Sehgal, Shradha and Robinson, Caleb and Banerjee, Arindam},
  journal={Advances in Neural Information Processing Systems},
  volume={36},
  pages={59787--59807},
  year={2023}
}

@article{dosovitskiy2020image,
  title={An image is worth 16x16 words: Transformers for image recognition at scale},
  author={Dosovitskiy, Alexey and Beyer, Lucas and Kolesnikov, Alexander and Weissenborn, Dirk and Zhai, Xiaohua and Unterthiner, Thomas and Dehghani, Mostafa and Minderer, Matthias and Heigold, Georg and Gelly, Sylvain and others},
  journal={arXiv preprint arXiv:2010.11929},
  year={2020}
}

@article{xiong2024neural,
  title={Neural plasticity-inspired foundation model for observing the earth crossing modalities},
  author={Xiong, Zhitong and Wang, Yi and Zhang, Fahong and Stewart, Adam J and Hanna, Jo{\"e}lle and Borth, Damian and Papoutsis, Ioannis and Le Saux, Bertrand and Camps-Valls, Gustau and Zhu, Xiao Xiang},
  journal={CoRR},
  year={2024}
}

@article{cong2022satmae,
  title={Satmae: Pre-training transformers for temporal and multi-spectral satellite imagery},
  author={Cong, Yezhen and Khanna, Samar and Meng, Chenlin and Liu, Patrick and Rozi, Erik and He, Yutong and Burke, Marshall and Lobell, David and Ermon, Stefano},
  journal={NeurIPS},
  volume={35},
  pages={197--211},
  year={2022}
}

@inproceedings{reed2023scale,
  title={Scale-mae: A scale-aware masked autoencoder for multiscale geospatial representation learning},
  author={Reed, Colorado J and Gupta, Ritwik and Li, Shufan and Brockman, Sarah and Funk, Christopher and Clipp, Brian and Keutzer, Kurt and Candido, Salvatore and Uyttendaele, Matt and Darrell, Trevor},
  booktitle={Proceedings of the IEEE/CVF International Conference on Computer Vision},
  pages={4088--4099},
  year={2023}
}

@article{liu2024remoteclip,
  title={Remoteclip: A vision language foundation model for remote sensing},
  author={Liu, Fan and Chen, Delong and Guan, Zhangqingyun and Zhou, Xiaocong and Zhu, Jiale and Ye, Qiaolin and Fu, Liyong and Zhou, Jun},
  journal={IEEE Transactions on Geoscience and Remote Sensing},
  year={2024}
}

@article{wang2024mtp,
  title={MTP: Advancing Remote Sensing Foundation Model via Multi-Task Pretraining},
  author={Wang, Di and Zhang, Jing and Xu, Minqiang and Liu, Lin and Wang, Dongsheng and Gao, Erzhong and Han, Chengxi and Guo, Haonan and Du, Bo and Tao, Dacheng and others},
  journal={IEEE Journal of Selected Topics in Applied Earth Observations and Remote Sensing},
  year={2024}
}

@InProceedings{Wei_2024_CVPR,
    author    = {Wei, Zhixiang and Chen, Lin and Jin, Yi and Ma, Xiaoxiao and Liu, Tianle and Ling, Pengyang and Wang, Ben and Chen, Huaian and Zheng, Jinjin},
    title     = {Stronger Fewer \& Superior: Harnessing Vision Foundation Models for Domain Generalized Semantic Segmentation},
    booktitle = {CVPR},
    month     = {June},
    year      = {2024},
    pages     = {28619-28630}
}

@inproceedings{zhang2023hivit,
  title={Hivit: A simpler and more efficient design of hierarchical vision transformer},
  author={Zhang, Xiaosong and Tian, Yunjie and Xie, Lingxi and Huang, Wei and Dai, Qi and Ye, Qixiang and Tian, Qi},
  booktitle={The eleventh international conference on learning representations},
  year={2023}
}

@article{oquab2023dinov2,
  title={Dinov2: Learning robust visual features without supervision},
  author={Oquab, Maxime and Darcet, Timoth{\'e}e and Moutakanni, Th{\'e}o and Vo, Huy and Szafraniec, Marc and Khalidov, Vasil and Fernandez, Pierre and Haziza, Daniel and Massa, Francisco and El-Nouby, Alaaeldin and others},
  journal={arXiv preprint arXiv:2304.07193},
  year={2023}
}

@article{zhirui2025air,
  title={AIR-PolSAR-Seg-2.0: Polarimetric SAR Ground Terrain Classification Dataset for Large-scale Complex Scenes},
  author={Zhirui, WANG and Liangjin, ZHAO and Yuelei, WANG and Xuan, ZENG and Jian, KANG and Jian, YANG and Xian, SUN},
  journal={Journal of Radars},
  volume={14},
  number={2},
  pages={353--365},
  year={2025},
  publisher={Journal of Radars}
}

@article{ren2022dual,
  title={A dual-stream high resolution network: Deep fusion of GF-2 and GF-3 data for land cover classification},
  author={Ren, Bo and Ma, Shibin and Hou, Biao and Hong, Danfeng and Chanussot, Jocelyn and Wang, Jianlong and Jiao, Licheng},
  journal={International Journal of Applied Earth Observation and Geoinformation},
  volume={112},
  pages={102896},
  year={2022},
  publisher={Elsevier}
}

@article{wu2021built,
  title={Built-up area mapping in China from GF-3 SAR imagery based on the framework of deep learning},
  author={Wu, Fan and Wang, Chao and Zhang, Hong and Li, Juanjuan and Li, Lu and Chen, Weirong and Zhang, Bo},
  journal={Remote Sensing of Environment},
  volume={262},
  pages={112515},
  year={2021},
  publisher={Elsevier}
}

@article{li2022mcanet,
  title={MCANet: A joint semantic segmentation framework of optical and SAR images for land use classification},
  author={Li, Xue and Zhang, Guo and Cui, Hao and Hou, Shasha and Wang, Shunyao and Li, Xin and Chen, Yujia and Li, Zhijiang and Zhang, Li},
  journal={International Journal of Applied Earth Observation and Geoinformation},
  volume={106},
  pages={102638},
  year={2022},
  publisher={Elsevier}
}

@article{adamw,
  title={Decoupled weight decay regularization},
  author={Loshchilov, I},
  journal={arXiv preprint arXiv:1711.05101},
  year={2017}
}

@inproceedings{coda,
  title={Coda: Instructive chain-of-domain adaptation with severity-aware visual prompt tuning},
  author={Gong, Ziyang and Li, Fuhao and Deng, Yupeng and Bhattacharjee, Deblina and Ma, Xianzheng and Zhu, Xiangwei and Ji, Zhenming},
  booktitle={European Conference on Computer Vision},
  pages={130--148},
  year={2024},
  organization={Springer}
}

@inproceedings{sta,
  title={Train one, generalize to all: Generalizable semantic segmentation from single-scene to all adverse scenes},
  author={Gong, Ziyang and Li, Fuhao and Deng, Yupeng and Shen, Wenjun and Ma, Xianzheng and Ji, Zhenming and Xia, Nan},
  booktitle={Proceedings of the 31st ACM International Conference on Multimedia},
  pages={2275--2284},
  year={2023}
}

@inproceedings{parsing,
  title={Parsing all adverse scenes: Severity-aware semantic segmentation with mask-enhanced cross-domain consistency},
  author={Li, Fuhao and Gong, Ziyang and Deng, Yupeng and Ma, Xianzheng and Zhang, Renrui and Ji, Zhenming and Zhu, Xiangwei and Zhang, Hong},
  booktitle={Proceedings of the AAAI Conference on Artificial Intelligence},
  volume={38},
  number={12},
  pages={13483--13491},
  year={2024}
}

@article{jia2025can,
  title={Can Generative Geospatial Diffusion Models Excel as Discriminative Geospatial Foundation Models?},
  author={Jia, Yuru and Marsocci, Valerio and Gong, Ziyang and Yang, Xue and Vergauwen, Maarten and Nascetti, Andrea},
  journal={arXiv preprint arXiv:2503.07890},
  year={2025}
}

@article{guan2021domain,
  title={Domain adaptation for medical image analysis: a survey},
  author={Guan, Hao and Liu, Mingxia},
  journal={IEEE Transactions on Biomedical Engineering},
  volume={69},
  number={3},
  pages={1173--1185},
  year={2021},
  publisher={IEEE}
}

@article{simeoni2025dinov3,
  title={Dinov3},
  author={Sim{\'e}oni, Oriane and Vo, Huy V and Seitzer, Maximilian and Baldassarre, Federico and Oquab, Maxime and Jose, Cijo and Khalidov, Vasil and Szafraniec, Marc and Yi, Seungeun and Ramamonjisoa, Micha{\"e}l and others},
  journal={arXiv preprint arXiv:2508.10104},
  year={2025}
}

@article{kumar2012environmental,
  title={Environmental monitoring systems: A review},
  author={Kumar, Anuj and Kim, Hiesik and Hancke, Gerhard P},
  journal={IEEE Sensors Journal},
  volume={13},
  number={4},
  pages={1329--1339},
  year={2012},
  publisher={IEEE}
}

@article{mcgill1998urban,
  title={Urban management in developing countries},
  author={McGill, Ronald},
  journal={Cities},
  volume={15},
  number={6},
  pages={463--471},
  year={1998},
  publisher={Elsevier}
}

@article{li2024sardet,
  title={Sardet-100k: Towards open-source benchmark and toolkit for large-scale sar object detection},
  author={Li, Yuxuan and Li, Xiang and Li, Weijie and Hou, Qibin and Liu, Li and Cheng, Ming-Ming and Yang, Jian},
  journal={Advances in Neural Information Processing Systems},
  volume={37},
  pages={128430--128461},
  year={2024}
}

@article{huang2021qxs,
  title={The QXS-SAROPT dataset for deep learning in SAR-optical data fusion},
  author={Huang, Meiyu and Xu, Yao and Qian, Lixin and Shi, Weili and Zhang, Yaqin and Bao, Wei and Wang, Nan and Liu, Xuejiao and Xiang, Xueshuang},
  journal={arXiv preprint arXiv:2103.08259},
  year={2021}
}

@article{wang2019sar,
  title={SAR-to-optical image translation using supervised cycle-consistent adversarial networks},
  author={Wang, Lei and Xu, Xin and Yu, Yue and Yang, Rui and Gui, Rong and Xu, Zhaozhuo and Pu, Fangling},
  journal={Ieee Access},
  volume={7},
  pages={129136--129149},
  year={2019},
  publisher={IEEE}
}

@article{wang2022air,
  title={AIR-PolSAR-Seg: A large-scale data set for terrain segmentation in complex-scene PolSAR images},
  author={Wang, Zhirui and Zeng, Xuan and Yan, Zhiyuan and Kang, Jian and Sun, Xian},
  journal={IEEE Journal of Selected Topics in Applied Earth Observations and Remote Sensing},
  volume={15},
  pages={3830--3841},
  year={2022},
  publisher={IEEE}
}

@article{xiang2020automatic,
  title={Automatic registration of optical and SAR images via improved phase congruency model},
  author={Xiang, Yuming and Tao, Rongshu and Wang, Feng and You, Hongjian and Han, Bing},
  journal={IEEE Journal of Selected Topics in Applied Earth Observations and Remote Sensing},
  volume={13},
  pages={5847--5861},
  year={2020},
  publisher={IEEE}
}

@inproceedings{wang2018sarptical,
  title={The sarptical dataset for joint analysis of sar and optical image in dense urban area},
  author={Wang, Yuanyuan and Zhu, Xiao Xiang},
  booktitle={IGARSS 2018-2018 IEEE International Geoscience and Remote Sensing Symposium},
  pages={6840--6843},
  year={2018},
  organization={IEEE}
}

@article{zheng2023airborne,
  title={Airborne multi-dimensional SAR land cover dataset and fusion classification method},
  author={Zheng, NR and Yang, Z and Shi, Z and Yang, H and Sun, Y and Wang, F},
  journal={National Remote Sensing Bulletin},
  year={2023}
}

@article{yan2024mpolsar,
  title={MPOLSAR-1.0: Multidimensional SAR multiband fully polarized fine classification dataset},
  author={Yan, JIN and Xiaolan, QIU and Jie, PAN and Songtao, SHANGGUAN and Zezhong, WANG and Wei, WANG and Hong, YANG},
  journal={Journal of Radars},
  volume={13},
  number={3},
  pages={525--538},
  year={2024},
  publisher={Journal of Radars}
}

@article{sar1,
  title={Accurate despeckling and estimation of polarimetric features by means of a spatial decorrelation of the noise in complex PolSAR data},
  author={Arienzo, Alberto and Argenti, Fabrizio and Alparone, Luciano and Gherardelli, Monica},
  journal={Remote Sensing},
  volume={12},
  number={2},
  pages={331},
  year={2020},
  publisher={MDPI}
}

@article{sar2,
  title={Improved Coherent Processing of Synthetic Aperture Radar Data through Speckle Whitening of Single-Look Complex Images},
  author={Alparone, Luciano and Arienzo, Alberto and Lombardini, Fabrizio},
  journal={Remote Sensing},
  volume={16},
  number={16},
  pages={2955},
  year={2024},
  publisher={MDPI}
}

@inproceedings{sar3,
  title={Effects and performance of speckle noise reduction filters on active radar and SAR images},
  author={Mansourpour, Mostafa and Rajabi, MA and Blais, JAR},
  booktitle={Proc. Isprs},
  volume={36},
  number={1},
  pages={W41},
  year={2006}
}

@inproceedings{sar4,
  title={A review of deep-learning techniques for SAR image restoration},
  author={Denis, Lo{\"\i}c and Dalsasso, Emanuele and Tupin, Florence},
  booktitle={2021 IEEE International Geoscience and Remote Sensing Symposium IGARSS},
  pages={411--414},
  year={2021},
  organization={IEEE}
}

@article{lee2005segmentation,
  title={Segmentation of SAR images},
  author={Lee, J-S and Jurkevich, Igor},
  journal={IEEE transactions on Geoscience and Remote Sensing},
  volume={27},
  number={6},
  pages={674--680},
  year={2005},
  publisher={IEEE}
}

@misc{yang2025fusarklip,
  title={FUSAR-KLIP: Towards Multimodal Foundation Models for Remote Sensing}, 
  author={Yi Yang and Xiaokun Zhang and Qingchen Fang and Jing Liu and Ziqi Ye and Rui Li and Li Liu and Haipeng Wang},
  year={2025},
  eprint={2509.23927},
  archivePrefix={arXiv},
  primaryClass={cs.CV},
  url={https://arxiv.org/abs/2509.23927}, 
}

@article{wu2021novel,
  title={A novel method for layover detection in mountainous areas with SAR images},
  author={Wu, Lin and Wang, Hongxia and Li, Yuan and Guo, Zhengwei and Li, Ning},
  journal={Remote Sensing},
  volume={13},
  number={23},
  pages={4882},
  year={2021},
  publisher={MDPI}
}

@article{robertson2023monitoring,
  title={Monitoring autumn agriculture activities using Synthetic Aperture Radar (SAR) and coherence change detection},
  author={Robertson, Laura Dingle and McNairn, Heather and van der Kooij, Marco and Jiao, Xianfeng and Ihuoma, Samuel and Joosse, Pamela},
  journal={Heliyon},
  volume={9},
  number={6},
  year={2023},
  publisher={Elsevier}
}

@article{wegmuller2011progress,
  title={Progress in the understanding of narrow directional microwave scattering of agricultural fields},
  author={Wegm{\"u}ller, U and Santoro, M and Mattia, F and Balenzano, A and Satalino, G and Marzahn, P and Fischer, G and Ludwig, R and Floury, N},
  journal={Remote Sensing of Environment},
  volume={115},
  number={10},
  pages={2423--2433},
  year={2011},
  publisher={Elsevier}
}

@inproceedings{qiu2025noise,
  title={Noise-consistent siamese-diffusion for medical image synthesis and segmentation},
  author={Qiu, Kunpeng and Gao, Zhiqiang and Zhou, Zhiying and Sun, Mingjie and Guo, Yongxin},
  booktitle={Proceedings of the Computer Vision and Pattern Recognition Conference},
  pages={15672--15681},
  year={2025}
}

@inproceedings{qiu2025adaptively,
  title={Adaptively Distilled ControlNet: Accelerated Training and Superior Sampling for Medical Image Synthesis},
  author={Qiu, Kunpeng and Zhou, Zhiying and Guo, Yongxin},
  booktitle={International Conference on Medical Image Computing and Computer-Assisted Intervention},
  pages={55--65},
  year={2025},
  organization={Springer}
}

\clearpage
\beginappendix
\section{CrossEarth-SAR-200K}

CrossEarthSAR-200K is composed of three parts: The 37K private SAR-optical paird dataset named CrossEarth-OPT-SAR, 126K publicly available SAR-optical paired datasets, and 40K publicly available SAR segmentation datasets. Among these data, 126,169 unlabeled images are drawn from QXS-SAROPT~\cite{huang2021qxs}, WHU-SEN-City~\cite{wang2019sar}, SEN12~\cite{Schmitt2019SEN12MS}, Osdataset~\cite{xiang2020automatic}, and SARptical~\cite{wang2018sarptical}. Moreover, 40,042 labeled images are drawn from AIR-PolSAR-Seg 1.0~\cite{wang2022air}, AIR-MDSAR-Map~\cite{zheng2023airborne}, DDHR-X~\cite{ren2022dual}, and MPOLSAR 1.0~\cite{yan2024mpolsar}. These data are released by or collected from different countries and institutions. Therefore, the data sources are extremely diverse, covering virtually all major regions around the world.

For data without semantic segmentation labels, the strongest RSDG model CrossEarth, is adopted to segment the optical images paired with SAR images. The R2U weights trained on 7 classes from the Loveda dataset are used. The resulting predictions are assigned as labels for the corresponding SAR images. Furthermore, all SAR images are cropped or resized to a resolution of five hundred twelve pixels. CrossEarth-SAR-200K-Val is randomly selected as validation for continuous pre-training.

The CrossEarthSAR-200K dataset comprises a total of 203,240 images with corresponding pixel-level annotations, covering seven semantic classes: building, road, water, barren, forest, agricultural, and background. To the best of our knowledge, CrossEarthSAR-200K is the first large-scale SAR semantic segmentation dataset, and its size surpasses that of the widely used COCO-Stuff benchmark (164K images) for general-purpose semantic segmentation. The scale and diversity of CrossEarthSAR-200K effectively emulate real-world deployment scenarios in which SAR semantic segmentation models are applied across multiple data sources, with imagery collected from 109 regions worldwide. CrossEarthSAR-200K provides a robust foundation for training and evaluation, thereby advancing research in SAR semantic segmentation and image understanding. 
The composition and introduction of the CrossEarthSAR-200K are shown in Tab.~\ref{tab:datasets}.

To quantitatively evaluate the confidence of generated pseudo-labels in remote sensing semantic segmentation, we propose a metric called ``Mean Agreement'' that measures the consistency of predictions across multiple foundation models. This metric is calculated as follows:

Let $\mathcal{I} = \{I_1, I_2, \ldots, I_N\}$ represent a set of $N$ optical images sampled from the unlabeled portion of the CrossEarth-SAR-200K dataset, where we set $N=1000$ in our experiments. For each image $I_i$, we denote its total number of pixels as $|I_i|$. Let $\mathcal{M} = \{M_1, M_2, M_3, M_4\}$ represent the set of four foundation models used for semantic segmentation (SatMAE, ScaleMAE, MTP, and CrossEarth).

For each pixel position $(x,y)$ in image $I_i$, let $M_j(I_i, x, y) \in \{1,2,\ldots,7\}$ denote the class label predicted by model $M_j$, where 7 represents the number of semantic categories in our task. We define the agreement indicator function $\mathcal{A}$ for pixel $(x,y)$ as:

$$\mathcal{A}(I_i, x, y) = 
\begin{cases} 
1, & \text{if } M_1(I_i, x, y) = M_2(I_i, x, y) = M_3(I_i, x, y) = M_4(I_i, x, y) \\
0, & \text{otherwise}
\end{cases}$$

The agreement ratio $R_i$ for image $I_i$ is calculated as:

$$R_i = \frac{\sum_{x,y} \mathcal{A}(I_i, x, y)}{|I_i|}$$

Finally, the Mean Agreement across all $N$ images is defined as:

$$\text{Mean Agreement} = \frac{1}{N} \sum_{i=1}^{N} R_i$$

This metric quantifies the proportion of pixels where all four foundation models (SatMAE, ScaleMAE, MTP, and CrossEarth) agree on the same semantic class. A higher Mean Agreement value indicates greater consistency among the models' predictions, suggesting higher confidence in the generated pseudo-labels. This approach leverages the collective intelligence of multiple strong foundation models to assess the reliability of semantic segmentation results in the absence of ground truth labels.

\section{Results}

For the one gap benchmarks, the detailed per-category performance of different methods on the N2S, S2N, K2C and C2K benchmarks are reported in Tab.~\ref{tab:gap1-1}. The detailed per-category performance of different methods on the VV2F, F2VV, HH2F and F2HH benchmarks are reported in Tab.~\ref{tab:gap1-2}. Additionally, the detailed per-category performance of different methods on the C(r)2R, R2C(r), C(i)2R and R2C(i) benchmarks are reported in Tab.~\ref{tab:gap1-3}. 

For the two gaps benchmarks, the detailed per-category performance of different methods on the F2A, A2F, O2D and D2O benchmarks are reported in Tab.~\ref{tab:gap2}. 

For the three gaps benchmarks, the detailed per-category performance of different methods on the D2F, F2D, W2D and D2W benchmarks are reported in Tab.~\ref{tab:gap3}.

These results further demonstrate that CrossEarth-SAR achieves state-of-the-art robustness across regions, radar bands, and polarization modes, as well as superior performance on multi-gaps SAR RSDG tasks.

\section{Visualization}

\begin{figure*}[t]
  \centering
  \includegraphics[width=1.0\textwidth]{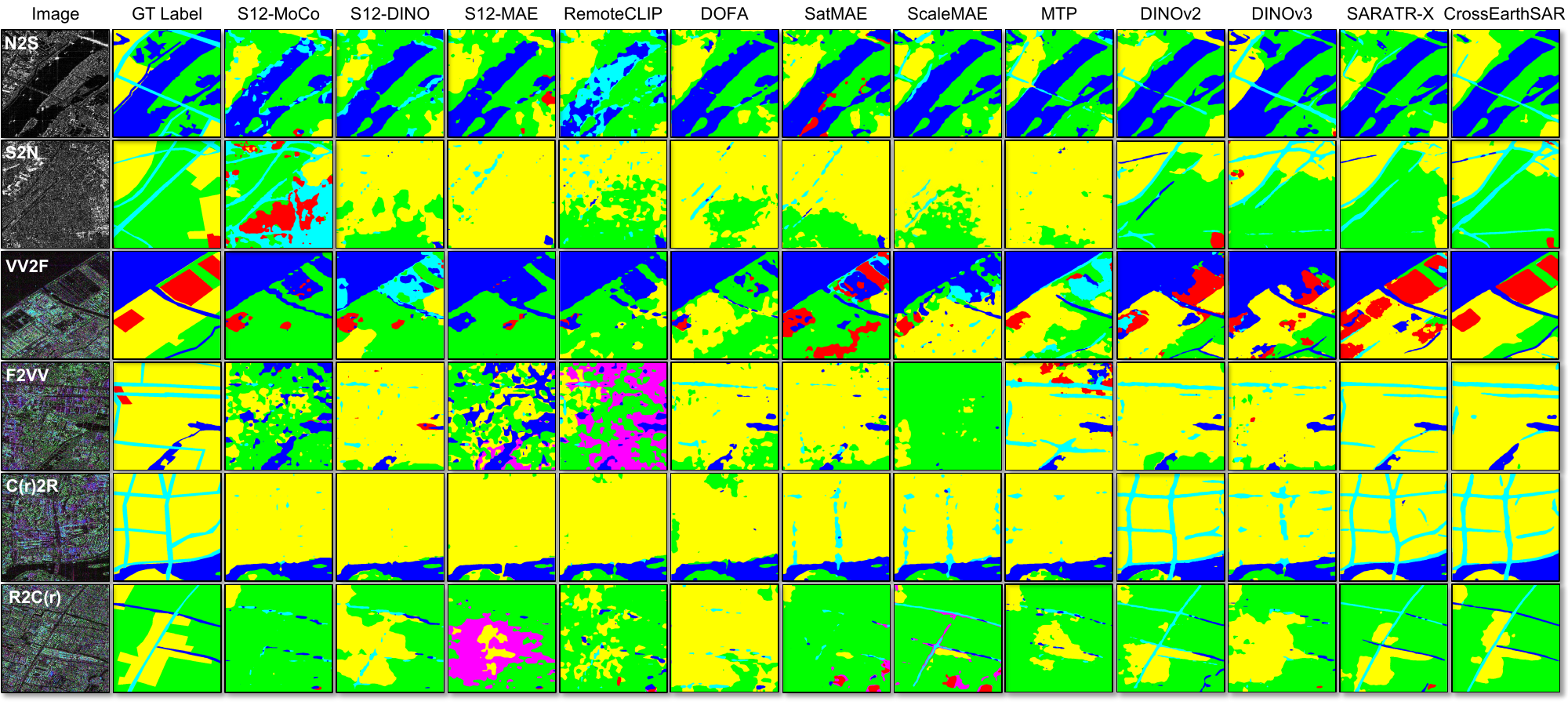}
  \caption{Visualizations of predicted segmentation maps on six representative benchmarks. The \textcolor[RGB]{0, 0, 255}{blue} is the water class, \textcolor[RGB]{0, 255, 0}{green} is the vegetation class, \textcolor[RGB]{255, 0, 0}{red} is the ground class, \textcolor[RGB]{0, 255, 255}{cyan} is the road class, \textcolor[RGB]{245, 245, 0}{yellow} is the building class, and \textcolor[RGB]{255, 0, 255}{purple} is the mountain class.}

  \label{fig:visualization_app1}
\end{figure*}

\begin{figure*}[t]
  \centering
  \includegraphics[width=1.0\textwidth]{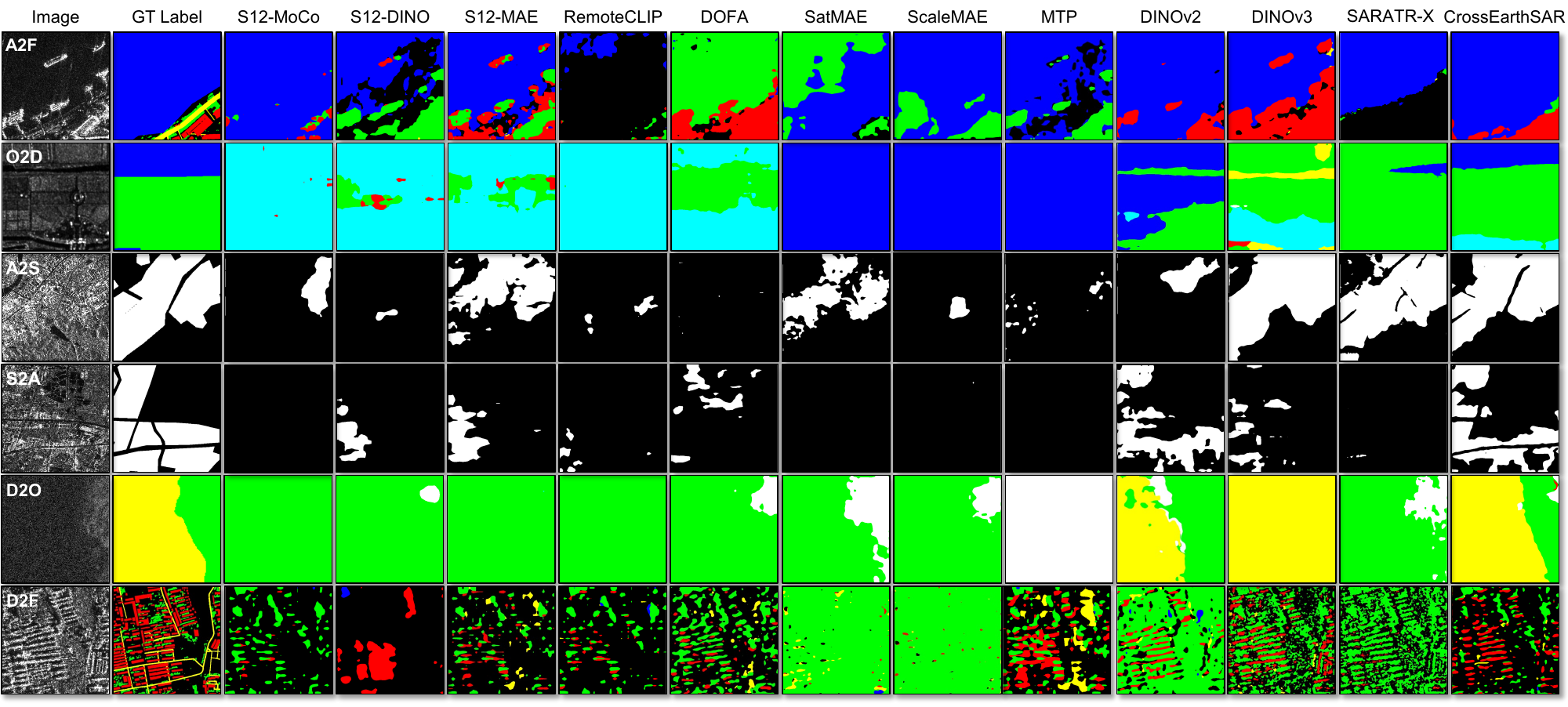}
  \caption{Visualizations of predicted segmentation maps on six representative benchmarks. The \textcolor[RGB]{0, 0, 255}{blue} is the farmland class, \textcolor[RGB]{0, 255, 0}{green} is the greenery class, \textcolor[RGB]{255, 0, 0}{red} is the road class, \textcolor[RGB]{0, 255, 255}{cyan} is the building class, \textcolor[RGB]{245, 245, 0}{yellow} is the water class, and white is the background class.}

  \label{fig:visualization_app2}
\end{figure*}

Fig.~\ref{fig:visualization_app1} and Fig.~\ref{fig:visualization_app2} respectively present the performance of CrossEarth-SAR under one-gap and multi-gap settings, demonstrating its strong superiority over competing methods. Overall, these visual prediction maps provide an intuitive representation of the qualitative findings. Based on this comprehensive set of experiments, we conclude that while CrossEarth is more effective in all gaps than current other models, there are still many challenges to be addressed in SAR RSDG research.


\begin{table*}[t]
\centering
\caption{CrossEarth-SAR-200K source datasets information. Res.:Resolution. GF-3: Gaofen-3. S-1: Sentinel-1. T-X: TerraSAR-X. Air: Airborne SAR.}
\label{tab:datasets}
\scriptsize
\setlength{\tabcolsep}{15pt}
\renewcommand{\arraystretch}{1.2}

\resizebox{1.0\textwidth}{!}{
\begin{tabular}{c c c c c}
\toprule
Datasets & Resolution (m) & Band & Polarization & Satellites \\
\midrule
QXS-SAROPT~\cite{huang2021qxs} & 1m & C & VV & Gaofen-3 \\
WHU-SEN-City~\cite{wang2019sar} & 20-22m & C & VV, VH & Sentinel-1 \\
SEN12~\cite{Schmitt2019SEN12MS} & 10m & C & VV, VH & Sentinel-1 \\
Osdataset~\cite{xiang2020automatic} & 1m & C & VV & Gaofen-3 \\
SARptical~\cite{wang2018sarptical} & 1m & X & VV, HH & TerraSAR-X \\
DDHR-X~\cite{ren2022dual} & 1m & C & VV & Gaofen-3 \\
MPOLSAR-1.0~\cite{yan2024mpolsar} & 0.5-1m & P, L, S, C, X & HH, HV, VH, VV & Airborne SAR \\
AIR-MDSAR-Map~\cite{zheng2023airborne} & 0.2-1m & P, L, S, C & HH, HV, VH, VV & Airborne SAR \\
AIR-PolSAR-Seg 1.0~\cite{wang2022air} & 8m & C & HH, HV, VH, VV & Gaofen-3 \\
CrossEarth-OPT-SAR & 0.5-3m & X & HH & HT-1 \\
\bottomrule
\end{tabular}}
\end{table*}



\begin{figure*}[h]
  \centering
  \includegraphics[width=1.0\textwidth]{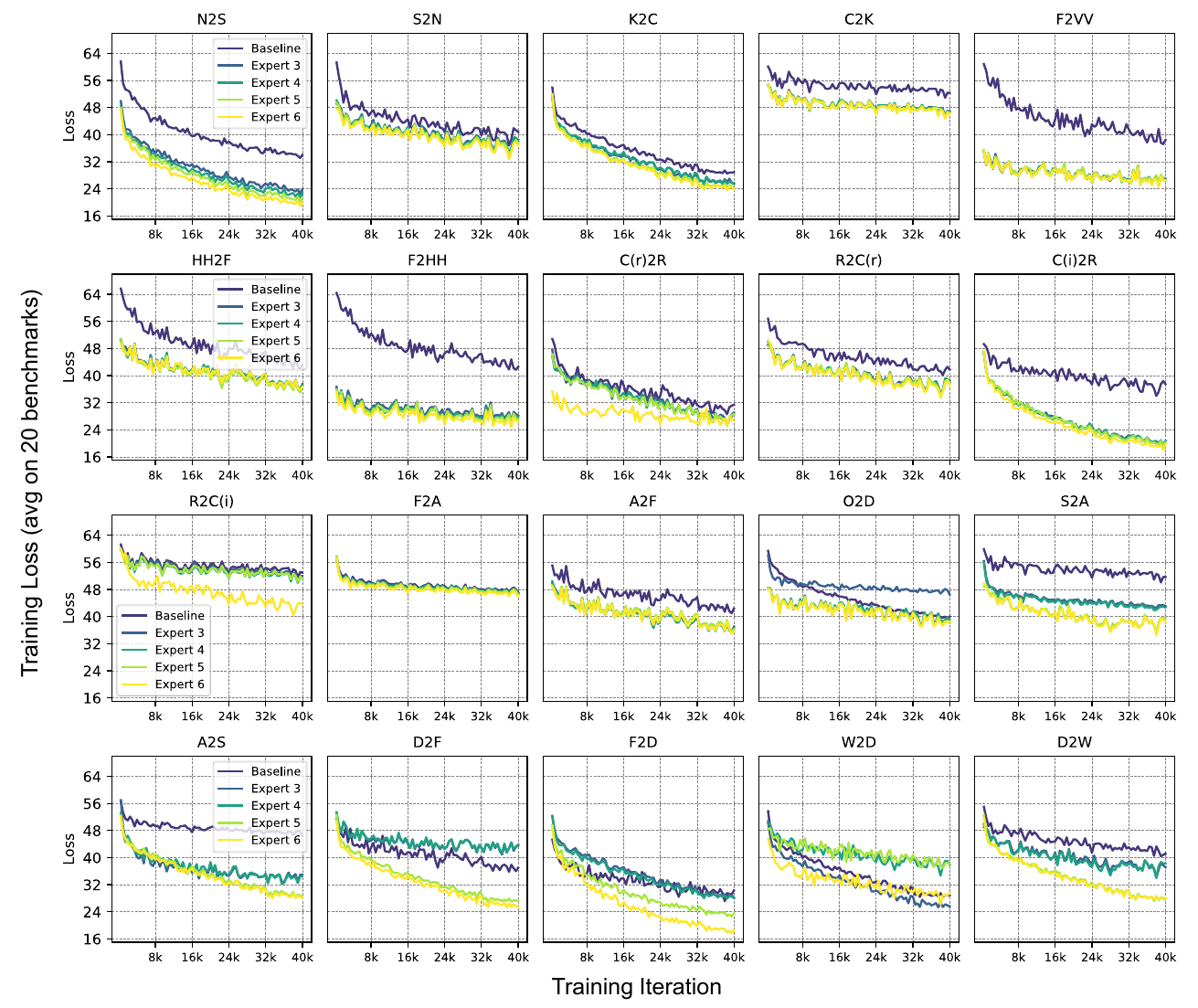}
  \caption{Loss Curve of CrossEarth-SAR in other 20 downstream tasks fine-tuning. CrossEarth-SAR consistently starts and converges at lower losses than Baseline, illustrating the effectiveness of CPT across diverse SAR domains.}

  \label{fig:appendix_loss_all}
\end{figure*}

\clearpage

\begin{table*}[t]
    \centering
    \caption{Performance comparison on benchmark for gaps1-1.
    \label{tab:gap1-1}
    \textbf{Bolds} = best; \underline{underlines} = second best. Bldg:Bulidings. Grnd:Ground. Road:Roads.  Veg:Vegetation. Wtr:Water. Farm:Farmland. Green:Greenery. Otr:Others.}
    \setlength{\tabcolsep}{4pt}
    \resizebox{\textwidth}{!}{%
    \begin{tabular}{ccccccccccccccccc}
    \toprule
    \multirow{3}{*}{Method}                       & \multirow{3}{*}{Backbone}                                                              & \multicolumn{1}{c}{Domain}                                                                 & \multicolumn{5}{c}{Classes}                                                                                                 & \multirow{3}{*}{mIoU (\%)}           & \multicolumn{1}{c}{Domain}                                                                 & \multicolumn{6}{c}{Classes}                                                                                                         & \multirow{3}{*}{mIoU (\%)}           \\
    \cmidrule(lr){3-8}\cmidrule(lr){10-16}
                                                 &                                                                                         & \multirow{2}{*}{Source $\rightarrow$ Unseen}                                               & \multirow{2}{*}{Bldg}                & \multirow{2}{*}{Grnd}                & \multirow{2}{*}{Road}                & \multirow{2}{*}{Veg}                 & \multirow{2}{*}{Wtr}                 &                                       & \multirow{2}{*}{Source $\rightarrow$ Unseen}                                               & \multirow{2}{*}{Bldg}                & \multirow{2}{*}{Farm}                & \multirow{2}{*}{Green}               & \multirow{2}{*}{Otr}                 & \multirow{2}{*}{Road}                & \multirow{2}{*}{Wtr}                 &                                       \\
                                                 &                                                                                         &                                                                                           &                                      &                                      &                                     &                                     &                                     &                                       &                                                                                           &                                     &                                     &                                     &                                     &                                     &                                     &                                       \\
    \midrule
    SARATR-X \cite{hoyer2022hrda}                & HiViT-B \cite{xie2021segformer}                                                        & \multirow{13}{*}{\begin{tabular}[c]{@{}c@{}}N2S\\ (Unseen Region)\end{tabular}}           & 55.6                                 & 5.7                                  & 23.1                                & 31.6                                & 20.3                                & 28.4                                  & \multirow{12}{*}{\begin{tabular}[c]{@{}c@{}}K2C\\ (Unseen Region)\end{tabular}}           & 35.5                                & 0.0                                 & 0.0                                 & 0.0                                 & 0.0                                 & 0.0                                 & 5.9                                   \\
    S12-MoCo \cite{stewart2023ssl4eo}           & ViT-S \cite{dosovitskiy2020image}                                                      &                                                                                           & 54.1                                 & 3.9                                  & 7.3                                 & 32.3                                & 31.6                                & 23.2                                  &                                                                                           & 37.4                                & 0.0                                 & 16.1                                & 0.0                                 & 0.0                                 & 0.0                                 & 8.9                                   \\
    S12-DINO \cite{stewart2023ssl4eo}           & ViT-S \cite{dosovitskiy2020image}                                                      &                                                                                           & 48.0                                 & 4.0                                  & 3.6                                 & 32.6                                & 35.8                                & 22.5                                  &                                                                                           & 43.1                                & 0.0                                 & 12.9                                & 0.0                                 & 0.2                                 & 19.6                                & 12.6                                  \\
    S12-MAE \cite{stewart2023ssl4eo}            & ViT-S \cite{dosovitskiy2020image}                                                      &                                                                                           & 48.5                                 & 4.1                                  & 5.2                                 & 28.7                                & 28.6                                & 22.9                                  &                                                                                           & 35.5                                & 0.0                                 & 0.0                                 & 0.0                                 & 0.0                                 & 0.0                                 & 5.9                                   \\
    DOFA \cite{xiong2024neural}                 & ViT-B \cite{dosovitskiy2020image}                                                      &                                                                                           & 53.0                                 & 12.9                                 & 10.3                                & 28.9                                & 21.4                                & 25.3                                  &                                                                                           & 54.1                                & 0.0                                 & 23.3                                & 0.0                                 & 0.0                                 & 17.3                                & 15.8                                  \\
    SatMAE \cite{cong2022satmae}                & ViT-L \cite{dosovitskiy2020image}                                                      &                                                                                           & 55.2                                 & 5.6                                  & 7.7                                 & 32.5                                & 36.0                                & 26.2                                  &                                                                                           & 60.2                                & 0.0                                 & 23.9                                & 0.0                                 & 0.0                                 & 12.0                                & 14.5                                  \\
    ScaleMAE \cite{reed2023scale}               & ViT-L \cite{dosovitskiy2020image}                                                      &                                                                                           & 55.7                                 & 6.2                                  & 15.5                                & 30.9                                & 32.4                                & 26.4                                  &                                                                                           & 51.1                                & 0.0                                 & 17.6                                & 0.0                                 & 0.1                                 & 8.5                                 & 12.9                                  \\
    RemoteCLIP \cite{liu2024remoteclip}         & ViT-L \cite{dosovitskiy2020image}                                                      &                                                                                           & 45.4                                 & 3.5                                  & 3.4                                 & 29.8                                & 30.0                                & 22.4                                  &                                                                                           & 35.5                                & 0.0                                 & 0.0                                 & 0.0                                 & 0.0                                 & 0.0                                 & 5.9                                   \\
    MTP \cite{wang2024mtp}                      & ViT-L \cite{dosovitskiy2020image}                                                      &                                                                                           & 59.0                                 & 11.9                                 & 30.7                                & 34.0                                & 17.1                                & 30.6                                  &                                                                                           & 68.1                                & 16.3                                & \textbf{28.4}                         & 0.4                                 & 11.2                                & 29.8                                & 25.7                                  \\
    DINOv2 \cite{Wei_2024_CVPR}                 & ViT-L \cite{dosovitskiy2020image}                                                      &                                                                                           & \underline{60.6}                      & \textbf{17.8}                         & 26.5                                & 35.1                                & 21.4                                & 32.3                        &                                                                                           & 68.2                                & \textbf{50.1}                         & 4.8                                 & 1.8                                 & 30.5                                & 48.3                                & 34.0                        \\
    DINOv3 \cite{simeoni2025dinov3}             & ViT-L \cite{dosovitskiy2020image}                                                      &                                                                                           & \textbf{61.47}                        & 7.6                                  & 22.4                                & 33.3                                & 15.7                                & 33.7                                   &                                                                                           & 64.2                                & 15.9                                & \underline{27.2}                      & 0.7                                 & 0.0                                 & 41.1                                & 29.9                                   \\
    \rowcolor[RGB]{235,235,235} CrossEarth-SAR-S & ViT-S \cite{dosovitskiy2020image}                                                      &                                                                                           & 56.5                                 & \underline{17.0}                      & 35.8                                & 35.2                                & 28.3                                & 34.6 (\textcolor{kcgreen}{+2.3})        &                                                                                           & 67.9                                & 45.6                                & 11.2                                & \underline{2.8}                        & 29.0                                & 56.1                                & 35.4 (\textcolor{kcgreen}{+1.4})        \\
    \rowcolor[RGB]{235,235,235} CrossEarth-SAR-B & ViT-B \cite{dosovitskiy2020image}                                                      &                                                                                           & 56.4                                 & 12.3                                 & 35.5                                & \textbf{39.1}                         & \underline{45.1}                      & 36.7 (\textcolor{kcgreen}{+4.4})            &                                                                                           & \underline{69.6}                      & 45.0                                & 15.5                                & 2.6                                 & \underline{36.3}                      & \underline{56.6}                      & 37.7 (\textcolor{kcgreen}{+3.7})            \\
    \rowcolor[RGB]{235,235,235} CrossEarth-SAR-L & ViT-L \cite{dosovitskiy2020image}                                                      &                                                                                           & 56.1                                 & 13.2                                 & \underline{39.0}                      & \underline{37.7}                      & \textbf{49.8}                         & \underline{37.8 (\textcolor{kcgreen}{+5.5})} &                                                                                           & 68.5                                & 41.4                                & 20.4                                & 1.6                                 & 35.9                                & 54.6                                & \underline{38.1 (\textcolor{kcgreen}{+4.1})} \\
    \rowcolor[RGB]{235,235,235} CrossEarth-SAR-L* & ViT-L \cite{dosovitskiy2020image}                                                     &                                                                                           & 56.4                                 & 14.7                                 & \textbf{40.9}                         & 36.6                                & 41.2                                & \textbf{38.0 (\textcolor{kcgreen}{+5.7})}   &                                                                                           & \textbf{71.5}                         & \underline{49.4}                      & 13.2                                & \textbf{3.2}                          & \textbf{36.4}                         & \textbf{61.1}                         & \textbf{39.1 (\textcolor{kcgreen}{+5.1})}   \\
   \midrule   
   SARATR-X \cite{hoyer2022hrda}                & HiViT-B \cite{xie2021segformer}                                                        & \multirow{13}{*}{\begin{tabular}[c]{@{}c@{}}S2N\\ (Unseen Region)\end{tabular}}           & 53.0                                 & 11.4                                 & 30.3                                & 60.2                                & 19.1                                & 34.8                                  & \multirow{12}{*}{\begin{tabular}[c]{@{}c@{}}C2K\\ (Unseen Region)\end{tabular}}           & 42.5                                & 0.0                                 & 9.5                                 & 0.0                                 & 0.1                                 & 20.6                                & 12.1                                  \\
   S12-MoCo \cite{stewart2023ssl4eo}           & ViT-S \cite{dosovitskiy2020image}                                                      &                                                                                           & 38.6                                 & 0.0                                  & 0.3                                 & 59.5                                & 24.5                                & 24.0                                  &                                                                                           & 54.8                                & 0.0                                 & 33.2                                & 0.0                                 & 0.0                                 & 23.5                                & 18.6                                  \\
   S12-DINO \cite{stewart2023ssl4eo}           & ViT-S \cite{dosovitskiy2020image}                                                      &                                                                                           & 45.2                                 & 4.9                                  & 8.8                                 & 52.2                                & 23.7                                & 27.0                                  &                                                                                           & 62.3                                & 10.5                                & 36.0                                & 0.0                                 & 0.3                                 & 75.5                                & 30.8                                  \\
   S12-MAE \cite{stewart2023ssl4eo}            & ViT-S \cite{dosovitskiy2020image}                                                      &                                                                                           & 49.9                                 & 6.2                                  & 13.5                                & 55.6                                & 24.3                                & 29.9                                  &                                                                                           & 72.7                                & \underline{29.6}                      & 17.8                                & 0.0                                 & 4.5                                 & 84.4                                & 34.8                                  \\
   DOFA \cite{xiong2024neural}                 & ViT-B \cite{dosovitskiy2020image}                                                      &                                                                                           & 47.4                                 & 6.7                                  & 15.1                                & 43.5                                & 24.9                                & 27.5                                  &                                                                                           & 71.7                                & 15.3                                & 27.5                                & 0.0                                 & 3.4                                 & 73.3                                & 31.9                                  \\
   SatMAE \cite{cong2022satmae}                & ViT-L \cite{dosovitskiy2020image}                                                      &                                                                                           & 53.4                                 & 15.5                                 & 15.7                                & 54.6                                & 21.1                                & 32.1                                  &                                                                                           & 68.0                                & 0.0                                 & \textbf{43.7}                         & 0.0                                 & 6.8                                 & 81.8                                & 33.4                                  \\
   ScaleMAE \cite{reed2023scale}               & ViT-L \cite{dosovitskiy2020image}                                                      &                                                                                           & \textbf{56.5}                         & 1.7                                  & 9.0                                 & 55.6                                & 13.1                                & 27.2                                  &                                                                                           & 58.4                                & 0.0                                 & \underline{39.8}                      & 0.0                                 & 4.9                                 & 79.5                                & 30.5                                  \\
   RemoteCLIP \cite{liu2024remoteclip}         & ViT-L \cite{dosovitskiy2020image}                                                      &                                                                                           & 46.8                                 & 6.5                                  & 15.7                                & 54.6                                & 27.1                                & 30.2                                  &                                                                                           & 68.3                                & 15.8                                & 29.6                                & 0.0                                 & 0.0                                 & 65.3                                & 29.8                                  \\
   MTP \cite{wang2024mtp}                      & ViT-L \cite{dosovitskiy2020image}                                                      &                                                                                           & 50.8                                 & 13.9                                 & 33.5                                & 60.5                                & 17.6                                & 35.3                                  &                                                                                           & 70.2                                & 21.1                                & 23.6                                & 0.0                                 & 8.5                                 & 80.0                                & 33.9                                  \\
   DINOv2 \cite{Wei_2024_CVPR}                 & ViT-L \cite{dosovitskiy2020image}                                                      &                                                                                           & 55.0                                 & \underline{28.3}                      & 34.4                                & 63.2                                & 38.2                                & 43.8                                  &                                                                                           & 74.1                                & 28.2                                & 15.2                                & 0.0                                 & 10.2                                & \underline{87.7}                      & 35.8                                  \\
   DINOv3 \cite{simeoni2025dinov3}             & ViT-L \cite{dosovitskiy2020image}                                                      &                                                                                           & 51.4                                 & 15.8                                 & 27.7                                & 60.4                                & 33.7                                & 42.8                                  &                                                                                           & 71.3                                & 21.7                                & 21.7                                & 0.0                                 & 7.8                                 & 87.4                                & 35.3                                  \\
   \rowcolor[RGB]{235,235,235} CrossEarth-SAR-S & ViT-S \cite{dosovitskiy2020image}                                                      &                                                                                           & 51.8                                 & 19.7                                 & 41.2                                & \textbf{65.3}                        & 39.1                                & 43.2 (\textcolor{kcred}{-0.6})        &                                                                                           & 75.6                                & \textbf{30.7}                        & 15.9                                & 0.1                                 & 13.9                                & 86.6                                & 35.6 (\textcolor{kcred}{-0.2})        \\
   \rowcolor[RGB]{235,235,235} CrossEarth-SAR-B & ViT-B \cite{dosovitskiy2020image}                                                      &                                                                                           & 53.9                                 & 13.7                                 & \textbf{43.6}                        & 61.8                                & \underline{45.3}                      & 43.4 (\textcolor{kcred}{-0.4})        &                                                                                           & \textbf{77.4}                        & 29.1                                & 15.4                                & 0.1                                 & \underline{18.5}                      & 85.6                                & 37.1 (\textcolor{kcgreen}{+1.3})      \\
   \rowcolor[RGB]{235,235,235} CrossEarth-SAR-L & ViT-L \cite{dosovitskiy2020image}                                                      &                                                                                           & 52.8                                 & 9.0                                  & 42.4                                & 63.4                                & \textbf{45.7}                         & \underline{45.6 (\textcolor{kcgreen}{+1.8})} &                                                                                           & \underline{75.7}                      & 26.8                                & 15.1                                & \underline{0.1}                        & \textbf{19.0}                         & 83.9                                & \underline{38.4 (\textcolor{kcgreen}{+2.6})} \\
   \rowcolor[RGB]{235,235,235} CrossEarth-SAR-L* & ViT-L \cite{dosovitskiy2020image}                                                     &                                                                                           & \underline{55.8}                      & \textbf{29.3}                        & \underline{43.1}                      & \underline{63.7}                      & 41.6                                & \textbf{46.7 (\textcolor{kcgreen}{+2.9})}   &                                                                                           & 74.6                                & 28.9                                & 20.3                                & \textbf{0.1}                          & 18.3                                & \textbf{89.4}                         & \textbf{38.6 (\textcolor{kcgreen}{+2.8})}   \\
   \bottomrule
\end{tabular}}
\end{table*}
\begin{table*}[t]
    \centering
    \caption{Performance comparison on benchmark for gaps1-2.
    \label{tab:gap1-2}
    \textbf{Bolds} = best; \underline{underlines} = second best. Bldg:Bulidings. Grnd:Ground. Mnt:Mountain. Veg:Vegetation. Wtr:Water.}
    \setlength{\tabcolsep}{4pt}
    \resizebox{\textwidth}{!}{%
    \begin{tabular}{cccccccccccccccccc}
    \toprule
    \multirow{3}{*}{Method} & \multirow{3}{*}{Backbone} & \multicolumn{1}{c}{Domain} & \multicolumn{6}{c}{Classes} & \multirow{3}{*}{mIoU (\%)} & \multicolumn{1}{c}{Domain} & \multicolumn{6}{c}{Classes} & \multirow{3}{*}{mIoU (\%)} \\
    \cmidrule(lr){3-9}\cmidrule(lr){11-17}
    & & \multirow{2}{*}{Source $\rightarrow$ Unseen} & \multirow{2}{*}{Bldg} & \multirow{2}{*}{Grnd} & \multirow{2}{*}{Mnt} & \multirow{2}{*}{Road} & \multirow{2}{*}{Veg} & \multirow{2}{*}{Wtr} & & \multirow{2}{*}{Source $\rightarrow$ Unseen} & \multirow{2}{*}{Bldg} & \multirow{2}{*}{Grnd} & \multirow{2}{*}{Mnt} & \multirow{2}{*}{Road} & \multirow{2}{*}{Veg} & \multirow{2}{*}{Wtr} & \\
    \\
    \midrule
    SARATR-X \cite{hoyer2022hrda}                 & HiViT-B \cite{xie2021segformer}                   & \multirow{12}{*}{\begin{tabular}[c]{@{}c@{}}VV2F\\ (Unseen Polarization)\end{tabular}} & 67.3                                 & 35.0                                 & 81.7                                 & 49.3                                 & 58.3                                 & 51.2                                 & 57.1                                      & \multirow{12}{*}{\begin{tabular}[c]{@{}c@{}}HH2F\\ (Unseen Polarization)\end{tabular}} & 61.0                                 & 23.4                                 & 73.3                                 & 33.4                                 & 55.7                                 & 44.6                                 & 48.6                                      \\ 
    S12-MoCo \cite{stewart2023ssl4eo}             & ViT-S \cite{dosovitskiy2020image}                 &                                                                                       & 10.5                                 & 14.6                                 & 27.2                                 & 3.5                                  & 40.2                                 & 35.6                                 & 21.9                                      &                                                                                       & 4.3                                  & 11.9                                 & 33.6                                 & 2.6                                  & 38.2                                 & 28.5                                 & 19.8                                      \\
    S12-DINO \cite{stewart2023ssl4eo}             & ViT-S \cite{dosovitskiy2020image}                 &                                                                                       & 21.3                                 & 7.4                                  & 41.8                                 & 2.3                                  & 32.1                                 & 32.8                                 & 23.0                                      &                                                                                       & 33.0                                 & 0.6                                  & 0.4                                  & 1.6                                  & 32.8                                 & 25.8                                 & 15.7                                      \\
    S12-MAE \cite{stewart2023ssl4eo}              & ViT-S \cite{dosovitskiy2020image}                 &                                                                                       & 2.8                                  & 3.8                                  & 4.0                                  & 1.1                                  & 36.2                                 & 28.6                                 & 12.8                                      &                                                                                       & 3.4                                  & 4.8                                  & 7.0                                  & 0.9                                  & 35.4                                 & 24.1                                 & 12.6                                      \\
    DOFA \cite{xiong2024neural}                   & ViT-B \cite{dosovitskiy2020image}                 &                                                                                       & 46.2                                 & 0.8                                  & 0.3                                  & 0.3                                  & 45.5                                 & 29.4                                 & 20.4                                      &                                                                                       & 52.0                                 & 2.4                                  & 0.0                                  & 0.3                                  & 48.6                                 & 26.5                                 & 21.6                                      \\
    SatMAE \cite{cong2022satmae}                  & ViT-L \cite{dosovitskiy2020image}                 &                                                                                       & 16.0                                 & 5.4                                  & 36.2                                 & 10.0                                 & 31.3                                 & 35.9                                 & 22.5                                      &                                                                                       & 10.4                                 & 5.7                                  & 17.4                                 & 9.2                                  & 33.1                                 & 26.9                                 & 17.1                                      \\
    ScaleMAE \cite{reed2023scale}                 & ViT-L \cite{dosovitskiy2020image}                 &                                                                                       & 11.0                                 & 3.8                                  & 11.5                                 & 13.8                                 & 35.4                                 & 15.2                                 & 15.1                                      &                                                                                       & 16.6                                 & 4.5                                  & 15.0                                 & 17.1                                 & 22.5                                 & 13.8                                 & 14.9                                      \\
    RemoteCLIP \cite{liu2024remoteclip}           & ViT-L \cite{dosovitskiy2020image}                 &                                                                                       & 20.1                                 & 1.5                                  & 0.0                                  & 0.3                                  & 24.5                                 & 33.5                                 & 13.3                                      &                                                                                       & 24.3                                 & 5.0                                  & 1.4                                  & 0.2                                  & 20.4                                 & 26.5                                 & 13.0                                      \\
    MTP \cite{wang2024mtp}                        & ViT-L \cite{dosovitskiy2020image}                 &                                                                                       & 40.1                                 & 6.4                                  & 21.9                                 & 32.1                                 & 32.4                                 & 49.5                                 & 30.4                                      &                                                                                       & 15.3                                 & 7.8                                  & 72.8                                 & 22.9                                 & 38.7                                 & 58.6                                 & 36.0                                      \\
    DINOv2 \cite{Wei_2024_CVPR}                   & ViT-L \cite{dosovitskiy2020image}                 &                                                                                       & 69.4                                 & 37.1                                 & 87.1                                 & 48.5                                 & 67.2                                 & 84.9                                 & 65.7                                      &                                                                                       & 70.7                                 & 34.5                                 & 76.2                                 & 44.4                                 & 66.9                                 & 81.7                                 & 56.8                                      \\
    DINOv3 \cite{simeoni2025dinov3}              & ViT-L \cite{dosovitskiy2020image}                 &                                                                                       & 20.7                                 & 19.4                                 & 64.9                                 & 17.9                                 & 42.9                                 & 47.0                                 & 48.3                                      &                                                                                       & 5.1                                  & 20.9                                 & 64.4                                 & 17.2                                 & 39.8                                 & 34.7                                 & 50.6                                      \\
    \rowcolor[RGB]{235,235,235} CrossEarth-SAR-S & ViT-S \cite{dosovitskiy2020image}                 &                                                                                       & 72.1                                 & 50.7                                 & \underline{91.0}                      & 55.7                                 & 71.3                                 & \underline{87.0}                      & 71.3 (\textcolor{kcgreen}{+5.6})        &                                                                                       & 62.5                                 & 53.1                                 & 90.1                                 & 53.0                                 & 65.4                                 & 86.6                                 & 68.5 (\textcolor{kcgreen}{+11.7})       \\
    \rowcolor[RGB]{235,235,235} CrossEarth-SAR-B & ViT-B \cite{dosovitskiy2020image}                 &                                                                                       & \textbf{77.4}                         & \textbf{58.8}                         & 89.6                                 & \textbf{57.9}                         & \underline{75.8}                      & 84.0                                 & 73.5 (\textcolor{kcgreen}{+7.8})        &                                                                                       & \underline{73.4}                      & 52.8                                 & 84.9                                 & 52.5                                 & \textbf{73.0}                         & \textbf{87.1}                         & 70.6 (\textcolor{kcgreen}{+13.8})       \\
    \rowcolor[RGB]{235,235,235} CrossEarth-SAR-L & ViT-L \cite{dosovitskiy2020image}                 &                                                                                       & 76.3                                 & 52.7                                 & 90.6                                 & 55.8                                 & \textbf{73.5}                         & 86.6                                 & \underline{73.8 (\textcolor{kcgreen}{+8.1})} &                                                                                       & \textbf{73.7}                         & \textbf{55.9}                         & \underline{90.1}                      & \underline{53.3}                      & \underline{72.8}                      & \underline{86.9}                      & \textbf{72.3 (\textcolor{kcgreen}{+15.5})} \\
    \rowcolor[RGB]{235,235,235} CrossEarth-SAR-L* & ViT-L \cite{dosovitskiy2020image}                &                                                                                       & \underline{77.3}                      & \underline{57.5}                      & \textbf{91.5}                         & \underline{56.5}                      & \textbf{76.0}                         & \textbf{88.0}                         & \textbf{73.9 (\textcolor{kcgreen}{+8.2})} &                                                                                       & 73.0                                 & \underline{55.6}                      & \textbf{91.2}                         & \textbf{55.2}                         & 71.3                                 & 84.4                                 & \underline{71.8 (\textcolor{kcgreen}{+15.0})} \\
    \midrule
    SARATR-X \cite{hoyer2022hrda}                 & HiViT-B \cite{xie2021segformer}                   & \multirow{12}{*}{\begin{tabular}[c]{@{}c@{}}F2VV\\ (Unseen Polarization)\end{tabular}} & 59.5                                 & 36.8                                 & 78.8                                 & 40.0                                 & 59.1                                 & 64.7                                 & 56.5                                      & \multirow{12}{*}{\begin{tabular}[c]{@{}c@{}}FF2H\\ (Unseen Polarization)\end{tabular}} & 64.3                                 & 32.3                                 & 76.1                                 & 35.0                                 & 61.5                                 & 68.2                                 & 56.2                                      \\
    S12-MoCo \cite{stewart2023ssl4eo}             & ViT-S \cite{dosovitskiy2020image}                 &                                                                                       & 19.4                                 & 1.5                                  & 0.2                                  & 2.2                                  & 40.5                                 & 33.0                                 & 16.1                                      &                                                                                       & 54.7                                 & 3.9                                  & 27.4                                 & 3.4                                  & 46.0                                 & 52.2                                 & 29.9                                      \\
    S12-DINO \cite{stewart2023ssl4eo}             & ViT-S \cite{dosovitskiy2020image}                 &                                                                                       & 57.8                                 & 18.0                                 & 26.7                                 & 12.8                                 & 41.8                                 & 68.1                                 & 37.5                                      &                                                                                       & 60.2                                 & 9.3                                  & 40.1                                 & 12.5                                 & 31.8                                 & 56.5                                 & 35.1                                      \\
    S12-MAE \cite{stewart2023ssl4eo}              & ViT-S \cite{dosovitskiy2020image}                 &                                                                                       & 34.8                                 & 0.0                                  & 3.9                                  & 1.0                                  & 39.8                                 & 26.8                                 & 17.7                                      &                                                                                       & 31.4                                 & 1.6                                  & 1.4                                  & 1.4                                  & 28.9                                 & 28.9                                 & 15.6                                      \\
    DOFA \cite{xiong2024neural}                   & ViT-B \cite{dosovitskiy2020image}                 &                                                                                       & 34.0                                 & 13.9                                 & 24.0                                 & 13.8                                 & 44.9                                 & 53.3                                 & 30.7                                      &                                                                                       & 49.2                                 & 11.4                                 & 31.7                                 & 19.1                                 & 33.3                                 & 39.0                                 & 30.6                                      \\
    SatMAE \cite{cong2022satmae}                  & ViT-L \cite{dosovitskiy2020image}                 &                                                                                       & 52.4                                 & 11.5                                 & 22.1                                 & 12.6                                 & 37.5                                 & 64.4                                 & 33.4                                      &                                                                                       & 62.6                                 & 7.7                                  & 24.0                                 & 10.1                                 & 42.9                                 & 32.8                                 & 30.0                                      \\
    ScaleMAE \cite{reed2023scale}                 & ViT-L \cite{dosovitskiy2020image}                 &                                                                                       & 17.5                                 & 10.1                                 & 67.9                                 & 6.7                                  & 41.8                                 & 55.8                                 & 33.3                                      &                                                                                       & 20.5                                 & 6.2                                  & 2.6                                  & 7.8                                  & 38.5                                 & 45.6                                 & 20.2                                      \\
    RemoteCLIP \cite{liu2024remoteclip}           & ViT-L \cite{dosovitskiy2020image}                 &                                                                                       & 33.1                                 & 7.5                                  & 4.7                                  & 3.6                                  & 37.0                                 & 39.5                                 & 20.9                                      &                                                                                       & 36.3                                 & 4.1                                  & 6.5                                  & 2.3                                  & 29.1                                 & 25.2                                 & 17.3                                      \\
    MTP \cite{wang2024mtp}                        & ViT-L \cite{dosovitskiy2020image}                  &                                                                                       & 50.2                                 & 20.5                                 & 37.4                                 & 19.7                                 & 53.7                                 & 64.7                                 & 40.2                                      &                                                                                       & 57.6                                 & 17.0                                 & 27.9                                 & 22.9                                 & 23.7                                 & 24.3                                 & 28.9                                      \\
    DINOv2 \cite{Wei_2024_CVPR}                   & ViT-L \cite{dosovitskiy2020image}                 &                                                                                       & 67.9                                 & 34.0                                 & 87.4                                 & 44.8                                 & 63.5                                 & 81.8                                 & 63.2                                      &                                                                                       & 68.9                                 & 15.6                                 & 76.2                                 & 35.6                                 & 63.4                                 & 71.4                                 & 55.2                                      \\
    DINOv3 \cite{simeoni2025dinov3}               & ViT-L \cite{dosovitskiy2020image}                 &                                                                                       & 66.9                                 & 32.1                                 & 72.5                                 & 37.6                                 & 61.6                                 & 68.9                                 & 56.6                                      &                                                                                       & 64.4                                 & 25.5                                 & 59.0                                 & 30.7                                 & 50.0                                 & 67.2                                 & 60.4                                      \\
    \rowcolor[RGB]{235,235,235} CrossEarth-SAR-S & ViT-S \cite{dosovitskiy2020image}                 &                                                                                       & 73.1                                 & 41.5                                 & 86.5                                 & 49.9                                 & 70.4                                 & 86.7                                 & 68.0 (\textcolor{kcgreen}{+4.8})        &                                                                                       & 65.0                                 & 30.9                                 & \textbf{90.4}                         & 44.2                                 & 66.4                                 & \underline{85.3}                      & 65.0 (\textcolor{kcgreen}{+9.8})        \\
    \rowcolor[RGB]{235,235,235} CrossEarth-SAR-B & ViT-B \cite{dosovitskiy2020image}                 &                                                                                       & \underline{74.8}                      & \underline{48.4}                      & 91.2                                 & \underline{51.9}                      & \underline{73.5}                      & 86.4                                 & 68.2 (\textcolor{kcgreen}{+5.0})        &                                                                                       & 66.4                                 & \underline{34.2}                      & 86.2                                 & \underline{49.5}                      & \underline{69.5}                      & 84.7                                 & 66.5 (\textcolor{kcgreen}{+11.3})       \\
    \rowcolor[RGB]{235,235,235} CrossEarth-SAR-L & ViT-L \cite{dosovitskiy2020image}                 &                                                                                       & 74.1                                 & 42.7                                 & \underline{92.6}                      & 50.1                                 & 72.5                                 & 86.8                                 & \underline{69.8 (\textcolor{kcgreen}{+6.6})} &                                                                                       & \textbf{72.6}                         & 30.2                                 & \underline{88.2}                      & 47.7                                 & 65.0                                 & 84.1                                 & \underline{67.1 (\textcolor{kcgreen}{+11.9})} \\
    \rowcolor[RGB]{235,235,235} CrossEarth-SAR-L* & ViT-L \cite{dosovitskiy2020image}                &                                                                                       & \textbf{75.6}                         & \textbf{52.6}                         & \textbf{93.1}                         & \textbf{55.2}                         & \textbf{73.9}                         & \textbf{87.7}                         & \textbf{72.2 (\textcolor{kcgreen}{+9.0})}  &                                                                                       & \underline{69.4}                      & \textbf{46.8}                         & 86.1                                 & \textbf{52.6}                         & \textbf{71.0}                         & \textbf{85.9}                         & \textbf{68.6 (\textcolor{kcgreen}{+13.4})}  \\

    \bottomrule
    \end{tabular}}
    \end{table*}

\begin{table*}[t]
    \centering
    \caption{Performance comparison on benchmark for gaps1-3.
    \label{tab:gap1-3}
    \textbf{Bolds} = best; \underline{underlines} = second best. Bldg:Bulidings. Grnd:Ground. Mnt:Mountain. Veg:Vegetation. Wtr:Water.}
    \setlength{\tabcolsep}{4pt}
    \resizebox{\textwidth}{!}{%
    \begin{tabular}{cccccccccccccccccc}
    \toprule
    \multirow{3}{*}{Method} & \multirow{3}{*}{Backbone}
    & \multicolumn{1}{c}{Domain} & \multicolumn{6}{c}{Classes} & \multirow{3}{*}{mIoU (\%)}
    & \multicolumn{1}{c}{Domain} & \multicolumn{6}{c}{Classes} & \multirow{3}{*}{mIoU (\%)}
    \\
    \cmidrule(lr){3-9}\cmidrule(lr){11-17}
    & &
    \multirow{2}{*}{Source $\rightarrow$ Unseen} & \multirow{2}{*}{Bldg} & \multirow{2}{*}{Grnd} & \multirow{2}{*}{Mnt}& \multirow{2}{*}{Road}& \multirow{2}{*}{Veg} &\multirow{2}{*}{Wtr} & 
    & \multirow{2}{*}{Source $\rightarrow$ Unseen} & \multirow{2}{*}{Bldg} & \multirow{2}{*}{Grnd} & \multirow{2}{*}{Mnt} & \multirow{2}{*}{Road} & \multirow{2}{*}{Veg} & \multirow{2}{*}{Wtr} &
    \\
    \\
    \midrule
    SARATR-X \cite{hoyer2022hrda}                 & HiViT-B \cite{xie2021segformer}                   & \multirow{13}{*}{\begin{tabular}[c]{@{}c@{}}C(r)2R\\ (Unseen Number Type)\end{tabular}}   & 71.0                 & 24.2                 & 0.0                  & 34.7                 & 61.3                 & 60.4                 & 73.8                                         & \multirow{12}{*}{\begin{tabular}[c]{@{}c@{}}C(i)2R\\ (Unseen Number Type)\end{tabular}}   & 74.0                 & 31.6                 & 22.0                 & 37.8                 & 66.0                 & 63.5                 & 72.3                                         \\
    S12-MoCo \cite{stewart2023ssl4eo}             & ViT-S \cite{dosovitskiy2020image}                 &                                                                                           & 59.3                 & 12.4                 & 0.9                  & 7.6                  & 47.0                 & 52.4                 & 29.9                                         &                                                                                           & 59.1                 & 8.0                  & 1.0                  & 7.7                  & 49.2                 & 51.0                 & 29.3                                         \\
    S12-DINO \cite{stewart2023ssl4eo}             & ViT-S \cite{dosovitskiy2020image}                 &                                                                                           & 56.2                 & 4.7                  & 0.0                  & 1.6                  & 46.1                 & 41.5                 & 25.0                                         &                                                                                           & 54.9                 & 4.3                  & 0.0                  & 2.0                  & 41.3                 & 37.8                 & 23.4                                         \\
    S12-MAE \cite{stewart2023ssl4eo}              & ViT-S \cite{dosovitskiy2020image}                 &                                                                                           & 53.8                 & 3.7                  & 0.0                  & 0.8                  & 46.3                 & 37.4                 & 23.7                                         &                                                                                           & 49.7                 & 0.9                  & 0.0                  & 0.3                  & 31.0                 & 35.4                 & 19.5                                         \\
    DOFA \cite{xiong2024neural}                   & ViT-B \cite{dosovitskiy2020image}                 &                                                                                           & 54.3                 & 4.8                  & 0.1                  & 0.6                  & 47.1                 & 40.8                 & 24.6                                         &                                                                                           & 54.3                 & 8.9                  & 0.0                  & 0.3                  & 45.1                 & 39.5                 & 24.7                                         \\
    SatMAE \cite{cong2022satmae}                  & ViT-L \cite{dosovitskiy2020image}                 &                                                                                           & 62.2                 & 22.2                 & 59.5                 & 17.9                 & 50.4                 & 66.0                 & 46.4                                         &                                                                                           & 61.7                 & 24.5                 & 34.3                 & 22.0                 & 49.0                 & 68.9                 & 43.4                                         \\
    ScaleMAE \cite{reed2023scale}                 & ViT-L \cite{dosovitskiy2020image}                 &                                                                                           & 62.7                 & 16.5                 & 7.8                  & 15.6                 & 50.6                 & 67.4                 & 36.8                                         &                                                                                           & 63.6                 & 21.5                 & 66.0                 & 26.5                 & 51.9                 & 70.7                 & 50.0                                         \\
    RemoteCLIP \cite{liu2024remoteclip}          & ViT-L \cite{dosovitskiy2020image}                 &                                                                                           & 53.6                 & 4.5                  & 0.0                  & 2.7                  & 38.1                 & 34.9                 & 22.3                                         &                                                                                           & 50.2                 & 3.1                  & 0.0                  & 0.9                  & 36.6                 & 32.6                 & 20.6                                         \\
    MTP \cite{wang2024mtp}                         & ViT-L \cite{dosovitskiy2020image}                 &                                                                                           & \underline{76.7}      & \underline{54.5}      & 82.3                 & \underline{55.8}      & 70.5                 & 85.0                 & 70.8                                         &                                                                                           & \underline{77.9}      & 55.4                 & 80.2                 & \underline{56.3}      & \underline{72.8}      & 85.6                 & 71.4                                         \\
    DINOv2 \cite{Wei_2024_CVPR}                   & ViT-L \cite{dosovitskiy2020image}                 &                                                                                           & 76.2                 & 51.4                 & 88.3                 & 54.4                 & 71.7                 & 85.5                 & 71.3                                         &                                                                                           & 77.1                 & 53.0                 & \underline{87.3}      & 54.6                 & 72.2                 & \underline{86.3}      & 71.7                                         \\
    DINOv3 \cite{simeoni2025dinov3}               & ViT-L \cite{dosovitskiy2020image}                 &                                                                                           & 75.9                 & 50.1                 & 85.2                 & 51.0                 & 70.5                 & 83.7                 & 69.9                                         &                                                                                           & 76.1                 & 50.5                 & 87.2                 & 50.0                 & 70.9                 & 80.8                 & 69.2                                         \\
    \rowcolor[RGB]{235,235,235} CrossEarth-SAR-S  & ViT-S \cite{dosovitskiy2020image}                 &                                                                                           & 79.4                 & 54.9                 & 91.8                 & 57.7                 & \underline{76.8}      & 86.6                 & 74.5 (\textcolor{kcgreen}{+3.2})              &                                                                                           & 74.1                 & \textbf{80.2}        & 50.6                 & \underline{92.6}      & 57.9                 & 77.4                 & 74.2 (\textcolor{kcgreen}{+2.5})              \\
    \rowcolor[RGB]{235,235,235} CrossEarth-SAR-B  & ViT-B \cite{dosovitskiy2020image}                 &                                                                                           & \underline{79.4}      & \underline{63.1}      & \underline{91.9}      & \underline{58.4}      & 76.4                 & \textbf{87.4}        & 76.1 (\textcolor{kcgreen}{+4.8})              &                                                                                           & 76.4                 & \underline{80.0}      & 62.7                 & 92.5                 & 58.3                 & 77.2                 & \underline{75.1 (\textcolor{kcgreen}{+3.4})}   \\
    \rowcolor[RGB]{235,235,235} CrossEarth-SAR-L  & ViT-L \cite{dosovitskiy2020image}                 &                                                                                           & 79.1                 & 55.5                 & \underline{92.5}      & 57.8                 & 75.6                 & 86.3                 & \underline{76.4 (\textcolor{kcgreen}{+5.1})}   &                                                                                           & \textbf{79.1}        & 57.1                 & \textbf{91.7}        & 58.2                 & \textbf{75.3}        & \textbf{86.3}        & \underline{76.4 (\textcolor{kcgreen}{+4.7})}   \\
    \rowcolor[RGB]{235,235,235} CrossEarth-SAR-L* & ViT-L \cite{dosovitskiy2020image}                 &                                                                                           & \textbf{79.8}        & \textbf{65.6}        & \textbf{93.5}        & \textbf{60.6}        & \textbf{76.9}        & \underline{87.2}      & \textbf{76.9 (\textcolor{kcgreen}{+5.6})}     &                                                                                           & 76.7                 & 79.9                 & 61.1                 & \textbf{93.7}        & 60.7                 & 77.4                 & \textbf{76.7 (\textcolor{kcgreen}{+5.0})}     \\
    \midrule
    SARATR-X \cite{hoyer2022hrda}                 & HiViT-B \cite{xie2021segformer}                   & \multirow{13}{*}{\begin{tabular}[c]{@{}c@{}}R2C(r)\\ (Unseen Number Type)\end{tabular}}   & 62.1                 & 20.2                 & 29.3                 & 24.5                 & 62.2                 & 47.7                 & 73.4                                         & \multirow{12}{*}{\begin{tabular}[c]{@{}c@{}}R2C(i)\\ (Unseen Number Type)\end{tabular}}   & \textbf{77.6}        & 58.8                 & 84.2                 & \underline{56.5}      & 76.7                 & 86.7                 & 72.4                                         \\
    S12-MoCo \cite{stewart2023ssl4eo}             & ViT-S \cite{dosovitskiy2020image}                 &                                                                                           & 58.2                 & 27.8                 & 55.5                 & 21.1                 & 59.6                 & 66.1                 & 48.1                                         &                                                                                           & 55.6                 & 24.0                 & 26.7                 & 8.8                  & 56.6                 & 54.3                 & 37.7                                         \\
    S12-DINO \cite{stewart2023ssl4eo}             & ViT-S \cite{dosovitskiy2020image}                 &                                                                                           & 66.8                 & 34.9                 & 79.4                 & 28.1                 & 66.3                 & 78.3                 & 59.0                                         &                                                                                           & 64.7                 & 30.5                 & 76.5                 & 20.6                 & 64.2                 & 73.3                 & 55.0                                         \\
    S12-MAE \cite{stewart2023ssl4eo}              & ViT-S \cite{dosovitskiy2020image}                 &                                                                                           & 53.4                 & 12.2                 & 18.7                 & 2.7                  & 55.2                 & 57.8                 & 33.3                                         &                                                                                           & 46.3                 & 4.8                  & 0.0                  & 1.3                  & 48.2                 & 36.5                 & 22.8                                         \\
    DOFA \cite{xiong2024neural}                   & ViT-B \cite{dosovitskiy2020image}                 &                                                                                           & 49.4                 & 25.2                 & 57.3                 & 8.6                  & 54.3                 & 61.0                 & 42.6                                         &                                                                                           & 46.6                 & 19.0                 & 8.7                  & 1.3                  & 48.7                 & 33.6                 & 26.3                                         \\
    SatMAE \cite{cong2022satmae}                  & ViT-L \cite{dosovitskiy2020image}                 &                                                                                           & 62.9                 & 14.8                 & 2.6                  & 1.7                  & 58.4                 & 66.0                 & 34.4                                         &                                                                                           & 44.6                 & 21.9                 & 29.9                 & 7.3                  & 50.2                 & 59.7                 & 35.6                                         \\
    ScaleMAE \cite{reed2023scale}                 & ViT-L \cite{dosovitskiy2020image}                 &                                                                                           & 36.7                 & 25.3                 & 48.4                 & 27.9                 & 47.2                 & 58.1                 & 40.6                                         &                                                                                           & 35.4                 & 24.7                 & 19.0                 & 20.3                 & 41.4                 & 42.6                 & 30.6                                         \\
    RemoteCLIP \cite{liu2024remoteclip}          & ViT-L \cite{dosovitskiy2020image}                 &                                                                                           & 52.2                 & 7.5                  & 0.3                  & 3.5                  & 50.1                 & 36.0                 & 24.9                                         &                                                                                           & 58.5                 & 14.4                 & 13.8                 & 6.8                  & 55.9                 & 36.7                 & 31.0                                         \\
    MTP \cite{wang2024mtp}                        & ViT-L \cite{dosovitskiy2020image}                 &                                                                                           & 68.5                 & 51.8                 & 77.4                 & \underline{45.7}      & 68.2                 & 80.6                 & 65.4                                         &                                                                                           & 71.2                 & 50.7                 & 82.2                 & 47.8                 & 70.6                 & 83.4                 & 67.6                                         \\
    DINOv2 \cite{Wei_2024_CVPR}                   & ViT-L \cite{dosovitskiy2020image}                 &                                                                                           & 70.5                 & \textbf{55.7}        & 84.0                 & 45.5                 & 71.0                 & \underline{86.2}      & 68.8                                         &                                                                                           & 70.7                 & 54.4                 & 72.8                 & 46.7                 & 70.2                 & 86.2                 & 66.8                                         \\
    DINOv3 \cite{simeoni2025dinov3}               & ViT-L \cite{dosovitskiy2020image}                 &                                                                                           & 68.1                 & 52.8                 & 79.6                 & 47.8                 & 68.9                 & 83.9                 & 71.9                                         &                                                                                           & 67.8                 & 56.0                 & 76.0                 & 46.8                 & 69.7                 & 84.5                 & 66.8                                         \\
    \rowcolor[RGB]{235,235,235} CrossEarth-SAR-S  & ViT-S \cite{dosovitskiy2020image}                 &                                                                                           & 75.2                 & 59.4                 & \underline{90.4}      & \underline{54.6}      & 75.1                 & 87.2                 & 73.6 (\textcolor{kcgreen}{+4.8})              &                                                                                           & 75.3                 & 56.6                 & \underline{89.5}      & 54.0                 & 74.9                 & 87.5                 & 72.9 (\textcolor{kcgreen}{+6.1})              \\
    \rowcolor[RGB]{235,235,235} CrossEarth-SAR-B  & ViT-B \cite{dosovitskiy2020image}                 &                                                                                           & \underline{75.9}      & \underline{64.0}      & 90.0                 & 53.8                 & \underline{75.6}      & \textbf{87.9}        & 74.5 (\textcolor{kcgreen}{+5.7})              &                                                                                           & 76.0                 & \underline{60.9}      & 88.0                 & 53.3                 & \underline{75.5}      & \underline{88.0}      & 73.5 (\textcolor{kcgreen}{+6.7})              \\
    \rowcolor[RGB]{235,235,235} CrossEarth-SAR-L  & ViT-L \cite{dosovitskiy2020image}                 &                                                                                           & 74.2                 & 55.3                 & \underline{81.7}      & 51.7                 & 73.3                 & \underline{87.2}      & \underline{76.4 (\textcolor{kcgreen}{+7.6})}   &                                                                                           & \underline{74.4}      & 57.2                 & \underline{89.2}      & \underline{52.0}      & \underline{74.0}      & 87.5                 & \underline{73.6 (\textcolor{kcgreen}{+6.8})}   \\
    \rowcolor[RGB]{235,235,235} CrossEarth-SAR-L* & ViT-L \cite{dosovitskiy2020image}                 &                                                                                           & \textbf{77.4}        & \textbf{65.6}        & \textbf{91.9}        & \textbf{56.7}        & \textbf{77.2}        & 8.3                  & \textbf{76.7 (\textcolor{kcgreen}{+7.9})}     &                                                                                           & \underline{77.4}      & \textbf{65.1}        & \textbf{90.2}        & \textbf{56.9}        & \textbf{77.2}        & \textbf{88.5}        & \textbf{75.1 (\textcolor{kcgreen}{+8.3})}     \\
    \bottomrule
    \end{tabular}}
    \end{table*}
    
\begin{table*}[t]
\centering
\caption{Performance comparison on benchmark for gaps2.
\label{tab:gap2}
\textbf{Bolds} = best; \underline{underlines} = second best. Bldg:Bulidings. Otr:Others. Veg:Vegetation. Wtr:Water. Green:Greenery.}
\setlength{\tabcolsep}{4pt}
\resizebox{\textwidth}{!}{%
\begin{tabular}{ccccccccccccccccc}
\toprule
\multirow{3}{*}{Method} & \multirow{3}{*}{Backbone}
& \multicolumn{1}{c}{Domain} & \multicolumn{5}{c}{Classes} & \multirow{3}{*}{mIoU (\%)}
& \multicolumn{1}{c}{Domain} & \multicolumn{6}{c}{Classes} & \multirow{3}{*}{mIoU (\%)}
\\
\cmidrule(lr){3-8}\cmidrule(lr){10-16}
& &
\multirow{2}{*}{Source $\rightarrow$ Unseen} & \multirow{2}{*}{Bldg} & \multirow{2}{*}{Otr} & \multirow{2}{*}{Road}& \multirow{2}{*}{Veg}& \multirow{2}{*}{Wtr} & 
& \multirow{2}{*}{Source $\rightarrow$ Unseen} & \multirow{2}{*}{Bldg} & \multirow{2}{*}{Farm} & \multirow{2}{*}{Green} & \multirow{2}{*}{Otr} & \multirow{2}{*}{Road} & \multirow{2}{*}{Wtr} &
\\
\\
\midrule
SARATR-X \cite{hoyer2022hrda}                 & HiViT-B \cite{xie2021segformer}             & \multirow{13}{*}{\begin{tabular}[c]{@{}c@{}}F2A\\ (Unseen Region and\\ Polarization)\end{tabular}}          & 0.0                     & \textbf{5.6}           & 0.0                     & 0.0                     & 15.2                    & 13.0                                         & \multirow{12}{*}{\begin{tabular}[c]{@{}c@{}}O2D\\ (Unseen Region and\\ Platform)\end{tabular}}          & 5.4                     & 0.0                     & \textbf{22.3}          & 2.1                     & 0.0                     & 2.6                     & 5.4                                          \\
S12-MoCo \cite{stewart2023ssl4eo}            & ViT-S \cite{dosovitskiy2020image}           &                                                                                                            & 0.0                     & 4.3                     & 0.0                     & 17.4                    & 15.1                    & 7.4                                         &                                                                                                            & 0.6                     & 0.0                     & \underline{21.4}       & 0.8                     & 0.0                     & 10.3                    & 5.5                                          \\
S12-DINO \cite{stewart2023ssl4eo}            & ViT-S \cite{dosovitskiy2020image}           &                                                                                                            & 0.0                     & 2.8                     & 0.0                     & 32.6                    & 9.0                     & 8.9                                         &                                                                                                            & 3.6                     & 0.0                     & 15.5                    & 2.5                     & 0.0                     & 4.2                     & 4.3                                          \\
S12-MAE \cite{stewart2023ssl4eo}             & ViT-S \cite{dosovitskiy2020image}           &                                                                                                            & 0.0                     & 3.0                     & 0.0                     & \underline{35.3}        & 22.0                    & 12.1                                        &                                                                                                            & 1.5                     & 0.0                     & 21.2                    & 0.9                     & 0.0                     & 8.7                     & 5.4                                          \\
DOFA \cite{xiong2024neural}                  & ViT-B \cite{dosovitskiy2020image}           &                                                                                                            & 0.0                     & 2.8                     & 0.8                     & 30.3                    & 17.6                    & 10.3                                        &                                                                                                            & 4.5                     & 0.0                     & 20.8                    & 2.6                     & 0.0                     & 0.6                     & 4.8                                          \\
SatMAE \cite{cong2022satmae}                 & ViT-L \cite{dosovitskiy2020image}           &                                                                                                            & 0.0                     & 2.9                     & 0.0                     & 22.8                    & 33.4                    & 11.8                                        &                                                                                                            & 0.0                     & 0.0                     & 20.4                    & 2.0                     & 0.0                     & 0.0                     & 12.1                                         \\
ScaleMAE \cite{reed2023scale}               & ViT-L \cite{dosovitskiy2020image}           &                                                                                                            & 0.0                     & 0.0                     & 0.0                     & 0.3                     & 11.9                    & 2.4                                         &                                                                                                            & 15.5                    & 0.0                     & 14.9                    & 2.4                     & 0.0                     & 26.2                    & 9.9                                          \\
RemoteCLIP \cite{liu2024remoteclip}         & ViT-L \cite{dosovitskiy2020image}           &                                                                                                            & 0.0                     & 3.3                     & 0.0                     & 34.0                    & 25.7                    & \underline{12.6}                             &                                                                                                            & 18.0                    & 0.0                     & 20.5                    & 0.0                     & 0.0                     & 0.0                     & 6.4                                          \\
MTP \cite{wang2024mtp}                       & ViT-L \cite{dosovitskiy2020image}           &                                                                                                            & 0.0                     & 0.2                     & 0.0                     & 1.8                     & 11.0                    & 2.6                                         &                                                                                                            & 0.3                     & 0.0                     & 21.1                    & 2.0                     & 0.0                     & 0.0                     & 3.9                                          \\
DINOv2 \cite{Wei_2024_CVPR}                  & ViT-L \cite{dosovitskiy2020image}           &                                                                                                            & 0.0                     & 4.3                     & 5.8                     & 0.4                     & 48.3                    & 13.4                                        &                                                                                                            & \textbf{51.3}          & 0.0                     & 13.5                    & 2.5                     & 16.6                    & 22.7                    & 17.8                                         \\
DINOv3 \cite{simeoni2025dinov3}              & ViT-L \cite{dosovitskiy2020image}           &                                                                                                            & 0.0                     & \underline{5.1}        & 0.1                     & \textbf{41.7}           & 51.3                    & 15.7                                        &                                                                                                            & 20.8                    & 0.0                     & 1.7                     & 2.5                     & 0.0                     & 36.4                    & 15.3                                         \\
\rowcolor[RGB]{235,235,235} CrossEarth-SAR-S & ViT-S \cite{dosovitskiy2020image}           &                                                                                                            & 0.0                     & 4.6                     & 11.4                    & 14.2                    & \textbf{65.6}           & 14.1 (\textcolor{kcgreen}{+0.7})           &                                                                                                            & \underline{49.7}       & \textbf{0.9}           & 11.5                    & 2.6                     & 18.5                    & 32.8                    & 19.0 (\textcolor{kcgreen}{+1.2})           \\
\rowcolor[RGB]{235,235,235} CrossEarth-SAR-B & ViT-B \cite{dosovitskiy2020image}           &                                                                                                            & 0.7                     & 4.3                     & \underline{12.1}        & 1.6                     & \underline{65.1}        & 15.2 (\textcolor{kcgreen}{+1.8})           &                                                                                                            & 49.1                    & 0.0                     & 15.6                    & 2.6                     & 14.0                    & 38.9                    & 20.0 (\textcolor{kcgreen}{+2.2})           \\
\rowcolor[RGB]{235,235,235} CrossEarth-SAR-L & ViT-L \cite{dosovitskiy2020image}           &                                                                                                            & \textbf{4.5}           & 3.3                     & 7.2                     & 1.6                     & 60.0                    & \textbf{16.9 (\textcolor{kcgreen}{+3.5})}   &                                                                                                            & 42.6                    & 0.0                     & 10.8                    & \underline{2.5}        & \underline{28.9}        & \underline{44.9}        & \textbf{23.7 (\textcolor{kcgreen}{+5.9})}   \\
\rowcolor[RGB]{235,235,235} CrossEarth-SAR-L* & ViT-L \cite{dosovitskiy2020image}          &                                                                                                            & \underline{1.8}        & 3.2                     & \textbf{16.3}           & 6.3                     & 54.2                    & \underline{16.1 (\textcolor{kcgreen}{+2.7})} &                                                                                                            & 33.1                    & \underline{0.0}        & 17.9                    & \textbf{2.6}           & \textbf{31.5}          & \textbf{46.5}          & \underline{23.1 (\textcolor{kcgreen}{+5.3})} \\
\midrule
SARATR-X \cite{hoyer2022hrda}                 & HiViT-B \cite{xie2021segformer}             & \multirow{13}{*}{\begin{tabular}[c]{@{}c@{}}A2F\\ (Unseen Region and\\ Polarization)\end{tabular}}          & 15.8                    & 22.6                    & 6.2                     & \textbf{28.4}          & 27.4                    & 21.3                                        & \multirow{12}{*}{\begin{tabular}[c]{@{}c@{}}D2O\\ (Unseen Region and\\ Platform)\end{tabular}}           & 10.8                    & 0.0                     & 5.4                     & 0.0                     & 5.9                     & 4.2                     & 4.4                                          \\
S12-MoCo \cite{stewart2023ssl4eo}            & ViT-S \cite{dosovitskiy2020image}           &                                                                                                            & 16.2                    & 11.4                    & 0.2                     & 20.9                    & 24.6                    & 14.7                                        &                                                                                                            & 17.9                    & 2.2                     & \underline{37.7}        & 0.0                     & 0.0                     & 27.8                    & 8.1                                          \\
S12-DINO \cite{stewart2023ssl4eo}            & ViT-S \cite{dosovitskiy2020image}           &                                                                                                            & 11.7                    & 43.4                    & 1.5                     & 5.7                     & 55.9                    & 23.6                                        &                                                                                                            & 24.2                    & 8.1                     & 4.1                     & 0.0                     & 0.5                     & 29.4                    & 9.5                                          \\
S12-MAE \cite{stewart2023ssl4eo}             & ViT-S \cite{dosovitskiy2020image}           &                                                                                                            & 12.4                    & 10.1                    & 0.7                     & 20.1                    & 13.6                    & 11.4                                        &                                                                                                            & 12.0                    & 1.7                     & \textbf{39.9}           & 0.0                     & 0.0                     & 10.4                    & 8.8                                          \\
DOFA \cite{xiong2024neural}                  & ViT-B \cite{dosovitskiy2020image}           &                                                                                                            & 10.3                    & \underline{46.6}        & 0.8                     & 2.4                     & 51.3                    & 22.3                                        &                                                                                                            & 14.3                    & 2.4                     & 19.4                    & 0.0                     & 4.2                     & \underline{32.8}        & 10.2                                         \\
SatMAE \cite{cong2022satmae}                 & ViT-L \cite{dosovitskiy2020image}           &                                                                                                            & 15.7                    & 4.8                     & 0.2                     & 16.0                    & 28.9                    & 13.1                                        &                                                                                                            & 13.6                    & 2.9                     & 4.4                     & 0.0                     & \textbf{8.8}            & 13.1                    & 9.9                                          \\
ScaleMAE \cite{reed2023scale}               & ViT-L \cite{dosovitskiy2020image}           &                                                                                                            & 15.8                    & 2.9                     & 0.7                     & \underline{20.9}        & 2.7                     & 8.6                                         &                                                                                                            & 13.5                    & 0.0                     & 17.3                    & 0.0                     & 0.0                     & 18.3                    & 8.2                                          \\
RemoteCLIP \cite{liu2024remoteclip}         & ViT-L \cite{dosovitskiy2020image}           &                                                                                                            & 10.0                    & 42.8                    & 1.0                     & 11.2                    & 43.0                    & 21.6                                        &                                                                                                            & 11.7                    & 0.6                     & 32.5                    & 0.0                     & 0.0                     & 19.4                    & 10.7                                         \\
MTP \cite{wang2024mtp}                       & ViT-L \cite{dosovitskiy2020image}           &                                                                                                            & 1.9                     & \textbf{53.1}          & 0.2                     & 11.6                    & 43.4                    & 22.0                                        &                                                                                                            & 6.2                     & 3.5                     & 6.7                     & 0.0                     & 0.0                     & 10.0                    & 4.1                                          \\
DINOv2 \cite{Wei_2024_CVPR}                  & ViT-L \cite{dosovitskiy2020image}           &                                                                                                            & 16.7                    & 18.1                    & 8.5                     & 6.7                     & 28.4                    & 15.5                                        &                                                                                                            & 18.0                    & 9.5                     & 8.5                     & 0.0                     & \underline{7.1}         & \textbf{35.0}           & 10.1                                         \\
DINOv3 \cite{simeoni2025dinov3}              & ViT-L \cite{dosovitskiy2020image}           &                                                                                                            & 11.2                    & 30.9                    & 6.0                     & 13.6                    & 57.9                    & 23.9                                        &                                                                                                            &                         &                         &                         &                         &                         &                         & 8.8                                          \\
\rowcolor[RGB]{235,235,235} CrossEarth-SAR-S & ViT-S \cite{dosovitskiy2020image}           &                                                                                                            & \underline{20.5}        & 3.2                     & \underline{13.3}        & 15.7                    & \underline{58.3}        & 21.3 (\textcolor{kcgreen}{+5.8})           &                                                                                                            & 13.6                    & \textbf{18.0}           & 1.7                     & 0.0                     & 0.7                     & 6.7                     & \underline{10.7 (\textcolor{kcgreen}{+0.6})} \\
\rowcolor[RGB]{235,235,235} CrossEarth-SAR-B & ViT-B \cite{dosovitskiy2020image}           &                                                                                                            & 18.0                    & 21.7                    & 11.8                    & 7.6                     & 58.1                    & 24.2 (\textcolor{kcgreen}{+8.7})           &                                                                                                            & \textbf{27.8}           & 3.56                    & 10.4                    & 0.0                     & 4.6                     & 16.0                    & 9.4 (\textcolor{kcred}{-0.7})                \\
\rowcolor[RGB]{235,235,235} CrossEarth-SAR-L & ViT-L \cite{dosovitskiy2020image}           &                                                                                                            & 16.1                    & 9.3                     & 9.7                     & 6.5                     & 29.0                    & \underline{25.0 (\textcolor{kcgreen}{+9.5})} &                                                                                                            & 19.9                    & 2.2                     & 14.7                    & \underline{0.0}         & 6.2                     & 16.8                    & 9.6 (\textcolor{kcred}{-0.5})                \\
\rowcolor[RGB]{235,235,235} CrossEarth-SAR-L* & ViT-L \cite{dosovitskiy2020image}          &                                                                                                            & \textbf{20.9}           & 13.5                    & \textbf{13.7}           & 14.9                    & \textbf{71.9}           & \textbf{27.0 (\textcolor{kcgreen}{+11.5})}  &                                                                                                            & \underline{27.4}        & \underline{10.0}        & 2.0                     & \textbf{0.0}            & 3.1                     & 21.6                    & \textbf{11.3 (\textcolor{kcgreen}{+1.2})}    \\

\bottomrule

\end{tabular}}
\end{table*}
\begin{table*}[t]
    \centering
    \caption{Performance comparison on benchmark for gaps3.
    \label{tab:gap3}
    \textbf{Bolds} = best; \underline{underlines} = second best. Bldg:Bulidings. Otr:Others. Veg:Vegetation. Wtr:Water. Farm:Farmland. Green:Greenery.}
    \setlength{\tabcolsep}{4pt}
    \resizebox{\textwidth}{!}{%
    \begin{tabular}{ccccccccccccccccc}
    \toprule
    \multirow{3}{*}{Method} & \multirow{3}{*}{Backbone}
    & \multicolumn{1}{c}{Domain} & \multicolumn{5}{c}{Classes} & \multirow{3}{*}{mIoU (\%)}
    & \multicolumn{1}{c}{Domain} & \multicolumn{6}{c}{Classes} & \multirow{3}{*}{mIoU (\%)}
    \\
    \cmidrule(lr){3-8}\cmidrule(lr){10-16}
    & &
    \multirow{2}{*}{Source $\rightarrow$ Unseen} & \multirow{2}{*}{Bldg} & \multirow{2}{*}{Otr} & \multirow{2}{*}{Road}& \multirow{2}{*}{Veg}& \multirow{2}{*}{Wtr} &
    & \multirow{2}{*}{Source $\rightarrow$ Unseen} & \multirow{2}{*}{Bldg} & \multirow{2}{*}{Farm} & \multirow{2}{*}{Green} & \multirow{2}{*}{Otr} & \multirow{2}{*}{Road} & \multirow{2}{*}{Wtr} &
    \\
    \\
    \midrule
    SARATR-X \cite{hoyer2022hrda}                         & HiViT-B \cite{xie2021segformer}                     & \multirow{13}{*}{\begin{tabular}[c]{@{}c@{}}D2F\\ (Unseen Region, Polarization\\ and Microwave Band)\end{tabular}}                                & 12.3                    & 0.0                     & 0.0                     & 0.0                     & 0.0                     & 21.7                                      & \multirow{12}{*}{\begin{tabular}[c]{@{}c@{}}W2D\\ (Unseen Region, Polarization\\ and Microwave Band)\end{tabular}}                                & 0.0                     & 0.0                     & \underline{23.0}        & 0.0                     & 0.0                     & \underline{50.4}        & 12.2                                    \\
S12-MoCo \cite{stewart2023ssl4eo}                     & ViT-S \cite{dosovitskiy2020image}                   &                                                                                                                                                   & 15.2                    & 0.4                     & 0.0                     & 22.0                    & 1.9                     & 6.7                                      &                                                                                                                                                   & 0.0                     & 0.6                     & 22.2                    & 0.0                     & 0.0                     & 39.0                    & 10.3                                    \\
S12-DINO \cite{stewart2023ssl4eo}                     & ViT-S \cite{dosovitskiy2020image}                   &                                                                                                                                                   & 18.9                    & 19.4                    & 0.0                     & 25.5                    & 13.6                    & 15.5                                     &                                                                                                                                                   & 7.4                     & 11.8                    & 20.3                    & 0.0                     & 0.0                     & 41.7                    & 13.6                                    \\
S12-MAE \cite{stewart2023ssl4eo}                      & ViT-S \cite{dosovitskiy2020image}                   &                                                                                                                                                   & 18.3                    & 3.7                     & 0.2                     & \underline{27.1}        & 48.5                    & 19.6                                     &                                                                                                                                                   & 0.0                     & 0.1                     & 22.4                    & 0.0                     & 0.0                     & 48.8                    & 11.9                                    \\
DOFA \cite{xiong2024neural}                           & ViT-B \cite{dosovitskiy2020image}                   &                                                                                                                                                   & 15.8                    & 4.9                     & 0.1                     & 13.6                    & 7.0                     & 8.3                                      &                                                                                                                                                   & 0.0                     & 0.0                     & 22.5                    & 0.0                     & 0.0                     & \textbf{51.5}           & 12.3                                    \\
SatMAE \cite{cong2022satmae}                          & ViT-L \cite{dosovitskiy2020image}                   &                                                                                                                                                   & 6.9                     & \textbf{43.9}           & 3.7                     & 0.0                     & 0.0                     & 20.3                                     &                                                                                                                                                   & 0.8                     & 0.0                     & 0.0                     & 0.0                     & 0.0                     & 11.9                    & 10.2                                     \\
ScaleMAE \cite{reed2023scale}                         & ViT-L \cite{dosovitskiy2020image}                   &                                                                                                                                                   & 16.6                    & 3.6                     & 2.8                     & 21.9                    & 63.6                    & 18.6                                     &                                                                                                                                                   & 9.1                     & 0.0                     & 20.8                    & 0.0                     & 0.0                     & 14.3                    & 7.4                                     \\
RemoteCLIP \cite{liu2024remoteclip}                  & ViT-L \cite{dosovitskiy2020image}                   &                                                                                                                                                   & 18.5                    & 0.6                     & 0.0                     & \textbf{27.3}           & 62.3                    & 21.7                                     &                                                                                                                                                   & 1.2                     & 0.0                     & 21.9                    & 0.0                     & 0.0                     & 30.0                    & 8.9                                     \\
MTP \cite{wang2024mtp}                                & ViT-L \cite{dosovitskiy2020image}                   &                                                                                                                                                   & 15.5                    & 18.0                    & 0.1                     & 7.7                     & 6.5                     & 7.1                                      &                                                                                                                                                   & 46.2                    & 0.1                     & 6.9                     & 0.2                     & 0.6                     & 21.4                    & 12.5                                    \\
DINOv2 \cite{Wei_2024_CVPR}                           & ViT-L \cite{dosovitskiy2020image}                   &                                                                                                                                                   & 15.0                    & \underline{42.9}        & 2.2                     & 1.4                     & \textbf{68.5}           & \underline{26.0}                         &                                                                                                                                                   & \underline{48.5}        & 13.0                    & 0.3                     & 0.8                     & 1.5                     & 36.3                    & 16.7                                    \\
DINOv3 \cite{simeoni2025dinov3}                       & ViT-L \cite{dosovitskiy2020image}                   &                                                                                                                                                   & 17.7                    & 15.3                    & 0.0                     & 7.1                     & 0.8                     & 22.1                                     &                                                                                                                                                   & 0.0                     & 0.0                     & \textbf{26.1}           & 0.0                     & 0.0                     & 26.5                    & 13.7                                    \\
\rowcolor[RGB]{235,235,235} CrossEarth-SAR-S         & ViT-S \cite{dosovitskiy2020image}                   &                                                                                                                                                   & \underline{20.3}        & 24.9                    & 3.2                     & 4.0                     & 58.5                    & 22.6 (\textcolor{kcred}{-3.4})           &                                                                                                                                                   & 40.0                    & 16.6                    & 21.5                    & 1.7                     & 13.2                    & 26.0                    & 16.4 (\textcolor{kcred}{-0.3})           \\
\rowcolor[RGB]{235,235,235} CrossEarth-SAR-B         & ViT-B \cite{dosovitskiy2020image}                   &                                                                                                                                                   & 19.3                    & 11.5                    & \textbf{6.8}            & 3.0                     & \underline{65.3}        & 21.2 (\textcolor{kcred}{-4.8})           &                                                                                                                                                   & 55.4                    & \underline{24.6}        & 10.1                    & 0.1                     & \textbf{37.1}           & 26.0                    & 20.0 (\textcolor{kcgreen}{+3.3})         \\
\rowcolor[RGB]{235,235,235} CrossEarth-SAR-L         & ViT-L \cite{dosovitskiy2020image}                   &                                                                                                                                                   & \textbf{20.4}           & 30.9                    & \underline{5.1}         & 5.0                     & 55.4                    & 25.1 (\textcolor{kcred}{-0.9})           &                                                                                                                                                   & \underline{67.7}        & 22.4                    & 19.4                    & \textbf{2.6}            & 16.5                    & 35.3                    & \underline{22.2 (\textcolor{kcgreen}{+5.5})} \\
\rowcolor[RGB]{235,235,235} CrossEarth-SAR-L*        & ViT-L \cite{dosovitskiy2020image}                   &                                                                                                                                                   & 19.2                    & 29.6                    & 4.6                     & 7.7                     & 61.9                    & \textbf{26.5 (\textcolor{kcgreen}{+0.5})} &                                                                                                                                                   & \textbf{67.8}           & \textbf{30.8}           & 6.2                     & \underline{2.0}         & \underline{29.6}        & 37.5                    & \textbf{25.6 (\textcolor{kcgreen}{+8.9})} \\

    \midrule
    SARATR-X \cite{hoyer2022hrda}                         & HiViT-B \cite{xie2021segformer}                     & \multirow{13}{*}{\begin{tabular}[c]{@{}c@{}}F2D\\ (Unseen Region, Polarization\\ and Microwave Band)\end{tabular}}                                & 0.0                     & 23.2                    & 1.6                     & 0.3                     & 49.7                    & 15.0                                     & \multirow{12}{*}{\begin{tabular}[c]{@{}c@{}}D2W\\ (Unseen Region, Polarization\\ and Microwave Band)\end{tabular}}                                & 10.3                    & 0.0                     & 0.0                     & 0.0                     & 0.0                     & 5.7                     & 2.7                                     \\
    S12-MoCo \cite{stewart2023ssl4eo}                     & ViT-S \cite{dosovitskiy2020image}                   &                                                                                                                                                   & 0.2                     & 16.5                    & 0.0                     & 18.8                    & 51.4                    & 17.4                                     &                                                                                                                                                   & 11.7                    & 6.8                     & 12.8                    & 0.0                     & 0.0                     & 0.0                     & 5.2                                     \\
    S12-DINO \cite{stewart2023ssl4eo}                     & ViT-S \cite{dosovitskiy2020image}                   &                                                                                                                                                   & 0.0                     & 21.0                    & 0.0                     & 10.1                    & 48.5                    & 15.9                                     &                                                                                                                                                   & 15.0                    & 0.2                     & 28.9                    & 0.0                     & 0.3                     & 16.4                    & 10.1                                    \\
    S12-MAE \cite{stewart2023ssl4eo}                      & ViT-S \cite{dosovitskiy2020image}                   &                                                                                                                                                   & 0.0                     & 15.1                    & 0.0                     & 16.0                    & 49.2                    & 16.1                                     &                                                                                                                                                   & 11.5                    & \textbf{44.2}           & 11.7                    & 0.0                     & 0.0                     & 21.1                    & 14.8                                    \\
    DOFA \cite{xiong2024neural}                           & ViT-B \cite{dosovitskiy2020image}                   &                                                                                                                                                   & 0.0                     & 15.8                    & 2.7                     & 13.9                    & 47.7                    & 16.0                                     &                                                                                                                                                   & 13.2                    & 26.8                    & 22.6                    & 0.0                     & 0.0                     & 12.2                    & 12.5                                    \\
    SatMAE \cite{cong2022satmae}                          & ViT-L \cite{dosovitskiy2020image}                   &                                                                                                                                                   & 0.0                     & 3.1                     & 0.0                     & \textbf{23.1}           & 41.5                    & 17.5                                     &                                                                                                                                                   & 14.9                    & 41.0                    & 5.6                     & 0.0                     & 0.7                     & 24.8                    & 7.7                                     \\
    ScaleMAE \cite{reed2023scale}                         & ViT-L \cite{dosovitskiy2020image}                   &                                                                                                                                                   & 0.0                     & 23.2                    & 0.2                     & 6.4                     & 51.4                    & 16.3                                     &                                                                                                                                                   & 10.0                    & 0.0                     & 6.5                     & 0.0                     & 0.5                     & 14.7                    & 5.3                                     \\
    RemoteCLIP \cite{liu2024remoteclip}                  & ViT-L \cite{dosovitskiy2020image}                   &                                                                                                                                                   & 0.0                     & 17.6                    & 0.0                     & 15.0                    & 45.0                    & 15.5                                     &                                                                                                                                                   & 14.2                    & 6.4                     & 19.9                    & 0.0                     & 0.0                     & 25.8                    & 11.0                                    \\
    MTP \cite{wang2024mtp}                                & ViT-L \cite{dosovitskiy2020image}                   &                                                                                                                                                   & 0.0                     & 0.0                     & 0.0                     & 19.5                    & 20.6                    & 8.0                                      &                                                                                                                                                   & 0.5                     & 27.6                    & 14.2                    & 0.0                     & 0.0                     & 26.4                    & 11.4                                    \\
    DINOv2 \cite{Wei_2024_CVPR}                           & ViT-L \cite{dosovitskiy2020image}                   &                                                                                                                                                   & 7.2                     & \underline{25.3}        & \underline{17.5}        & 0.2                     & \underline{56.9}        & \underline{21.4}                        &                                                                                                                                                   & 10.3                    & 38.3                    & 1.8                     & 0.0                     & 0.3                     & \underline{32.3}        & 13.8                                    \\
    DINOv3 \cite{simeoni2025dinov3}                       & ViT-L \cite{dosovitskiy2020image}                   &                                                                                                                                                   & 0.0                     & 11.8                    & 0.0                     & \underline{21.3}        & 39.4                    & 19.5                                     &                                                                                                                                                   & 2.4                     & 8.9                     & 32.4                    & \textbf{1.9}            & 1.5                     & 19.6                    & 13.4                                    \\
    \rowcolor[RGB]{235,235,235} CrossEarth-SAR-S         & ViT-S \cite{dosovitskiy2020image}                   &                                                                                                                                                   & \underline{9.0}         & 25.2                    & \underline{22.7}        & 0.6                     & 59.3                    & 21.7 (\textcolor{kcgreen}{+0.3})          &                                                                                                                                                   & 15.8                    & 37.3                    & 29.9                    & 0.0                     & \textbf{3.3}            & \textbf{34.0}           & \underline{18.9 (\textcolor{kcgreen}{+5.1})} \\
    \rowcolor[RGB]{235,235,235} CrossEarth-SAR-B         & ViT-B \cite{dosovitskiy2020image}                   &                                                                                                                                                   & 8.1                     & \underline{25.4}        & \textbf{25.3}           & 0.0                     & \textbf{61.7}           & 23.2 (\textcolor{kcgreen}{+1.8})          &                                                                                                                                                   & \underline{16.6}        & 33.8                    & 8.1                     & 0.0                     & 0.3                     & 29.4                    & 15.3 (\textcolor{kcgreen}{+1.5})          \\
    \rowcolor[RGB]{235,235,235} CrossEarth-SAR-L         & ViT-L \cite{dosovitskiy2020image}                   &                                                                                                                                                   & \textbf{14.8}           & \textbf{25.5}           & 20.2                    & 0.2                     & 58.5                    & \textbf{25.0 (\textcolor{kcgreen}{+3.6})} &                                                                                                                                                   & \textbf{21.8}           & \underline{41.9}        & \textbf{37.1}           & 0.0                     & \underline{2.9}         & 30.0                    & 18.0 (\textcolor{kcgreen}{+4.2})          \\
    \rowcolor[RGB]{235,235,235} CrossEarth-SAR-L*        & ViT-L \cite{dosovitskiy2020image}                   &                                                                                                                                                   & 1.7                     & 25.2                    & 18.6                    & 0.6                     & \underline{60.9}        & \underline{24.1 (\textcolor{kcgreen}{+2.7})} &                                                                                                                                                  & 14.3                    & 39.7                    & \underline{36.2}        & \underline{0.1}         & 0.3                     & 32.0                    & \textbf{19.4 (\textcolor{kcgreen}{+5.6})} \\
    
    \bottomrule
    \end{tabular}}
    \end{table*}

\begin{figure*}[t]
  \centering
  \includegraphics[width=0.9\textwidth]{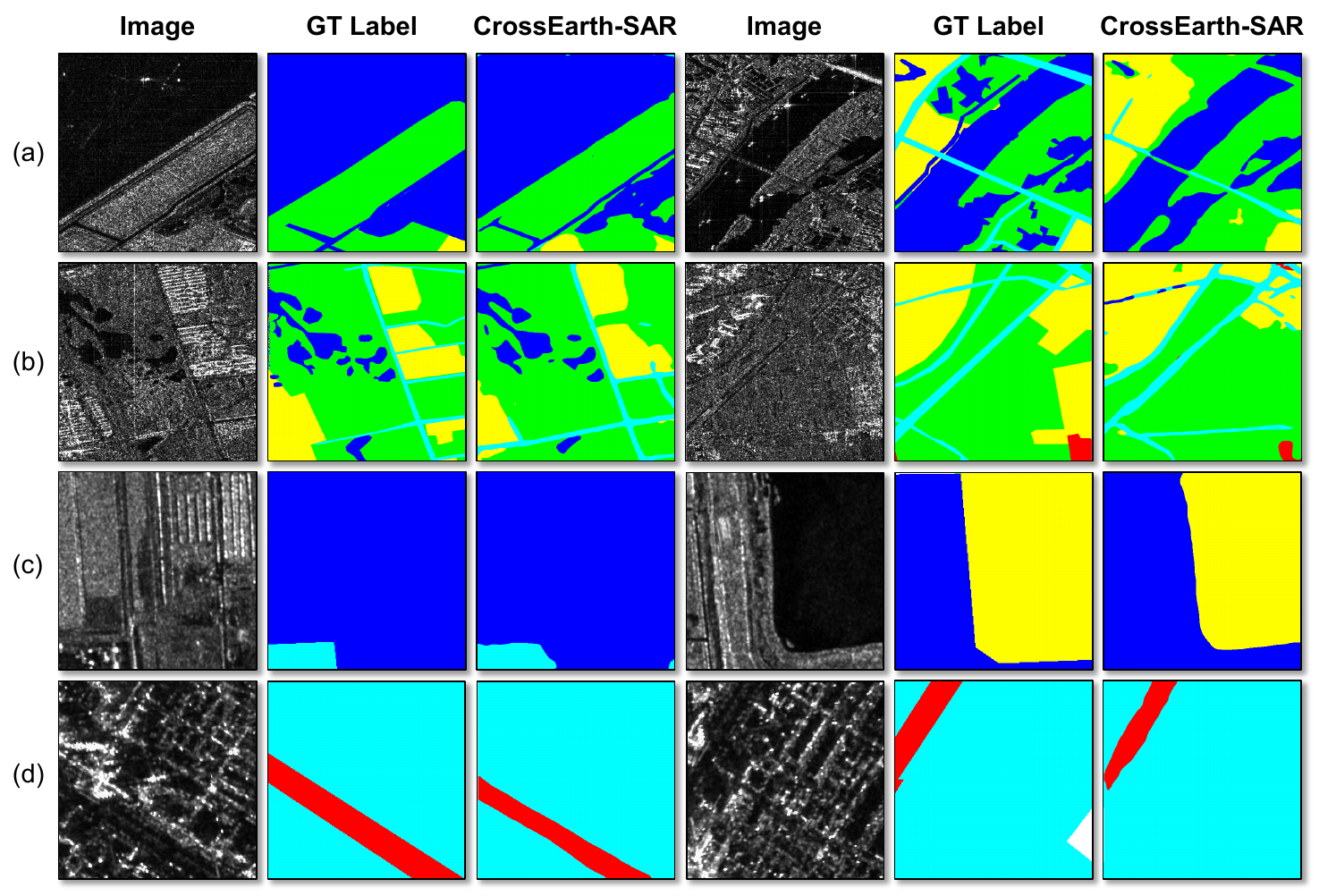}
  \caption{Visualizations of predicted segmentation maps on 1-4 benchmarks. Images in (a) and (b) respectively represent N2S and S2N. The \textcolor[RGB]{0, 0, 255}{blue} is the water class, \textcolor[RGB]{0, 255, 0}{green} is the vegetation class, \textcolor[RGB]{255, 0, 0}{red} is the ground class, \textcolor[RGB]{0, 255, 255}{cyan} is the road class, \textcolor[RGB]{245, 245, 0}{yellow} is the building class, and \textcolor[RGB]{255, 0, 255}{purple} is the mountain class. Images in (c) and (d) respectively represent K2C and C2K. The \textcolor[RGB]{0, 0, 255}{blue} is the farmland class, \textcolor[RGB]{0, 255, 0}{green} is the greenery class, \textcolor[RGB]{255, 0, 0}{red} is the road class, \textcolor[RGB]{0, 255, 255}{cyan} is the building class, \textcolor[RGB]{245, 245, 0}{yellow} is the water class, and white is the background class.}

  \label{fig:appendix_visual1}
\end{figure*}

\begin{figure*}[t]
  \centering
  \includegraphics[width=1.0\textwidth]{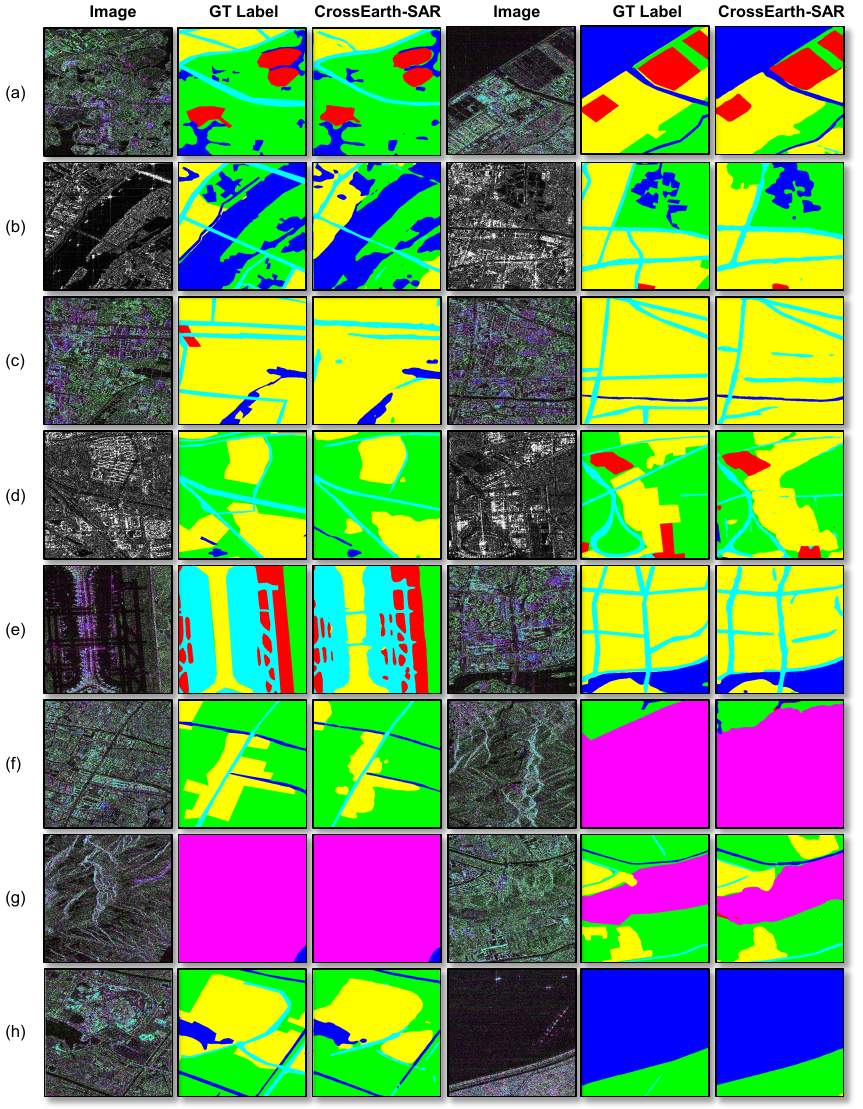}
  \caption{Visualizations of predicted segmentation maps on 5-12 benchmarks. Images in (a), (b), (c), and (d) respectively represent VV2F, F2VV, HH2F, F2HH. Images in (e), (f), (g), and (h) respectively represent C(r)2R, R2C(r), C(i)2R and R2C(i). The \textcolor[RGB]{0, 0, 255}{blue} is the water class, \textcolor[RGB]{0, 255, 0}{green} is the vegetation class, \textcolor[RGB]{255, 0, 0}{red} is the ground class, \textcolor[RGB]{0, 255, 255}{cyan} is the road class, \textcolor[RGB]{245, 245, 0}{yellow} is the building class, and \textcolor[RGB]{255, 0, 255}{purple} is the mountain class.}

  \label{fig:appendix_visual2}
\end{figure*}

\begin{figure*}[t]
  \centering
  \includegraphics[width=1.0\textwidth]{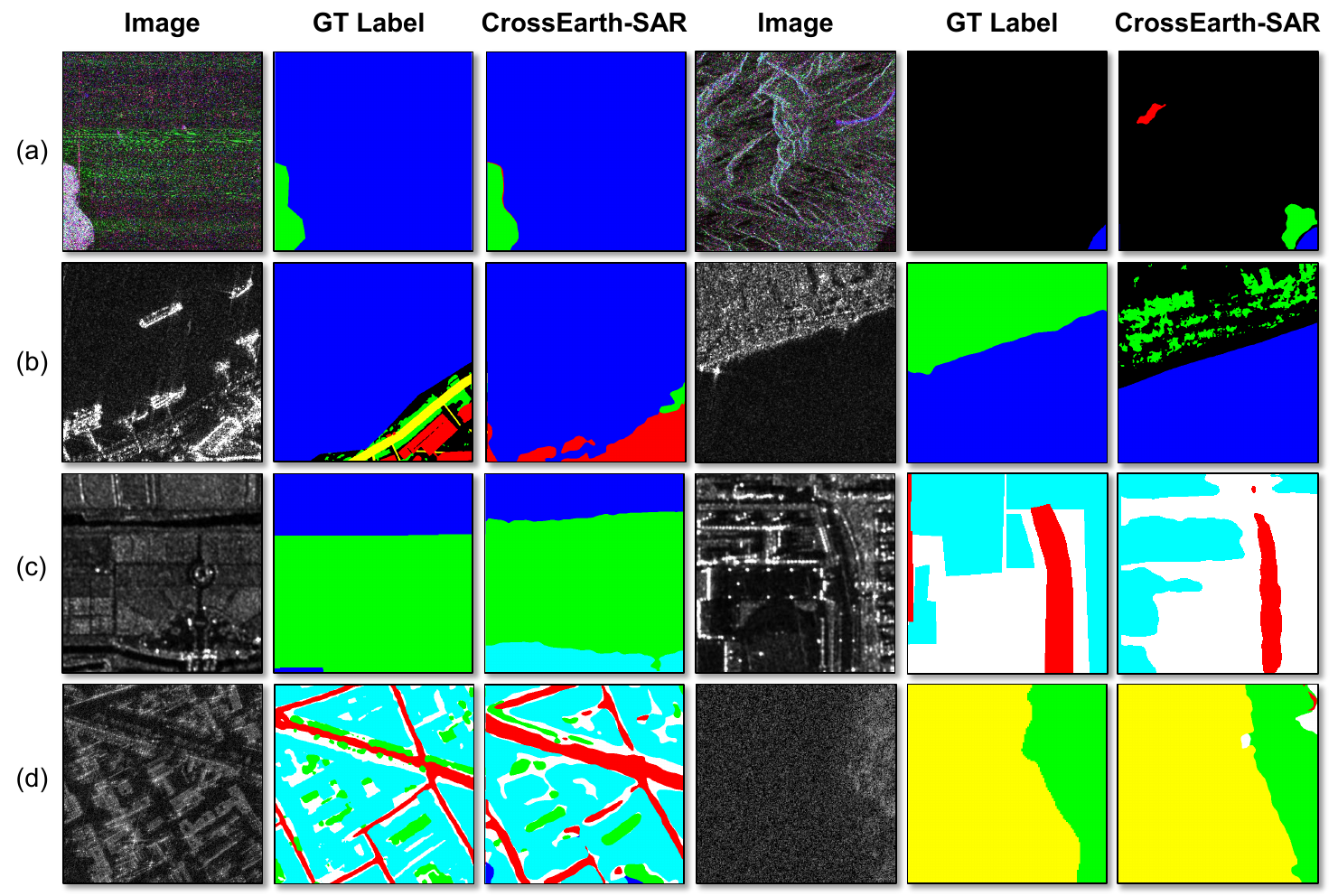}
  \caption{Visualizations of predicted segmentation maps on 13-16 benchmarks. Images in (a) and (b) respectively represent F2A, A2F. The \textcolor[RGB]{0, 0, 255}{blue} is the water class, \textcolor[RGB]{0, 255, 0}{green} is the vegetation class, \textcolor[RGB]{255, 0, 0}{red} is the building class, \textcolor[RGB]{245, 245, 0}{yellow} is the road class, and black is the other class. Images in (c) and (d) respectively represent O2D, D2O. The \textcolor[RGB]{0, 0, 255}{blue} is the farmland class, \textcolor[RGB]{0, 255, 0}{green} is the greenery class, \textcolor[RGB]{255, 0, 0}{red} is the road class, \textcolor[RGB]{0, 255, 255}{cyan} is the building class, \textcolor[RGB]{245, 245, 0}{yellow} is the water class, and white is the background class.}

  \label{fig:appendix_visual3}
\end{figure*}

\begin{figure*}[t]
  \centering
  \includegraphics[width=1.0\textwidth]{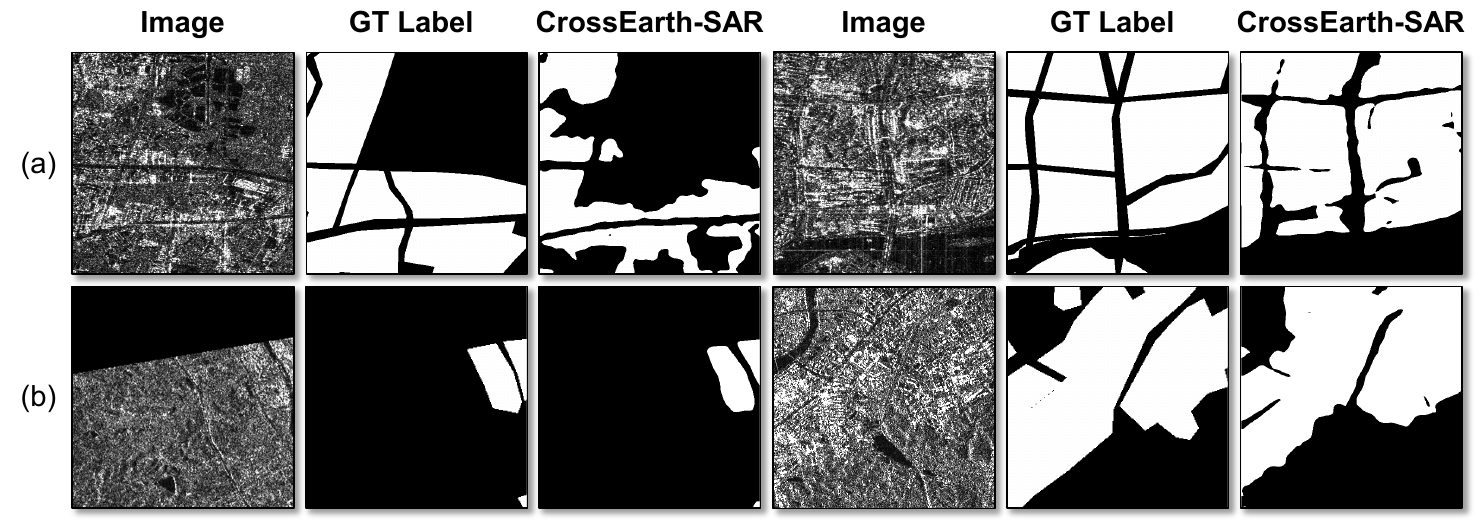}
  \caption{Visualizations of predicted segmentation maps on 17-18 benchmarks. Images in (a) and (b) respectively represent S2A and A2S.  The white is the building class, black is the other class.}

  \label{fig:appendix_visual4}
\end{figure*}

\begin{figure*}[t]
  \centering
  \includegraphics[width=1.0\textwidth]{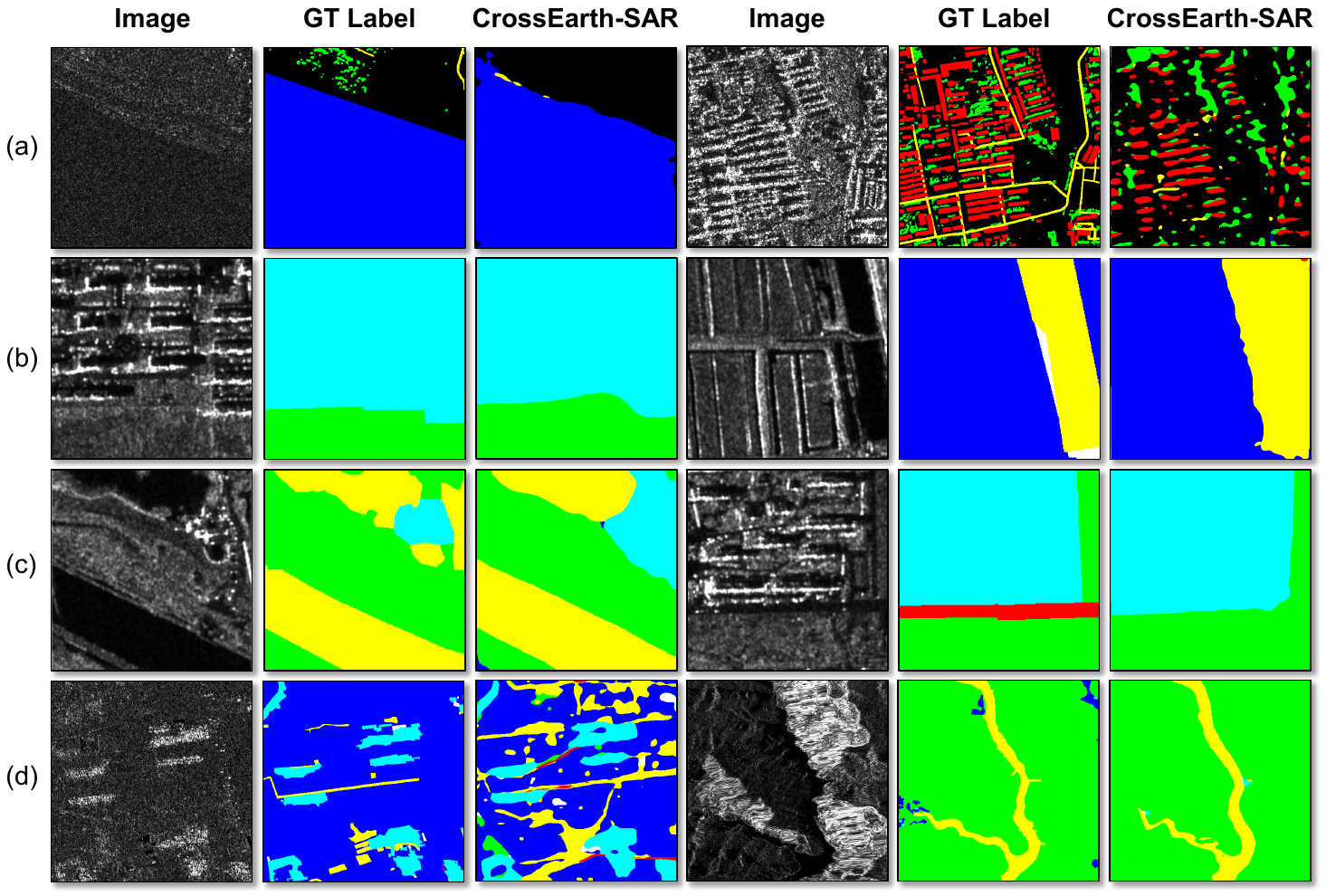}
  \caption{Visualizations of predicted segmentation maps on 19-22 benchmarks. Images in (a), (b), (c), and (d) respectively represent D2F, F2D, W2D and D2W. For the D2F benchmark, the \textcolor[RGB]{0, 0, 255}{blue} is the water class, \textcolor[RGB]{0, 255, 0}{green} is the vegetation class, \textcolor[RGB]{255, 0, 0}{red} is the building class, \textcolor[RGB]{245, 245, 0}{yellow} is the road class, and black is the other class. For the other three banechmarks, the \textcolor[RGB]{0, 0, 255}{blue} is the farmland class, \textcolor[RGB]{0, 255, 0}{green} is the greenery class, \textcolor[RGB]{255, 0, 0}{red} is the road class, \textcolor[RGB]{0, 255, 255}{cyan} is the building class, \textcolor[RGB]{245, 245, 0}{yellow} is the water class, and white is the background class.}

  \label{fig:appendix_visual5}
\end{figure*}

\end{document}